\def\revmode{clean}
\documentclass[journal]{IEEEtran}

\newif\ifarxivversion
\arxivversiontrue

\ifarxivversion
  \def\ARXIVCOMBINED{1}
\fi

\usepackage{amsmath,amssymb}
\usepackage{amsthm}
\newtheorem{proposition}{Proposition}
\usepackage{graphicx}

\graphicspath{{./}{figures/}}
\usepackage{booktabs}
\usepackage{multirow}
\usepackage{subcaption}
\usepackage{pgfplots}
\pgfplotsset{compat=1.18}
\usepackage{tikz}
\usetikzlibrary{arrows.meta, positioning, shapes.geometric, calc, fit, backgrounds}
\usepackage{stfloats}
\renewcommand{\topfraction}{0.92}
\renewcommand{\bottomfraction}{0.75}
\renewcommand{\textfraction}{0.06}
\renewcommand{\floatpagefraction}{0.80}
\renewcommand{\dbltopfraction}{0.92}
\renewcommand{\dblfloatpagefraction}{0.80}
\usepackage{xcolor}
\usepackage{colortbl}
\definecolor{useGPU}{HTML}{DCEBF7}\definecolor{useINT}{HTML}{FBE4D5}
\usepackage{url}
\usepackage{revmarks}
\ifarxivversion
  \usepackage{placeins}
  \definecolor{rowhl}{HTML}{E6F0F8}\definecolor{colhl}{HTML}{EFEFEF}
  \usepackage[hidelinks]{hyperref}
\else
  \usepackage{xr-hyper}
  \usepackage[hidelinks]{hyperref}
\fi
\extrafloats{60}

\definecolor{myblue}{RGB}{31,119,180}
\definecolor{myorange}{RGB}{214,104,14}
\definecolor{mygreen}{RGB}{44,160,44}
\definecolor{mygray}{RGB}{130,130,130}

\begin{document}
\ifarxivversion
  \newcommand{\submissionnote}{}
\else
  \newcommand{\submissionnote}{\thanks{Manuscript submitted September 8, 2026\add{; revised October 30, 2026}.}}
\fi

\title{\chg{Cross-Architecture Foundation-Model Distillation for Edge Flood Segmentation}{Distilling a Geospatial Foundation Model into Compact Surface-Water Segmenters for Edge Flood Mapping}}

\author{Fabian Schmalstieg, Karsten Müller, and Wojciech Samek%
\thanks{Fabian Schmalstieg, Karsten Müller, and Wojciech Samek are with the Department of Artificial Intelligence, Fraunhofer Heinrich Hertz Institute,
10587 Berlin, Germany (e-mail: fabian.schmalstieg@hhi.fraunhofer.de;
karsten.mueller@hhi.fraunhofer.de; samek@hhi.fraunhofer.de).
\chg{Corresponding author: Fabian Schmalstieg.}{(Corresponding author: Fabian Schmalstieg.)}}%
\submissionnote}

\ifarxivversion\else
\markboth{IEEE Journal of Selected Topics in Applied Earth Observations and Remote Sensing}%
{Schmalstieg: Distilling a Geospatial Foundation Model into Compact Surface-Water Segmenters}
\fi

\maketitle
\revlegend

\begin{abstract}
\begin{oldtext}
\revst{Geospatial foundation models can provide strong flood-segmentation performance, but their size limits deployment on memory-constrained edge hardware. We distill a 300-million-parameter Prithvi-EO-2.0 teacher, fine-tuned on the 252 manually labeled Sen1Floods11 training scenes, into a 0.7-million-parameter EfficientViT-B0 student. The teacher supervises additional unlabeled Sentinel-2 imagery, allowing the student training set to grow without new manual annotations. At the matched budget of 252 scenes, teacher-supervised training is competitive with direct training and improves STURM-Flood performance across tested configurations; a geometry-matched control shows that label source alone does not explain the difference. Scaling the teacher-supervised pool to 2,500 scenes narrows the remaining student--teacher gap: the float student reaches 0.787 water intersection over union on the Sen1Floods11 test split against 0.822 for the teacher, matches the teacher on STURM-Flood under our evaluation protocol, and remains below it on WorldFloods-v2. After activation replacement and quantization-aware training, the student runs as a 1.5-megabyte 8-bit integer (INT8) TensorRT engine on a Jetson Xavier NX at 5.57 milliseconds of graphics processing unit (GPU) compute per 512-by-512 image, with approximately 14 megabytes of runtime device memory. A fixed modified normalized difference water index (MNDWI) threshold is competitive with both models on the two clean external benchmarks, so we interpret those benchmarks as generalization tests rather than as evidence of learned-model superiority over a spectral rule. The results support the conclusion: foundation-model supervision can amplify a fixed manual annotation budget into a substantially larger training set and yield a compact, deployable edge model.}
\end{oldtext}
\begin{newtext}
Geospatial foundation models segment surface water accurately, but their size limits deployment on edge hardware. Therefore, we distill a 300-million-parameter Prithvi-EO-2.0 teacher, fine-tuned on the 252 hand-labeled Sen1Floods11 training images, into compact student models. The teacher labels 2,500 Sentinel-2 images around flood alerts and supervises the students via its labels and probability maps, without manual annotation. A 0.7-million-parameter EfficientViT-B0 student achieves the teacher's accuracy in water intersection over union (IoU) on WorldFloods-v2, STURM-Flood and 97 held-back WorldFloods events. Unlike prior work, we evaluate every model on four test sets from three public datasets and on two new dry-scene sets. Controlled comparisons show that teacher labels achieve human-level annotation quality for student training on the same images. The teacher's probability maps account for much of its value, and the full pipeline adds 5.7 IoU points on the held-back events over training on hand labels alone. All students, with parameter counts ranging from about 1,000 to 32.5 million, perform within 2 IoU points of the teacher on WorldFloods-v2 and STURM-Flood. On a Jetson Xavier NX, the main student runs as a 1.6-megabyte mixed 8-bit integer and 16-bit floating-point engine at 4.9 milliseconds per 512-by-512 image, and a 12,230-parameter student runs in full 8-bit integer format within 0.5 IoU points of its 32-bit version. The main student outperforms a fixed water index by 11.5 IoU points on the held-back events and raises false alarms on under 10\% rather than 57.0\% of water-free windows, as unattended use requires.
\end{newtext}
\end{abstract}

\begin{IEEEkeywords}
edge inference, \chg{flood segmentation}{flood mapping}, geospatial foundation models, knowledge distillation, model compression, pseudo-labeling, \chg{quantization-aware training}{quantization}, Sentinel-2\add{, surface-water segmentation}.
\end{IEEEkeywords}

\section{Introduction}

\IEEEPARstart{F}{lood} is among the most frequent and most costly natural disasters globally, and
timely spatial mapping of inundated areas directly informs evacuation routing,
damage assessment, and relief logistics. Operational flood response typically
combines numerical weather prediction, hydrological runoff modeling, and hydraulic
inundation simulation; satellite-based segmentation provides complementary
observational evidence that is not dependent on model assumptions and can validate
or correct model output after an event.

For satellite observation, both synthetic aperture radar (SAR) and optical sensors are used.
Sentinel-1 SAR can penetrate cloud cover, which is important during active precipitation
events, while Sentinel-2 multispectral imagery
provides a richer spectral signature and is the basis for most large-scale
geospatial pretraining. In particular, the Harmonized Landsat Sentinel-2 (HLS)
corpus on which Prithvi-EO-2.0 was pretrained covers six spectral bands
(B2, B3, B4, B8A, B11, B12), and the same six-band input is used throughout
this work. Real-time acquisition during cloud-covered flood events remains a
practical limitation of this modality; here we focus on the mapping and
compression problem, treating near-real-time applicability as future work.

On the modeling side, geospatial foundation models (GFMs) pretrained on large
multispectral satellite corpora generalize well to flood events and geographic
regions not seen during fine-tuning~\cite{szwarcman2025prithvi}. Prithvi-EO-2.0 (300\,M
parameters), fine-tuned on the Sen1Floods11 labeled dataset, \chg{reaches 0.755 mean
intersection over union (mIoU) on our protocol
of the STURM out-of-distribution (OOD) benchmark, a considerably harder evaluation set
than the Sen1Floods11 test split}{serves as our teacher. Under our protocol, its water
intersection over union (IoU) per flood event is 79.1 on the Sen1Floods11 test split. On the two out-of-distribution (OOD)
benchmarks, it reaches 79.4 on the WorldFloods test split and 49.9 on the more difficult STURM-Flood evaluation set}.

The difficulty lies in deployment. Running flood segmentation close to the sensor,
rather than in a data center, shortens the path between acquisition and an
actionable map and removes the dependency on connectivity, which is often degraded
in a disaster zone. On-board inference for Earth observation is already being
demonstrated, for example in the European Space Agency (ESA) $\Phi$-sat missions~\cite{giuffrida2021varphi}\add{ and
for in-orbit flood and water mapping with very small models~\cite{mateogarcia2023inorbit}}.\del{ We do not
claim a quantified industry-wide trend.} The hardware that local authorities and
relief organizations operate is not well documented and varies considerably, from
ruggedized laptops through embedded graphics processing units (GPUs) to custom
field-programmable gate arrays (FPGAs). What these platforms share is a strict memory budget.
A 300\ M-parameter Vision Transformer (ViT) at 32-bit floating point (FP32) is a poor fit for
memory-constrained embedded targets, whereas a compact 8-bit integer (INT8) model offers a more
accessible deployment path. Sixteen-bit floating point (FP16) or structured pruning remain alternatives
where more memory headroom is available. \add{Some low-power accelerators support only integer
arithmetic. A model that runs strictly in INT8 therefore offers the largest hardware compatibility.} The gap between the models that perform
best in research and those that can run in the field thus remains \chg{little}{hardly} studied for
geospatial \chg{flood segmentation}{water and flood mapping}.

Knowledge distillation (KD) is a natural route from a large foundation model to a compact student, but our setting adds a data constraint. Sen1Floods11 contains only 252 manually labeled training scenes. Training a compact model directly on that split fixes the amount and geographic scope of supervision, whereas a teacher that has already been adapted to the task can apply its supervision to additional unlabeled imagery. We therefore use Prithvi-EO-2.0 as a frozen task teacher, \chg{generate supervision on Sentinel-2 tiles}{use it to label Sentinel-2 images} retrieved around flood events from the Global Disaster Alert and Coordination System (GDACS), and train \chg{a compact EfficientViT-B0 student on those scenes}{compact students on those images, with EfficientViT-B0 as the main deployable student}. No additional manual annotation is acquired.

\begin{oldtext}
\revst{This design separates two questions that are often conflated. First, at the same scene budget, does teacher-supervised training provide a viable substitute for direct training on the original human-labeled split? Second, once a task teacher exists, does scaling the supervised image pool beyond the fixed annotation set narrow the student--teacher gap? The first question is tested at $N=252$ with matched architectures, objectives, augmentation and seeds. The second is tested by scaling the teacher-supervised pool to $N=2{,}500$ while keeping the teacher fixed. External evaluation on STURM-Flood and WorldFloods-v2 then tests whether the resulting student remains useful outside Sen1Floods11.}

\revst{We therefore restrict the interpretation in two ways. Synthetic gain, haze, and per-band
radiometric perturbations are treated as diagnostics rather than as a robustness contribution.
The two training sources differ in acquisition geometry as well as in imagery, and no experiment
isolates training-pool diversity as a cause. Because a fixed-threshold modified normalized difference water index (MNDWI) baseline is
competitive on both external benchmarks, we present external accuracy as evidence of
generalization rather than superiority of a learned model over a spectral rule.}
\end{oldtext}
\begin{newtext}
This design raises three questions that are often conflated. First, do teacher labels achieve human-level annotation quality on the same images, and what do the teacher's probability maps add? Second, does labeling more images with an existing task teacher improve performance, and does the benefit depend on the student? Third, how small can the student be, and can it run strictly in INT8 on an embedded device without losing accuracy? Section~\ref{sec:onechange} answers the first question, Sections~\ref{sec:onechange} and~\ref{sec:analysis} the second, and Sections~\ref{sec:analysis} and~\ref{sec:deployment} the third. We address these questions under one controlled protocol. Every model is trained and evaluated on $512\!\times\!512$ images. Each comparison changes one factor where possible, and each setting is trained with four seeds. We report differences with 95\% intervals obtained by resampling whole flood events. In addition to the Sen1Floods11 test split, STURM-Flood and WorldFloods-v2, we evaluate on 97 held-back WorldFloods events and on water-free ground.
\end{newtext}

The deployment question is \chg{independent of those secondary diagnostics}{treated separately}. The practical target is a student small enough to \chg{compile to TensorRT and run}{run} with low latency and memory use on embedded hardware while retaining most of the teacher's \chg{in-distribution flood-water accuracy}{surface-water IoU accuracy}. The resulting compression ratio is about $430\!\times$ in parameter count, from 300\,M parameters to 0.7\,M\add{, and about $24{,}500\!\times$ for a 12,230-parameter student}.

\begin{newtext}
Du et al.~\cite{du2025onorbit} compressed Prithvi-EO-2.0 about 16-fold into smaller transformers and ran one in orbit on the International Space Station in FP32, with task heads fine-tuned on labeled data. We reach about 430-fold and 24,500-fold compression across architectures, down to full INT8 and without new manual labels. Under one controlled protocol, we assess where the teacher's supervision adds value and how far it carries students of different architectures and sizes. We also measure the cost of each number format on an edge device. Unlike prior work, we evaluate the teacher and each student on four test sets from three public datasets and on two dry-scene sets introduced here, so that each claim is tested for transfer across datasets. To our knowledge, no earlier study of foundation-model distillation for water mapping has separated these factors, evaluated them this broadly or followed a student down to integer-only execution.
\end{newtext}

\ifrevmarked\paragraph*{\revst{Contributions}}\leavevmode\fi 
\begin{oldtext}
\begin{itemize}
  \item \revst{\textbf{Annotation-budget amplification through teacher supervision.} At the matched budget of 252 scenes, teacher-supervised training is a competitive substitute for direct training on the Sen1Floods11 labels and raises STURM-Flood mIoU in all five tested architecture--objective cells, four of which survive Holm correction over the block. A geometry-matched control shows that the size of the effect is partly entangled with acquisition geometry. The claim is therefore substitution at a fixed manual annotation budget, not superiority of teacher-generated labels over human annotations.}

  \item \revst{\textbf{Scaling beyond the fixed manual split.} Because additional teacher-supervised scenes require no new manual labeling, the student training set can grow beyond the 252-scene annotation budget. At $N=2{,}500$, the EfficientViT-B0 student narrows the in-distribution water intersection-over-union (IoU) gap to the teacher and matches its STURM-Flood mIoU under our evaluation protocol, while remaining lower on WorldFloods-v2. This establishes both the benefit and the limit of the scaling strategy, without claiming parity across out-of-distribution settings in general.}

  \item \revst{\textbf{Edge deployment of a highly compressed student.} ReLU6 retraining and quantization-aware training produce a 1.5\,MB INT8 TensorRT engine for the 0.7\,M-parameter student. On a Jetson Xavier NX it runs at 5.57\,ms of GPU compute per $512\!\times\!512$ image, with approximately 14\,MB of runtime device memory and a further 0.4\,ms of host--device transfer. The deployment contribution is the small, measured engine footprint and latency, not a claim that the 300\,M teacher is impossible to execute on every embedded platform.}
\end{itemize}
\end{oldtext}
\begin{newtext}
Therefore, this paper contributes the following novel methods:
\begin{enumerate}
  \item \textbf{A controlled account of what a foundation-model teacher contributes.} We separate the teacher's labels,
  its probability maps, the spatial extent of the training images and their number under one protocol. We use four seeds and
  intervals obtained by resampling whole flood events. Teacher labels achieve human-label quality for student training. When images are limited,
  the probability maps account for most of the teacher's value. The full pipeline brings student performance to the
  teacher's level on test sets from different places, dates and labeling teams.

  \item \textbf{Students of different architectures and sizes perform similarly to the teacher.} Six students, ranging from a 1,266-parameter network to
  a 32.5\,M-parameter U-Net, all achieve performance within 2 IoU points of the teacher on WorldFloods and STURM-Flood when the teacher
  supervises them. The teacher's benefit is therefore approximately equal to each student's shortfall when trained on hand labels alone. With the proposed setup, every
  design produces fewer false alarms on water-free ground. We further propose the smallest student model for deployment that stays within 2 IoU points of the teacher on all four test sets.

  \item \textbf{Compression measured and deployed on an edge device.} We trace accuracy from a 418-parameter per-pixel network to the
  300\,M-parameter teacher. We measure three student models on an edge device in four number formats. For the student with attention, we provide a
  quantization setup that prevents performance collapse under INT8: ReLU6 from the first update, quantization-aware training (QAT) with quantizers only on
  convolution inputs and weights, frozen batch normalization and FP16 attention. The main student runs
  as a 1.59\,MB mixed INT8/FP16 engine, and a 12,230-parameter student (12\,kB of weights), about $24{,}500\!\times$ smaller than the teacher, runs
  in full INT8, as integer-only accelerators require, with almost no loss.

  \item \textbf{Better model robustness for unattended use.} The main student model outperforms a fixed-threshold modified normalized difference water index (MNDWI) by 11.5 IoU points across 97 varied flood events and raises
  a false alarm on 5.4\% rather than 57.0\% of water-free windows, both of which matter for unattended use. On small curated test sets, the index remains competitive.

  \item \textbf{Strict evaluation across datasets.} We test the teacher and every student on four test
  sets from three public datasets: the Sen1Floods11 and WorldFloods test splits, STURM-Flood and 97 held-back WorldFloods
  events. We also introduce two dry-scene sets, 200 water-free
  WorldFloods windows and 49 dry tiles, because flood-centered test sets say little about false alarms on dry ground. Each claim is thus tested for
  transfer across datasets.
\end{enumerate}
These novel contributions are combined in a distillation pipeline with two deployable student models. The 0.7\,M EfficientViT-B0 uses a mixed INT8/FP16 engine for edge devices. The 12,230-parameter student uses full INT8 for integer-only hardware. The pipeline builds on established techniques (pseudo-labeling, farthest-point sampling, logit distillation and quantization), each credited to its source. The remaining errors, on shadows, snow and the small events of STURM-Flood, are shared with the teacher (Sections~\ref{sec:index} and~\ref{sec:inherit}), so further gains must come from the teacher and the training data, not from larger students.
\end{newtext}

\section{Related Work}
\label{sec:related}

\subsection{Geospatial Foundation Models}
Satellite foundation models such as SatMAE~\cite{cong2022satmae},
SpectralGPT~\cite{hong2024spectralgpt}, and Prithvi~\cite{szwarcman2025prithvi}
use large-scale self-supervised pretraining to transfer across geographies and tasks.
Prithvi-EO-2.0, our teacher, is pretrained globally on Harmonized Landsat Sentinel-2
imagery. Earlier flood studies show the value of this pretraining outside the fine-tuning
distribution: Li et al.~\cite{li2023assessment} report stronger geographic generalization
for the original Prithvi than conventional segmentation baselines, while Polushko
et al.~\cite{polushko2026igarss} study label efficiency when adapting Prithvi-EO-2.0 to
airborne RGB imagery. Garg et al.~\cite{garg2023cross} use a large pool of unlabeled
pairs to distill an optical flood teacher into a Sentinel-1 student. Their goal is
cross-modal label efficiency rather than compact edge deployment. \chgx{To our knowledge,
distilling a geospatial foundation model into an INT8-deployable flood-segmentation
student without acquiring additional ground-truth labels has not been studied.}{Du et al.~\cite{du2025onorbit}
distilled the pretrained Prithvi-EO-2.0 encoder into smaller transformers, fine-tuned task heads on labeled data and ran the small model in orbit.
Beyond this, we were able to distill the fine-tuned task model across architectures into students with as few as
1,266 parameters, train on teacher-labeled images and evaluate strict INT8 execution.}

\subsection{Knowledge Distillation and Compact Segmentation}
Knowledge distillation transfers a teacher's output distribution to a student
\cite{hinton2015distilling} and has since been extended to feature- and
attention-level objectives~\cite{romero2014fitnets,zagoruyko2016paying}; Himeur
et al.~\cite{himeur2025applications} survey its use in remote sensing.
Cross-architecture transfer is harder when teacher and student expose incompatible
intermediate representations~\cite{liu2022cross}. One-for-All (OFA)~\cite{hao2023one} avoids direct
feature matching by attaching lightweight student branches that predict the teacher's
final output at several depths, \chg{which makes it a natural fit for}{and was designed for} transfer from a ViT to a convolutional neural network (CNN).
\chg{The closest}{A related} foundation-model deployment pipeline is InstaGeo~\cite{yusuf2025instageo}:
it distills fine-tuned geospatial foundation models into smaller versions of the same
architecture for cloud serving. Our setting differs in the much larger cross-architecture
compression, training on teacher-labeled scenes rather than ground-truth scenes, and
quantized on-device evaluation.

UNet-ResNet50~\cite{ronneberger2015u} is a standard flood-segmentation baseline.
EfficientViT~\cite{cai2023efficientvit} combines local depthwise convolutions with
linear attention and provides a favorable accuracy--efficiency trade-off for dense
prediction. \add{MobileNetV3~\cite{howard2019searching} and LightFormer~\cite{chen2026lightformer}, a
lightweight decoder for remote-sensing segmentation, are further compact designs that we compare under the same
protocol. Compact water segmenters have also been obtained by distillation, for example WaterSegformer with relational
distillation~\cite{yang2023watersegformer} and a light network for agricultural water bodies~\cite{liao2025lightweight}.
Furthermore, a SAM2-based foundation model has been distilled into an EfficientNet-B0 student for urban scenes~\cite{shi2026sam2arafnet}.
Their spectral bands, labels and test data differ from ours, so we limit these comparisons to a qualitative discussion. Very small models have mapped floods on low-cost satellite
hardware~\cite{mateogarcia2019flood,mateogarcia2021towards} and have even been used in orbit~\cite{mateogarcia2023inorbit}.}
We use EfficientViT-B0 as the \chg{primary}{main} student because it is compact, remains
competitive in our OOD controls, and can be carried through the deployment path described
in Section~\ref{sec:qat}.

\subsection{Flood Benchmarks and Acquisition Robustness}
Sen1Floods11~\cite{bonafilia2020sen1floods11} is the in-distribution training benchmark.
STURM-Flood~\cite{notarangelo2025sturm} and WorldFloods-v2~\cite{portales2023global}
provide complementary Sentinel-2 OOD evaluations built from Copernicus emergency-mapping
activations; recent multi-modal datasets~\cite{montello2022mmflood,chen2026sen2gf3data}
address a broader sensor setting than the optical-only student studied here.
\delx{Robustness to acquisition conditions is distinct from clean-benchmark accuracy.
Sentinel-2 radiometric uncertainty is naturally expressed in relative terms
\cite{gorrono2017radiometric}, motivating our gain diagnostic, while atmospheric haze
is well modeled as an additive wavelength-dependent contribution
\cite{makarau2014haze}. We therefore test gain and haze directly in reflectance space
(Section~IV-F) rather than relying on generic natural-image corruptions.}\add{Synthetic gain and haze diagnostics
from an earlier, 224-pixel protocol of this study are given in Supplementary Section~\ref{S-sec:s_harmonisation}.}

\section{\chghead{Method}{Proposed Method}}
\label{sec:method}

\begin{delfloat}
\begin{figure*}[t]
\revdelfloat{FIG. 1}
\centering

\includegraphics[width=\linewidth]{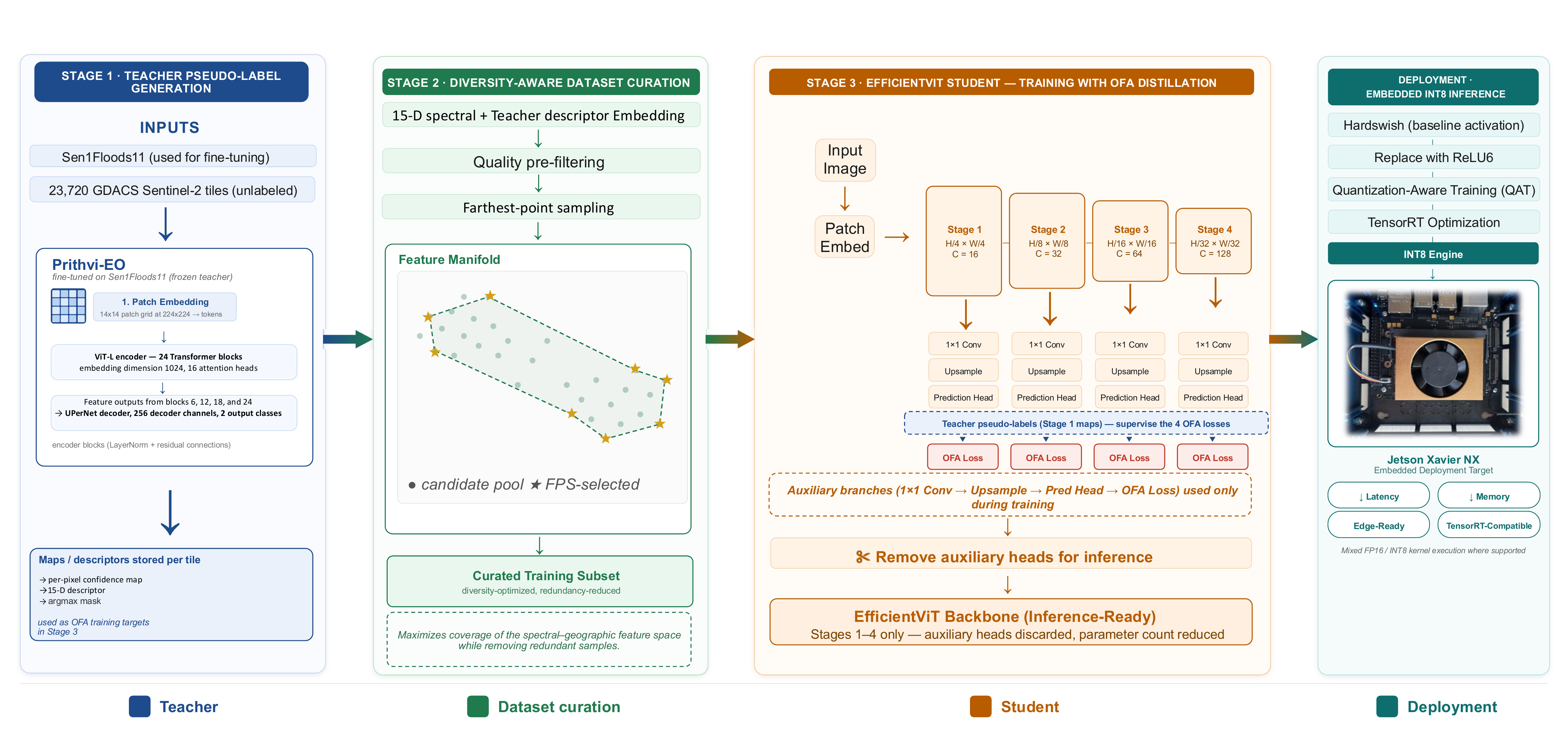}

{\footnotesize \revst{Submitted Fig.~1. Overview of the distillation-to-deployment pipeline: pseudo-labeling of the
unlabeled GDACS pool by the frozen foundation-model teacher, filtering and subset
selection, cross-architecture distillation into the compact student, and quantized
deployment. Only the hard mask, confidence map, and 15-dimensional descriptor are stored
during pool generation; soft targets are produced by live frozen-teacher inference during
student training. The pool holds two acquisitions per location, one inside the event window
and one pre-event baseline, which enables the pre-event teacher-semantics audit of
Section~IV-G.}\par}
\end{figure*}
\end{delfloat}

\begin{figure*}[t]
\revnewfloat
\centering
\includegraphics[width=\linewidth]{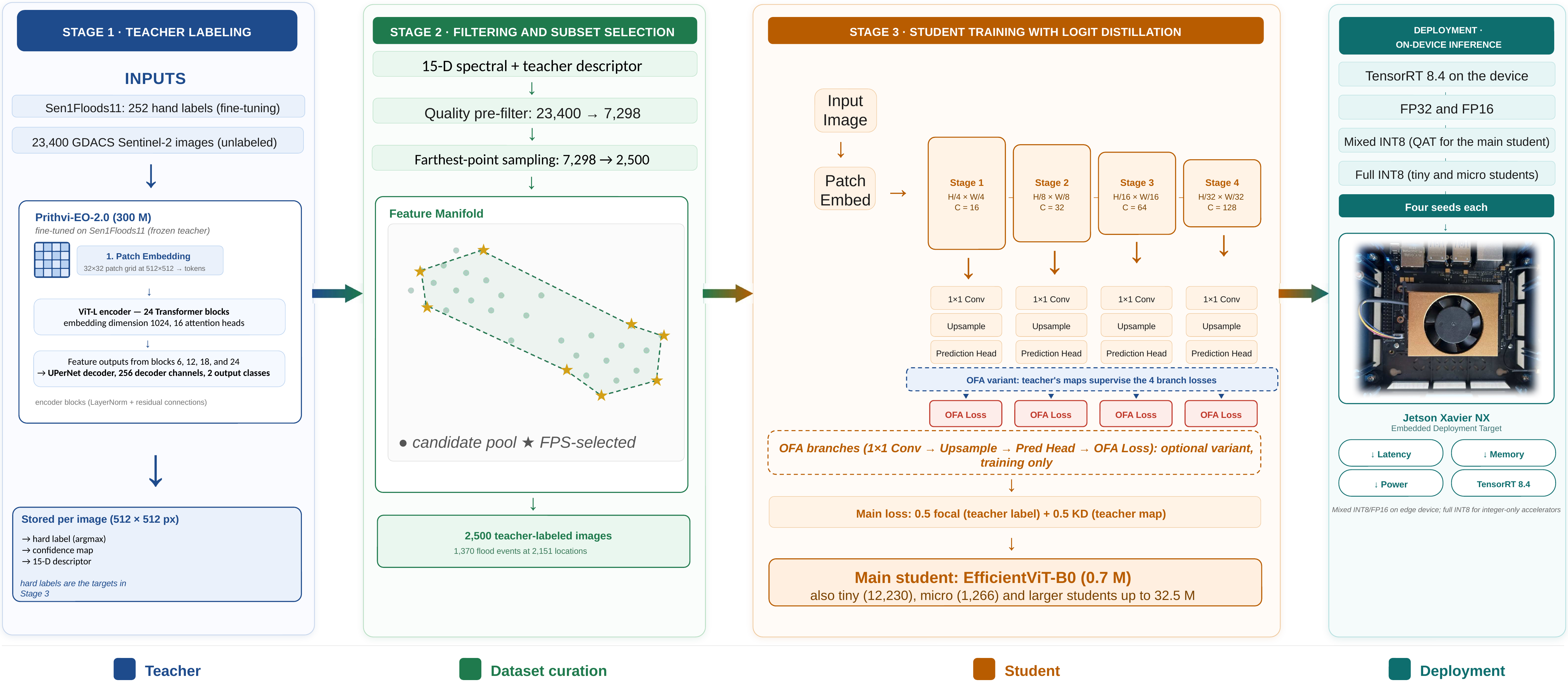}
\caption{\add{\textbf{Overview of the pipeline.} (1) The frozen teacher, fine-tuned on the 252 hand-labeled Sen1Floods11 images,
labels 23,400 Sentinel-2 images downloaded around GDACS flood alerts and stores a hard label, a confidence map and a
15-dimensional descriptor for each. (2) A quality filter and farthest-point sampling select 2,500 of them. (3) Students
learn from the teacher's labels and probability maps. The frozen teacher computes the probability maps for each augmented
image during training. One-for-All side branches are an optional training-only variant. (4) The students are compiled and
measured on a Jetson Xavier NX in four number formats. Checkpoints, learning rates and 8-bit calibration
use only the Sen1Floods11 validation split. The pool holds two acquisitions per location, one inside the event window
and one pre-event baseline, which enables the teacher-semantics audit of Section~\ref{sec:inherit}.}}
\label{fig:overview}
\end{figure*}

Figure~\ref{fig:overview} summarizes the pipeline. Its four stages are \chg{pseudo-labeling}{teacher labeling}, filtering and subset selection, distillation, and \chg{quantization}{on-device inference}.
They are described in turn below.

\subsection{Foundation Model Teacher and \chghead{Pseudo-label}{Teacher-Label} Generation}
\label{sec:teacher}

We use Prithvi-EO-2.0~\cite{szwarcman2025prithvi} fine-tuned on the Sentinel-2 subset of
Sen1Floods11 (S11) as our teacher. We take the official IBM/NASA
Prithvi-EO-2.0-300M-TL-Sen1Floods11 segmentation checkpoint released on Hugging Face,
without any further fine-tuning of our own, so the teacher is the
publicly available model.
On the S11 test split we measure \chgx{0.897~mIoU and 0.822
water IoU against the 0.900 and 0.826}{89.5~mIoU and 81.7 water IoU over all pixels against the 90.0 and 82.6} reported for this
checkpoint~\cite{szwarcman2025prithvi}; the remaining \chg{0.003--0.004 is a protocol difference}{difference of 0.5 to 0.9 points comes from the evaluation protocol}, not
a different model. \add{We report IoU in points from 0 to 100.} At inference it processes a single $512\!\times\!512$ six-band
Sentinel-2 image (bands B2, B3, B4, B8A, B11, B12) and produces per-pixel class
probabilities $p_T(x) \in [0,1]^{H\times W}$ for binary \chg{flood}{water} segmentation.

\paragraph*{Choice of teacher}
The labeler is applied only to scenes no one has annotated, so the property that matters
is its accuracy outside the fine-tuning distribution. We select Prithvi-EO-2.0 because it provides strong external performance after task adaptation. This study does not test whether foundation-model pretraining itself is required for the observed transfer. This is a design decision
supported by measured out-of-distribution accuracy, not a demonstration that a foundation
model is required: we did not compare Prithvi with a second learned teacher of comparable
capacity and provenance (Section~\ref{sec:limitations}). Section~\ref{sec:inherit} measures which of the teacher's
annotation conventions the student reproduces, and Supplementary Sections~\chg{S4 and~S8}{\ref{S-sec:s_mndwi} and~\ref{S-app:diag}} contain
additional labeler diagnostics.

\begin{oldtext}
\paragraph*{\revst{What the pseudo-labels contain}}
\revst{Supervision enters on two axes, the labeling of the GDACS pool and the objective.
Adding the temperature-scaled soft term to the hard-target loss increases STURM mIoU by
$0.018$ at a matched budget of 252 scenes (Table~I). This difference is
consistent with the soft teacher probabilities carrying useful spatial information beyond the
hard argmax. The two rows also differ in temperature and in the weight on the hard target
(Section~III-C), so the contribution of the soft target alone is not isolated. OFA adds a
smaller further gain; the gradient diagnostics in Section~V are consistent with
deep supervision contributing to this effect.}
\end{oldtext}
\begin{newtext}
\paragraph*{What the teacher provides}
The teacher supplies two forms of supervision for each image: a hard label (its argmax) and a probability map. We call
the hard labels \emph{teacher labels} (pseudo-labels in the literature) and the human masks of Sen1Floods11 \emph{hand
labels}. On the same images, teacher labels achieve human-level annotation quality for student training. When images are limited, the probability maps provide most of the
teacher's additional value (Section~\ref{sec:onechange}).
\end{newtext}

The teacher is used strictly for inference from here on; nothing in this paper trains
or fine-tunes it on the unlabeled pool. We run it once, frozen, over all \chg{23{,}720
GDACS tiles at the $224\!\times\!224$ pool-labeling resolution}{23{,}400 GDACS images at their native $512\!\times\!512$ resolution} (Sentinel-2 imagery downloaded via Google Earth Engine (GEE) for GDACS flood events; see
Section~\ref{sec:datasets}) and store per \chg{tile}{image}: the hard argmax \chg{pseudo-mask}{label}, a per-pixel confidence map, and a
15-dimensional summary vector (flood fraction, mean confidence, confidence standard
deviation, and per-band mean and std across the 6 input channels) used for curation.

\subsection{\chghead{Pseudo-label Filtering}{Filtering} and Subset Selection}
\label{sec:curation}

The GDACS pool $\mathcal{U}$ (Section~\ref{sec:datasets}) is spatially and temporally
correlated: large flood events generate many overlapping acquisition strips, so the most
frequent event types and geographies are heavily over-represented, and half of it is a
pre-event acquisition that the teacher labels no differently from the rest. A selection
stage therefore has two plausible jobs, removing tiles whose labels are unusable and
spreading the remainder over the feature space. We designed the stage for both; only the
first is measurable at our operating point (Supplementary Section~\chg{S2}{\ref{S-sec:s_curation}}).

After pre-filtering and selection the curated pool is capped at $N\!=\!2{,}500$
scenes drawn from the 
\chg{5{,}987}{7{,}298} that survive the filter, and this is the operating point for
all full-pool experiments.

Our curation uses three steps.

\paragraph*{Step 1: Pre-filtering}
We first restrict $\mathcal{U}$ to tiles that contain informative boundary signal.
Let $\hat{y}_T(u) = \arg\max_c\,p_T^{(c)}(u)$ be the teacher's hard class prediction
at pixel $u$ (\chg{flood or non-flood}{water or land}), and $p_T(u) \in [0,1]$ the teacher's \chg{flood}{water}
probability at pixel $u$.
We define:
\begin{itemize}
\item $f(x) = \frac{1}{|\Omega|}\sum_{u\in\Omega}\mathbf{1}[\hat{y}_T(u)=\text{water}]$:
  the fraction of pixels in tile $x$ that the teacher predicts as \chg{flood}{water}.
\item $\mu_T(x) = \frac{1}{|\Omega|}\sum_{u\in\Omega} p_T(u)$:
  the spatial mean of the teacher's \chg{flood}{water} probability across the tile.
\item $\sigma_T(x) = \sqrt{\frac{1}{|\Omega|}\sum_{u\in\Omega}(p_T(u)-\mu_T(x))^2}$:
  the spatial standard deviation of the teacher's confidence map, which measures
  \chg{how much uncertainty is present within the tile}{how much the teacher's water probability varies within the tile, rather than
  calibrated uncertainty}.
\end{itemize}
We then define:
\begin{equation}
  \mathcal{U}' = \bigl\{x \in \mathcal{U}
    \;\big|\;
    f_{\min} \le f(x) \le f_{\max}
    \;\land\;
    \sigma_T(x) \ge \sigma_{\min}
  \bigr\},
  \label{eq:prefilter}
\end{equation}
with $f_{\min}\!=\!0.05$, $f_{\max}\!=\!0.85$, $\sigma_{\min}\!=\!0.05$, fixed a priori
and never tuned against a benchmark.
The lower bound on $f$ removes background-only tiles, the upper bound removes tiles
that are almost entirely flooded with no land boundary, and the $\sigma_T$ threshold
removes tiles where the teacher is uniformly confident, which are scenes that provide no
gradient information about the flood boundary. This reduces \chg{23{,}720 to 5{,}987}{23{,}400 to 7{,}298}
candidates.

The pre-filter \del{is the curation component that measurably matters, and it acts in both
directions. It also} has two side effects that follow from the rule rather than from our data.
\chgx{The lower bound $f\!\ge\!f_{\min}$ admits 2{,}127 of the 11{,}825 pre-event baselines
(35.5\% of the candidate pool), which are by construction the pre-event tiles the teacher
marked as water}{The lower bound $f\!\ge\!f_{\min}$ admits pre-event baselines that the teacher marked as water:
831 of the 2,500 training images are pre-event acquisitions} (Section~\ref{sec:inherit}); and it guarantees that no training \chg{tile}{image} is
\chg{flood-free}{free of teacher-predicted water}, so the student never sees a fully dry scene (Section~\ref{sec:index}).\del{ Neither
was intended.}

\paragraph*{Step 2: Feature embedding}
Each remaining tile is embedded into a compact descriptor built entirely from
quantities already stored during \chg{pseudo-label}{teacher-label} generation; no additional teacher
inference is required:
\begin{equation}
\begin{aligned}
  \boldsymbol{\phi}(x) = \bigl[\,
      & f(x),\; \mu_T(x),\; \sigma_T(x),\; \\
      & \bar{x}_1, \ldots, \bar{x}_K,\;
        \sigma_{x_1}, \ldots, \sigma_{x_K}
    \,\bigr] \;\in\; \mathbb{R}^{2K+3},
  \end{aligned}
  \label{eq:embedding}
\end{equation}
where $K\!=\!6$ is the number of spectral bands, $\bar{x}_k$ and
$\sigma_{x_k}$ are the mean and standard deviation of band $k$ over the tile.
This gives a 15-dimensional vector ($2\times6 + 3$). The first three dimensions,
$f(x)$, $\mu_T(x)$, and $\sigma_T(x)$, are the same quantities used in
Step~1 pre-filtering and capture \chg{flood}{water} content and label uncertainty. The remaining
twelve dimensions are the per-band mean and standard deviation of the raw Sentinel-2
reflectance, which encode spectral appearance and, because spectral statistics vary
systematically across geographic regions and seasons, serve as a proxy for
geographic and environmental diversity. A flood over dry, bare ground reflects light
very differently from a flood in a dense green landscape, so the two scenes have
different per-band statistics even when the same fraction of each is under
water. By including both the \chg{pseudo-label}{teacher-label} statistics and the raw
spectral statistics in $\boldsymbol{\phi}$, the farthest-point sampling (FPS) step diversifies the selected
set along both the ``what the teacher predicts'' and ``what the scene looks like''
axes simultaneously.

\paragraph*{Step 3: Farthest-point sampling}
Given the embedded pool $\{\boldsymbol{\phi}(x) : x \in \mathcal{U}'\}$, we
select a diversity-maximising subset $\mathcal{D}^* \subset \mathcal{U}'$ of size
$N$ by greedy FPS. Starting from an arbitrary seed
$d_0 \in \mathcal{U}'$, we iteratively add the point that maximizes the minimum
distance to all already-selected points:
\begin{equation}
  d_{t+1} = \operatorname*{arg\,max}_{x\,\in\,\mathcal{U}' \setminus \{d_0,\ldots,d_t\}}
    \;\min_{0 \le j \le t}\,
    \bigl\|\boldsymbol{\phi}(x) - \boldsymbol{\phi}(d_j)\bigr\|_2.
  \label{eq:fps}
\end{equation}
Greedy FPS repeatedly adds the candidate farthest from its nearest selected
neighbor, giving the standard $k$-center coverage approximation in the
15-dimensional descriptor space~\cite{sener2017active}. This is a geometric coverage
property, not a distributional guarantee.

FPS is expected to outperform confidence-ranked selection, which concentrates the pool in the
region of $\boldsymbol{\phi}$-space with high $\mu_T$ and low $\sigma_T$. We tested this
against alternative selectors and random controls \add{under an earlier, 224-pixel protocol} (Supplementary Section~\chg{S2}{\ref{S-sec:s_curation}}). It does not hold
at our operating point: once the pre-filter has run, differences between ranking rules are
within the seed spread. It does hold \emph{without} the pre-filter, where confidence ranking
degrades as the active-learning literature predicts. We therefore report FPS as the selector we
deploy rather than as a component whose contribution we can measure.

\subsection{Distillation \chghead{Objective: OFA}{Objectives}}
\label{sec:ofa}

\begin{oldtext}
\revst{Given pseudo-label set $\mathcal{D} = \{(x_i, z_i)\}$ with $z_i = T(x_i)$, we
train EfficientViT-B0 ($S_\theta$) using the OFA objective~\cite{hao2023one}. Lightweight
branch decoder heads $\{B_{\phi_i}\}_{i=1}^{L}$ are attached to the student's
intermediate feature maps $F_i^\theta(x)$ at each of the $L$ encoder stages.
Branch $i$ produces a prediction $b_i(x) = B_{\phi_i}(F_i^\theta(x))$ targeting
the teacher's final output. The full training loss is:}
\begin{equation*}
  \mathcal{L}_\text{OFA}(\theta,\phi)
  \;=\;
  \mathcal{L}_\text{head}(\theta)
  \;+\;
  \alpha_\text{OFA}
  \sum_{i=1}^{L}
  \mathbb{E}_{x\sim\mathcal{D}}\!\left[\,
    \ell_\text{br}\!\left(b_i(x),\;p_T(x)\right)
  \right]
  \tag*{(old 4)}
\end{equation*}
\revst{where the main-head loss is the fixed mixture}
\end{oldtext}
\begin{newtext}
Given the teacher-labeled set $\mathcal{D} = \{(x_i, z_i)\}$ with $z_i = T(x_i)$, we
train the student $S_\theta$ with logit distillation~\cite{hinton2015distilling}. Its loss is the fixed mixture
\end{newtext}
\begin{equation}
  \mathcal{L}_\text{head}(\theta)
  \;=\;
  (1{-}\alpha)\,\ell_\text{focal}\!\left(S_\theta(x),\tilde{y}\right)
  \;+\;
  \alpha\,\ell_\text{KD},
  \qquad \alpha=0.5,
  \label{eq:headloss}
\end{equation}
of a hard-label focal term~\cite{lin2017focal} and the temperature-scaled logit distillation
term\add{ ($T\!=\!4$, scaled by $T^2$)}. The hard target $\tilde{y}$ is the teacher's argmax
$\hat{y}_T(x) = \arg\max_c p_T(x)$ in the \chg{pseudo-label}{teacher-label} configurations and the human
annotation in the \chg{Sen1Floods11 ones}{hand-label ones}\chg{ (\emph{Supervision in the Sen1Floods11 experiments},
below), $L\!=\!4$ is the number of encoder stages, and the branch
parameters $\phi$ are discarded at inference.}{. The teacher probabilities come from a live forward pass of the frozen teacher on the
same augmented image the student sees. Eq.~\eqref{eq:headloss} is the objective of the proposed setup.}

\begin{newtext}
\paragraph*{One-for-All variant}
OFA~\cite{hao2023one} attaches lightweight branch heads $\{B_{\phi_i}\}_{i=1}^{L}$ to the student's
intermediate feature maps $F_i^\theta(x)$ at each of the $L\!=\!4$ encoder stages. Branch $i$ produces a prediction
$b_i(x) = B_{\phi_i}(F_i^\theta(x))$ targeting the teacher's final output, and the training loss becomes
\begin{equation}
  \mathcal{L}_\text{OFA}(\theta,\phi)
  =
  \mathcal{L}_\text{head}(\theta)
  +
  \frac{\alpha_\text{OFA}}{L}
  \sum_{i=1}^{L}
  \mathbb{E}_{x\sim\mathcal{D}}\!\left[\,
    \ell_\text{br}\!\left(b_i(x),\;p_T(x)\right)
  \right],
  \label{eq:ofa}
\end{equation}
with $\alpha_\text{OFA}\!=\!0.3$. The branch losses are averaged, not summed. The branch parameters $\phi$ are
discarded at inference.
\end{newtext}
The branch loss $\ell_\text{br}$ is
the OFA loss of Hao et al.~\cite{hao2023one} (their Eq.~6), applied per pixel:
\begin{equation}
  \ell_\text{br}(b_i, p_T)
  \;=\;
  -\bigl(1+p_T^{\tilde{y}}\bigr)^{\gamma}\,\log b_i^{\tilde{y}}
  \;-\;
  \sum_{c \neq \tilde{y}} p_T^{c}\,\log b_i^{c},
  \label{eq:branchloss}
\end{equation}
where $p_T^{c}$ is the temperature-scaled teacher probability of class $c$ and
$b_i^{c}$ the branch softmax output at that pixel. The modulation exponent
$\gamma\!=\!1.2$ raises the weight of the hard-target class where the teacher
is confident in it. \chgx{Setting $\gamma\!=\!0$ removes the modulation and leaves standard
knowledge-distillation cross-entropy on the branch; $\gamma\!=\!1$ retains a factor
$(1{+}p_T^{\tilde y}) \in [1,2]$ on the target term.}{At $\gamma\!=\!1$ the branch loss equals hard
cross-entropy plus soft cross-entropy. Setting $\gamma\!=\!0$ does not reduce it to soft distillation, because a
term $-(1{-}p_T^{\tilde y})\log b_i^{\tilde y}$ remains.} \chg{Equation~\ref{eq:headloss} is}{With cross-entropy in place
of the focal term, Equation~\ref{eq:headloss} becomes} the
KD${+}$cross-entropy (CE) mixture analyzed in Supplementary Section~\chg{S9}{\ref{S-sec:ceiling}}. 
\chgx{The same head loss without the distillation term (focal only) is the ``CE''
baseline of the tables; with the distillation term but without branch heads it is the
Logit-KD baseline.}{The focal loss on the hard label alone, without the distillation term, is the
\emph{labels-only} variant. Plain cross-entropy is used for the reference students that learn from the hand labels
without any teacher, and as the hard-label loss of the larger students in Section~\ref{sec:analysis}.} The \add{deployed} model is therefore
identical in architecture to a vanilla \chg{EfficientViT-B0}{student of the same design}.

OFA was designed for heterogeneous teacher-student pairs~\cite{hao2023one}, which is
our case (ViT\,$\to$\,CNN). \delx{Beyond the architecture argument, we show in
Section~V that OFA is a natural fit for pseudo-label training
specifically: every term in Eq.~1 targets the teacher's output and
introduces no competing supervisory signal. Its gain over standard logit KD comes
from deep-supervision gradient conditioning~\cite{lee2015deeply} rather than from a
different optimal student, a distinction that has practical consequences for how to interpret and
extend the results.}\add{Under our controlled protocol it gives no reliable gain over logit distillation for any of three
student designs (Section~\ref{sec:onechange}), so we report it as a variant.}

\chgx{Note that no term in Eq.~\ref{eq:ofa} matches intermediate \emph{features}, which is
deliberate.}{No term in Eqs.~\eqref{eq:headloss}--\eqref{eq:branchloss} matches intermediate features of the teacher.}
\begin{deltext}
\revst{The reason is empirical: adding a mean-squared-error matching term between
intermediate student and teacher features did not improve any benchmark in our experiments.
A representation-alignment measurement in Supplementary Section~S8 is consistent with that
null, showing that the student trained on the full pseudo-label pool already reaches high
deep-stage alignment with the teacher through logit-level supervision alone. That
measurement is a similarity diagnostic rather than a demonstration that feature matching
cannot help, and the alignment it reports depends on both the label source and the
objective.}
\end{deltext}

\begin{oldtext}
\paragraph*{\revst{Supervision in the Sen1Floods11 experiments}}
\revst{Both label sources are run through the same objectives, so ``Sen1Floods11 Logit-KD'' and
``Sen1Floods11 OFA'' are distillation configurations too, not teacher-free baselines. They
load the identical frozen teacher, and the soft term of Eq.~4 and the
non-target term of Eq.~6 are that teacher's output on the
\emph{Sen1Floods11} images. What the label source switches is the hard target $\tilde{y}$:
in the pseudo-label configurations it is the teacher's argmax $\hat{y}_T(x)$, and in the
Sen1Floods11 configurations it is the human annotation, with Eqs.~4
and~6 otherwise unchanged. The two label sources therefore share the
teacher, the objective, the architecture, the recipe and the same 252-scene annotation
provenance, and differ in which images are trained on and in the source of the hard target. The
two pools are not matched in native chip extent; the geometry-matched control is reported in
Section~IV-C.
The only teacher-free configuration in this paper is \emph{Sen1Floods11}\,CE.}
\end{oldtext}
\begin{newtext}
\paragraph*{Hand labels and teacher labels}
Both label sources use the same objectives. In the teacher-label configurations, the hard target $\tilde{y}$ is
the teacher's argmax on a downloaded image. In the hand-label configurations, it is the human annotation of a
Sen1Floods11 image, and the probability maps are the teacher's output on that image. A third configuration uses the teacher's
labels on the Sen1Floods11 images and changes only the label source (Section~\ref{sec:onechange}). The GDACS images are downloaded
at the Sen1Floods11 size, so all configurations share the same image extent. The only teacher-free
configurations are those trained on the hand labels alone.
\end{newtext}

\begin{deltext}
\paragraph*{\revst{Use of the Sen1Floods11 labels}}
\revst{The Sen1Floods11 labels are available, so a reader might ask why the student does not
use them as well. Mixing the ground-truth scenes into the pseudo-label
pool degrades performance at the full pool,
and helps only when the pseudo pool is very small, where a same-size pseudo pool
does better anyway. A two-stage schedule is cleaner to interpret and gives the same
answer: initializing from a pseudo-label-trained checkpoint and fine-tuning on the
ground truth costs $0.017$~mIoU on STURM on all four seeds and multiplies the seed
spread by six, while the reverse order (ground truth first, then pseudo-labels)
reproduces the pseudo-only result to within $0.003$ on every benchmark. The 2{,}500
tiles of second-stage training simply overwrite the initialization.}

\revst{One possible explanation is a mismatch in annotation conventions, rather than a lack of information.
Boundaries are subjective at the pixel level, and human annotations
frequently assign a hard label where the teacher's probability is 0.3--0.7; training on
both at once asks the student to fit two conventions on the same object. The consequence is
that there is no measured benefit to using the Sen1Floods11 labels directly, and on STURM-Flood a measured penalty, so the pseudo-label-only pipeline is not a compromise
forced by their absence.}
\end{deltext}

\paragraph*{Adaptation to dense prediction}
The original OFA paper~\cite{hao2023one} targets image \emph{classification}: branch
heads are implemented as global average pooling followed by a linear layer,
producing a single class-score vector per image. Flood segmentation is a dense
prediction task: each branch head must produce a $C\!\times\!H\!\times\!W$ output
at the same spatial resolution as the teacher's mask. We implement each branch
$B_{\phi_i}$ as a lightweight decoder comprising a \chgx{$1\!\times\!1$ convolutional
projection}{$1\!\times\!1$ convolution that halves the channels, batch normalization, a rectified linear unit and a
$1\!\times\!1$ output convolution,} followed by bilinear upsampling to the student's main-output resolution. In the
main experiments, branch outputs, student logits, and live teacher targets are evaluated at
the \chg{$224\!\times\!224$}{$512\!\times\!512$} training resolution. The branch loss $\ell_\text{br}$ of
Eq.~\ref{eq:branchloss} is then computed per pixel and averaged over the tile.

EfficientViT-B0 has four encoder stages with channel widths $C_0\!=\!16$,
$C_1\!=\!32$, $C_2\!=\!64$, $C_3\!=\!128$. \chgx{Each branch head $B_{\phi_i}$ projects
from $C_i$ to 2 classes (flood / non-flood) via a single $1\!\times\!1$ convolution
with bias: the per-branch parameter counts are 34, 66, 130, and 258 respectively,
summing to 488 parameters in total, which is less than 0.07\% of the 0.7\,M student
budget.}{The four branch heads have 11{,}368 parameters in total, less than 2\% of the 0.7\,M student.} This overhead is \chg{negligible}{small} during training and absent at inference; the
deployed model is architecturally identical to a vanilla EfficientViT-B0.

\begin{newtext}
\subsection{Students}
\label{sec:students}
The main student model is EfficientViT-B0~\cite{cai2023efficientvit} (0.7\,M parameters), initialized from a segmentation
model pretrained on Cityscapes street scenes. It retains its native Hardswish activations. To compare architectures, we
train MobileNetV3-Small~\cite{howard2019searching} (3.55\,M), LightFormer-ResNet18~\cite{chen2026lightformer} (11.43\,M,
using the authors' code) and UNet-ResNet50~\cite{ronneberger2015u} (32.5\,M), a common flood-segmentation baseline. All three use
ImageNet-pretrained encoders. We also train four small students from scratch: a per-pixel network with two hidden layers
(418 parameters), a two-layer CNN with a 5-pixel receptive field (1,058), and two convolutional encoder--decoders without
attention. The encoder--decoders use depthwise-separable residual blocks in three stages at 1/4 to 1/16 resolution. The \emph{tiny}
student has 12,230 parameters. The \emph{micro} student uses the same design with fewer channels and has 1,266 parameters. We propose EfficientViT-B0 due to its balance between parameters and accuracy (IoU and precision).
\end{newtext}

\subsection{Activation Replacement and Quantization-Aware Training (QAT)}
\label{sec:qat}

\add{For deployment, we utilize NVIDIA TensorRT~8.4, which compiles a trained network into an optimized engine.}

\paragraph*{Activation replacement}
The standard EfficientViT-B0 uses Hardswish activations throughout the backbone.
Under TensorRT 8.4's per-tensor INT8 quantization, Hardswish \chg{was not viable in our
deployment: the activation's piecewise shape produces strongly varying dynamic ranges
in adjacent layers, and per-tensor scale calibration degenerates to accuracy collapse
after quantization}{is costly in our deployment because post-training quantization (PTQ) fails the accuracy check, and quantization-aware
training requires additional rescaling of the convolution inputs. The resulting mixed-precision engine is slower than FP32
on the device (Section~\ref{sec:jetson})}. We therefore train a separate \chg{ReLU6-from-scratch model}{ReLU6 model, replacing Hardswish from the first training update,} on the same
\chg{pseudo-label set}{2,500 teacher-labeled images, with the same training setup and initialization}. \chg{The activation
replacement effects are reported in Section~VI.}{In FP32 on the edge device it scores 0.6 to 2.6 IoU points below the Hardswish student on the three public test
sets, with every interval including zero (Table~\ref{tab:device}, Supplementary Table~\ref{S-tab:s_device}).}

Direct fine-tuning from the Hardswish checkpoint (swapping activations
then continuing training) does not work: the Hardswish pre-trained BatchNorm
running statistics are computed for a different activation distribution, and they
do not settle within a reasonable fine-tuning budget. Training ReLU6 \chg{from scratch}{from the first update}
is required.
\begin{newtext}
\paragraph*{Number formats}
For deployment, we compare four number formats in TensorRT~8.4: FP32, FP16, mixed INT8 (INT8 where the
builder supports it, floating point elsewhere) and full INT8. In full INT8, every tensor between layers is INT8.
EfficientViT-B0 cannot run in full INT8 with this toolchain. Its linear-attention step multiplies two computed
tensors, for which TensorRT~8.4 has no INT8 kernel. Per-tensor INT8 also collapses the attention accumulator. The deployed ReLU6 model
therefore retains attention in FP16, producing a mixed INT8/FP16 engine with 56 of its 57 convolutions in INT8. The
tiny and micro students contain no attention and are built for full INT8. They retain their native activations.
\end{newtext}

\paragraph*{Quantization-aware training}
We fine-tune the ReLU6 model with 8-bit weights and activations (W8A8) quantization-aware training\add{ (QAT) for ten
epochs on the 2,500 teacher-labeled images, keeping the epoch that scores best on the Sen1Floods11 validation images}. Quantize--dequantize (QDQ) nodes are placed
on convolution inputs and weights, but not on convolution outputs; keeping the final logits
in floating point avoids a background-dominated output scale that otherwise collapses the
\chg{flood}{water} range. Batch normalization (BN) affine parameters and running statistics are frozen during QAT.
Without that constraint, folded biases in a few channels inflate the per-tensor activation
scale by up to $9\times$ and crush the remaining channels toward zero. \add{The tiny and micro students use post-training quantization
for 13 of their 16 INT8 engines. Three engines required QAT or corrected
quantizer placement (Supplementary Section~\ref{S-sec:s_deploy}). Calibration and acceptance use only the Sen1Floods11
validation images: an INT8 engine is accepted when it stays within 2 IoU points of its own FP32 engine there, a common margin
for 8-bit quantization~\cite{krishnamoorthi2018quantizing}.}

\paragraph*{Export and compilation}
The QAT model is exported to the Open Neural Network Exchange (ONNX) format with
\texttt{torch.onnx.export}, simplified, and fixed to a static
$1\!\times\!6\!\times\!512\!\times\!512$ input for TensorRT~8.4. LiteMLA MatMul
operations remain in FP16 because per-tensor INT8 collapses their attention accumulator;
the deployed engine is therefore a mixed INT8/FP16 graph.\add{ TensorRT~8.4 does not support cubic interpolation. The two cubic upsampling steps of
the EfficientViT-B0 decoder therefore use linear interpolation on the device. All reported device accuracies include this change.} Compilation is performed on the
Jetson Xavier NX itself (JetPack~5.0, TensorRT~8.4), and the resulting engine is tied to
that target architecture. Supplementary Section~\chg{S7}{\ref{S-sec:s_deploy}} gives the export-graph rewrite and the
numerical precision diagnostics.

\section{\chghead{Experiments}{Experimental Setup}}
\label{sec:experiments}

\subsection{Datasets and Evaluation}
\label{sec:datasets}

\textbf{Sen1Floods11 (S11)}~\cite{bonafilia2020sen1floods11}: 446 paired Sentinel-1 and
Sentinel-2 images across 11 flood events in 11 countries, with pixel-level binary \chg{flood}{water}
labels. We use the Sentinel-2 six-band (B2, B3, B4, B8A, B11, B12) subset to match the
Prithvi-EO-2.0 input specification, and the official split of 252 / 89 / 90
hand-annotated chips with the 15 Bolivia chips reserved as a transfer set. \delx{Our
matched-budget series is capped at $N\!=\!252$, the full training list. Performance is reported as mean intersection
over union (mIoU) averaged over flood and non-flood classes.}\add{The 90 test chips come from 10 flood events.}

One property of these labels bears on the interpretation of every result below. Sen1Floods11 is
a \emph{surface-water} dataset: its analysts were asked to mark water areas inside a
flood scene, starting from a Sentinel-2 water classification, so permanent water is part
of the positive class rather than something subtracted from
it~\cite{bonafilia2020sen1floods11}, and the Prithvi-EO-2.0 authors describe it the same
way and score it as a water-class IoU~\cite{szwarcman2025prithvi}. A model trained on this target is therefore a water segmenter, not a flood-increment detector,
and so is any label it writes. We claim no flood-versus-permanent-water discrimination anywhere.
Accordingly we report the positive class as \emph{water IoU} throughout\chg{, including on
the two emergency-mapping benchmarks, where the ignore
protocol of the following paragraphs removes reference water before scoring}{ and score all surface water as positive on
every test set, including the two emergency-mapping benchmarks. The task is surface-water segmentation, with flood mapping as
its application}.

\textbf{STURM-Flood}~\cite{notarangelo2025sturm}: OOD benchmark from the Copernicus Emergency Management Service (EMS)
activations disjoint from S11, acquired by the same Sentinel-2 sensor; we select the six
S11-matching bands from the nine it stores (Appendix~\ref{app:impl}). Our evaluation uses
\add{all} 2{,}675 chips from 46 flood maps spanning 29 EMS activations\add{. Its authors randomly split the chips 80:10:10 across all
events~\cite{notarangelo2025sturm}. Their 267 test chips therefore share events with their training chips, and the release does
not identify the split. No model in this study is trained on STURM chips, so we use all chips as test data and group them
by event}. The release's
often-quoted 60 events is the Sentinel-1\,$\cup$\,Sentinel-2 union, of which our
optical-only evaluation touches 29. The grouping hierarchy chip $\to$ flood map $\to$
event is what the per-group statistics of \chg{Sections IV-C and IV-F}{Section~\ref{sec:stats}} use.
\chgx{STURM labels five water classes; we score flooded (1) and river-related (2) as flood and
set open water, reservoirs and lakes (3--5) to \emph{ignore} ($\approx$3.3\% of pixels).
\emph{This is not STURM's own protocol}, which merges classes 1--5 into one water class
over a 267-chip test split, so our STURM values are not comparable to any published STURM
baseline.}{STURM labels five water classes. We score all five as water, as in STURM's own protocol, but evaluate all
2,675 chips rather than the authors' random 267-chip subset. Scores are computed per flood event.} Because our \del{class mapping and} evaluation protocol \chg{differ}{differs} from the published STURM protocol, we use STURM as a controlled
OOD benchmark and do not compare its absolute scores with published baselines.
Appendix~\ref{app:impl} gives evidence that STURM also carries a processing-level shift
relative to the Level-1C products used everywhere else here, and substitutes B8 for B8A\add{. Its authors also sharpen
the 20\,m bands to 10\,m~\cite{notarangelo2025sturm}}.
Both affect every model identically, which is why we use this benchmark only for paired
comparisons in which the two experiments share architecture, objective, budget and augmentation:
such a comparison is insensitive to a shift common to both experiments, whereas using an absolute score
is not. Where we do report an  STURM number, as in the student--teacher comparison
of Section~\ref{sec:main}, the processing-level shift remains a candidate explanation for part of
the value and we say so there.

\textbf{WorldFloods-v2}~\cite{portales2023global}: the second OOD benchmark, 509
Sentinel-2 Level-1C (L1C) scene pairs, of which the test split holds 18 Copernicus EMS areas of interest (AOIs) from 11
events, tiled to \del{785} native $512\!\times\!512$ windows. \chgx{Pixels marked permanent in the
co-registered Joint Research Centre (JRC) yearly-history layer are set to ignore and seasonal water stays
scorable; tables abbreviate this flood-only protocol as WF-fo.}{All reference water is scored.}

\begin{newtext}
\textbf{WorldFloods-v2, held back}: the official test splits of Sen1Floods11 and WorldFloods contain only 10 and 11
events. To increase the number of events, we use the training and validation splits of WorldFloods-v2 (116 events) as a fourth test set. No model
or decision in this study used these splits. Before inspecting any labels, we removed every flood map whose outline
overlaps our training images or another test set. This leaves 97 events with 349 maps. Each finished model is scored once
using the WorldFloods test-split protocol.

\textbf{Water-free ground}: The four test sets described above primarily test how accurately water is identified, rather than how often the model falsely predicts water on dry land.
We assess false alarms on two additional water-free sets. They
differ in source and in how the absence of water is established:
\begin{itemize}
\item \emph{Water-free WorldFloods windows}: the WorldFloods test scenes are tiled into native $512\!\times\!512$
windows, and 200 of them contain no water in the expert reference map. They come from 9 of the 11 events, and three events
hold 159 of them, so the alarm rate is averaged over the 9 events.
\item \emph{Dry tiles}: 49 Sentinel-2 tiles at random land locations where ESA WorldCover~\cite{zanaga2022worldcover}
and the JRC Global Surface Water layer~\cite{pekel2016high} show no water, at least 25\,km from any training image,
downloaded using the GDACS procedure. One author inspected each full tile in true-color and short-wave-infrared
images. Every tile with the slightest sign of water, or any doubt, was removed. The set contains 28 ordinary tiles
and 21 with shadows or snow and has no expert reference (Supplementary Section~\ref{S-sec:s_audit}).
\end{itemize}
On both, a false alarm means that more than 1\% of the valid pixels are predicted as water.
\end{newtext}

\begin{deltext}
\paragraph*{\revst{Permanent-water harmonization}}
\revst{STURM explicitly separates permanent-water classes, whereas WorldFloods does not.
For a common flood-increment target we ignore permanent-water pixels in both OOD
evaluations. On WorldFloods, scoring those pixels as positive raises every model's
water IoU by 0.08--0.09 but does not change any model ranking. For water-class
scoring, the masking is the same operation as applying a reference-water mask before evaluation;
the surface-water semantics retained from the teacher are examined separately in
Section~IV-G.}
\end{deltext}
\begin{newtext}
\paragraph*{Surface water on every test set}
Because the teacher and its students segment surface water (Section~\ref{sec:inherit}), we score all surface water on every
test set, including permanent water. Removing permanent water to approximate flood water would lower every method's score
on STURM-Flood and WorldFloods.
\end{newtext}

\textbf{GDACS \chg{pseudo-label}{teacher-label} pool}: unlabeled Sentinel-2 L1C \chg{tiles}{images} acquired around flood
events listed by the Global Disaster Alert and Coordination System (GDACS), downloaded
via Google Earth Engine~\cite{gorelick2017google}. GDACS assigns each event a
Green/Orange/Red alert level from its estimated humanitarian impact; we query only
\emph{Orange} and \emph{Red} alerts, on the rationale that higher-impact events are the
ones likely to have produced an inundation large enough to be visible in a Sentinel-2
tile. GDACS reports an event as a point rather than a polygon, so a footprint has to be chosen
around it: we fetch the six Prithvi fine-tuning bands over a \chgx{$1{,}320$\,m radius of that
point, which at the delivered $\approx$9.9\,m sampling is a $264\!\times\!264$ pixel tile
covering $\approx$2.60\,km. This
is the largest window we could retrieve for every event in the catalog within the Earth
Engine request budget, and it is the one respect in which the pool is unlike the two OOD
benchmarks: a STURM chip ($128^2$, 1.28\,km) or a WorldFloods AOI (a full scene, tiled) is cut from a
delineated flood extent, whereas}{$512\!\times\!512$-pixel window of 10\,m pixels (5.12\,km) around that point, the size of a
Sen1Floods11 chip. To measure the effect of the extent, we also trained on the central $264\!\times\!264$ pixels (2.6\,km)
of the same images (Table~\ref{tab:onechange}). Unlike a STURM chip or a WorldFloods AOI, which are cut from a
delineated flood extent,} a GDACS \chg{tile}{image} is centered on a reported epicentre and may or
may not contain inundation at all, which is precisely what the pre-filter of
Section~\ref{sec:curation} then has to decide. \add{Before any teacher labeling, the downloader kept an event acquisition only if
MNDWI marked at least 1\% of it as water outside the permanent water of the Joint Research Centre (JRC) layer.} All three are mapped to a common input size and a
common normalization before any model sees them (Appendix~\ref{app:impl}). Because floods are caused by severe weather, the
flood-dense scenes we \chg{most want}{want most} are also \chg{the most}{severely} cloud-contaminated, so a strict cloud
filter would preferentially discard the imagery of interest; we accept up to 50\% cloud
cover, falling back to 80\% for events where nothing else passes, and apply no cloud
masking at any stage.

What survives download and screening is \textbf{\chg{23{,}720 tiles from 3{,}603 distinct
events}{23{,}400 images}}, acquired between \chg{December 2015 and July 2026}{July 2015 and August 2026} and spanning
\chg{$46^\circ$S}{$50^\circ$S}--$68^\circ$N; the $N\!=\!2{,}500$ training subset draws on \chg{1{,}404 of them}{1{,}370 events at 2{,}151 locations}.
Half the pool is a second acquisition: \chg{11{,}895 tiles lie inside an event window and
11{,}825}{11{,}735 images lie inside an event window and 11{,}665} are \emph{pre-event baselines} of the same location, drawn 30--395 days before
the event start. Relative to the paired event-window acquisition, the realized separation
is 25--502 days (median 300); baselines are retrieved under a stricter 20\% cloud gate, and an audit of the
acquisition dates finds every baseline strictly earlier than every event-window scene at
its location.

\paragraph*{Acquisition and labeling of the pool}
Event-window scenes are retrieved inside the GDACS event's own reported interval
(\texttt{fromdate}--\texttt{todate}), widened by seven days on each side only when no scene
passes the cloud gate; pre-event baselines use the same fixed 30--395-day window before
the event start described above. Where several Sentinel-2 Level-1C scenes fall inside a window we take the single
least-cloudy one and never composite. \chgx{Each tile is stored at native resolution
($264\!\times\!264$ for the 10\,m bands, $132\!\times\!132$ for the 20\,m bands); the 20\,m
bands are bilinearly resampled onto the 10\,m grid and the tile is bilinearly resized to
$224\!\times\!224$ before the teacher, which is also the resolution at which the teacher's hard
mask and per-pixel confidence map are stored. The $512\!\times\!512$ input size used elsewhere in
this paper applies to benchmark evaluation, not to pool generation.}{Each image is stored on the 10\,m grid at
$512\!\times\!512$ pixels, with the 20\,m bands repeated onto that grid. The teacher labels the image and the students train
at this size.}

\paragraph*{Contamination audit}\ifrevmarked\leavevmode\fi 
\begin{oldtext}
\revst{Because GDACS and the OOD benchmarks describe some of the same flood events, we checked
geographic and temporal overlap explicitly. At the matched budget, no STURM chip or
WorldFloods test AOI lies within 25\,km of any of the 252 pseudo-labeled training tiles.
At $N\!=\!2{,}500$, a stricter 5\,km/2-day test leaves no overlap on either benchmark.
A looser 25\,km/30-day screen flags three STURM activations; removing them leaves the
matched-budget pseudo-minus-Sen1Floods11 gap essentially unchanged ($+0.0193$~mIoU per flood
map over all 46, $+0.0189$ over the remaining 42). Distances are computed between chip and AOI
centroids. STURM and WorldFloods also share a small number of EMS activations, so we
treat them as complementary rather than independent; removing the overlapping WorldFloods
AOIs changes all reported values by less than 0.001. The full distance/date audit is
reported in Supplementary Section~S1.}
\end{oldtext}
\begin{newtext}
Because GDACS and the test sets describe some of the same flood events, we compared the projected footprints and dates
of every training image with those of every test image. No training image overlaps a test image on the ground (Supplementary
Section~\ref{S-sec:s_audit}). STURM and WorldFloods share three EMS activations, so we treat them as complementary rather
than independent.
\end{newtext}

\chg{All models are evaluated on the same held-out test splits. Reported numbers are
mean $\pm$ std across 4 random training seeds, except the teacher (single run).}{All models are evaluated on the same test
sets, and no model is trained on any of them. Reported numbers are means over four training seeds. The
teacher and MNDWI are deterministic.}

\begin{table}[t]
\revnewfloat
\centering
\caption{\add{\textbf{Data roles.} Every checkpoint, learning rate and 8-bit calibration is chosen on the Sen1Floods11 validation split
only. The test sets evaluate finished models.}}
\label{tab:roles}
\footnotesize
\setlength{\tabcolsep}{3pt}
\begin{tabular}{@{}p{0.43\columnwidth}p{0.53\columnwidth}@{}}
\toprule
Data & Role \\
\midrule
S11 train, 252 images & Teacher fine-tuning (public checkpoint) and hand labels for the hand-label students \\
S11 validation, 89 images & Checkpoints, learning rates of the small students, 8-bit calibration and acceptance. No gradient updates \\
GDACS pool, 2,500 of 23,400 images & Student training with teacher labels and no human labels \\
\midrule
S11 test, 90 images, 10 events & Test only \\
WorldFloods test split, 11 events & Test only \\
STURM-Flood, 2,675 chips, 29 events & Test only (no official split) \\
WorldFloods train/val, held back, 97 events & Test only, scored once per finished model \\
Water-free ground, 200 windows and 49 tiles & Test of false alarms only \\
\bottomrule
\end{tabular}
\end{table}

\begin{newtext}
\paragraph*{Data roles and selection}
Table~\ref{tab:roles} lists the data used for training, selection and testing. We select the checkpoint with the highest water IoU
per event on the 89 Sen1Floods11 validation images. Validation is performed after every epoch, or every 156 updates in runs with a
fixed number of updates. The same images determine each small student's learning rate and the 8-bit
calibration and acceptance. The thresholds of the tuned water-index variants in the supplement are
set on the Sen1Floods11 training and validation images. No test set is used to choose a model, a threshold or a
hyperparameter.
\end{newtext}

\begin{oldtext}
\paragraph*{\revst{Micro and macro aggregation}}
\revst{We report both pixel-pooled (micro) and unweighted per-scene (macro) IoU. This matters
most on WorldFloods, where scene sizes and flood-pixel counts are highly imbalanced and
the model ordering can change with the aggregation rule. STURM is also strongly
size-imbalanced, and scene size correlates with difficulty there (Spearman $+0.709$ over
46 flood maps, $p<10^{-4}$). Absolute micro and macro values therefore answer different
weighting questions and are never compared directly.}

\paragraph*{\revst{Statistical Protocol and Units of Replication}}
\revst{Most comparisons here can be tested along two axes.
\emph{Seed level} ($n\!=\!4$) asks whether another training seed would reproduce an effect.
We report such results as ``consistent across four seeds'' and not as significance.
\emph{Group level} asks whether another flood scene would reproduce it, and there
$n\!=\!46$ flood maps or 29 events on STURM and 18 scenes or 11 events on WorldFloods. We
average the four seeds within each group first, so the unit of replication is the scene
and not the seed, pair them across groups, and report the two-sided Wilcoxon signed-rank
test~\cite{wilcoxon1945} with Pratt's treatment of zero differences~\cite{pratt1959}.
For predefined families of related comparisons, we control the family-wise error rate using
Holm's step-down correction~\cite{holm1979}. A 10{,}000-sample group bootstrap gives the one confidence interval we
report, on the student--teacher difference. Because scenes inside one EMS activation are
correlated, we also give the coarser event-level result for the two comparisons on which the
paper's claims rest. Secondary statistics, including exact sign tests, win rates, and alternative
groupings, are in Supplementary Section~S1.}
\end{oldtext}
\begin{newtext}
\paragraph*{Measure: water IoU per flood event}
\label{sec:stats}
Our main measure is the IoU of the water class, $\mathrm{TP}/(\mathrm{TP}+\mathrm{FP}+\mathrm{FN})$, computed over the
pixels of each flood event and then averaged with equal weight across events. Pooling all pixels of a
test set instead gives greater influence to events with more water. On WorldFloods, a few large scenes contain most of the
water pixels. For the learned models, per-event values are lower than pooled values on Sen1Floods11 and STURM-Flood,
with the largest decrease on STURM-Flood. Many events there contain little water, so a small boundary error can substantially
reduce an event's IoU. On WorldFloods, the learned models instead have higher per-event than pooled scores. We fixed this measure before
training any of the models reported here. The comparison of student and teacher does not depend on the measure: on the
Sen1Floods11 test split the teacher reaches 81.7 and the main student 78.5 pooled water IoU, against 79.1 and 76.1 per
event, and on WorldFloods the student is level with or ahead of the teacher under both. Supplementary
Section~\ref{S-sec:s_measures} gives pooled and per-event water IoU for every model.

\paragraph*{Statistics}
Each setting is trained with four random seeds to account for variation between training runs. To compare two settings, we first average the results across seeds within each flood event and calculate the paired performance difference. These differences are then averaged across events. We estimate 95\% confidence intervals using 10{,}000 bootstrap resamples of whole events. This preserves the event as the unit of replication and accounts for variation across flood events while keeping the trained seeds fixed. For the planned comparisons, we additionally apply a two-sided Wilcoxon signed-rank test~\cite{wilcoxon1945}, using Pratt's method for zero differences~\cite{pratt1959} and Holm's correction for multiple comparisons~\cite{holm1979}. The Sen1Floods11 and WorldFloods test splits contain 10 and 11 events, STURM-Flood 29 and the held-back set 97. We report confidence intervals alongside the performance differences to make the uncertainty visible, particularly for the two smaller official test splits.
\end{newtext}

\begin{oldtext}
\paragraph*{\revst{Choice of student architecture}}
\revst{The choice is a deployment constraint first and an accuracy result second, so we state
what it does and does not rest on. Under the two objectives that all backbones share, an
ablation against MobileNetV3-Small (3.55\,M) and a non-pretrained 1.64\,M CNN puts the
three architectures within 0.015~mIoU of each other under each shared objective on STURM, of the order of their seed
spreads (Supplementary Section~S8; those runs share one recipe with each other but not with
Tables~I and~II): no architecture
dominates on a comparable basis, and we do not claim EfficientViT-B0 is
intrinsically the best OOD backbone. The deciding property is the deployment path. EfficientViT-B0 with OFA is the best row of that
ablation at $5\times$ fewer parameters than the runner-up (MobileNetV3-Small with Logit-KD,
3.55\,M), it is the smallest of the three, and it is the configuration we carried to INT8
on-device after the activation replacement of Section~III-D. We did not port OFA's branch heads to the other backbones, so we cannot
rule out that OFA would help them too.}
\end{oldtext}

\emph{Naming.} Two supervision sources are compared throughout, and each has exactly one
name. \chgx{\emph{Sen1Floods11} (abbreviated \emph{S11} in table headers and figure legends) means
the hand-drawn masks of the Sen1Floods11 training split, at most 252 of them; \emph{pseudo}
means masks written by the frozen teacher on GDACS tiles. A configuration is named by its
label source, so ``the Sen1Floods11 student'' is the student trained on those masks, never
the student evaluated on that benchmark.}{\emph{Hand labels} are the human masks of the 252 Sen1Floods11 training images.
\emph{Teacher labels} are the masks produced by the frozen teacher on GDACS images unless stated otherwise. A student is
named by its training data, never by the benchmark it is evaluated on. We call EfficientViT-B0 the \emph{main student}.}
Every benchmark is named explicitly where it is used.

\subsection{\chghead{Baselines and Implementation Details}{Models and Training}}
\label{sec:baselines}

\begin{oldtext}
\revst{We compare against the following baselines:}

\begin{itemize}
  \item \revst{\textbf{Teacher (Prithvi-EO-2.0, 300\,M)}: upper reference only; not benchmarked on the embedded targets in this study.}
  \item \revst{\textbf{UNet-ResNet50, CE, S11} (32.5\,M): standard UNet~\cite{ronneberger2015u}
  with ResNet-50 encoder, trained on Sen1Floods11 labels.}
  \item \revst{\textbf{EfficientViT-B0, CE, S11} (0.7\,M): student architecture trained
  on Sen1Floods11 labels only. This is the one configuration in the paper with no
  teacher involvement at all.}
  \item \revst{\textbf{EfficientViT-B0, Logit-KD and OFA, S11} (0.7\,M) and
  \textbf{UNet-ResNet50, Logit-KD, S11} (32.5\,M): the label-source counterparts of the
  pseudo-label entries below, with the same objectives and the \emph{same frozen teacher},
  with the Sen1Floods11 human masks as the hard target (Section~III-C). These are the
  Sen1Floods11 rows of Table~I and the Sen1Floods11 curves of
  Figs.~4 and~5, and Supplementary Fig.~S4.}
  \item \revst{\textbf{EfficientViT-B0, CE, pseudo} (0.7\,M): student trained on
  pseudo-labels with plain cross-entropy, no distillation loss.}
  \item \revst{\textbf{UNet-ResNet50, CE, pseudo} (32.5\,M): the heavy baseline trained
  on the same pseudo-label pool.}
  \item \revst{\textbf{EfficientViT-B0, Logit-KD, pseudo} (0.7\,M): standard
  temperature-scaled logit distillation.}
  \item \revst{\textbf{UNet-ResNet50, Logit-KD, pseudo} (32.5\,M): a higher-capacity non-edge reference.}
  \item \revst{\textbf{EfficientViT-B0, OFA, pseudo (ours)} (0.7\,M): our primary
  model.}
\end{itemize}

\revst{Every pseudo-label entry at a given budget is trained on the same FPS-curated subset of the GDACS pool, so
architecture and loss comparisons are made on identical data. EfficientViT-B0 is
initialized from ImageNet-pretrained weights and UNet-ResNet50 from an
ImageNet-pretrained ResNet-50 encoder; all models use AdamW with a cosine
learning-rate schedule. Every Sen1Floods11-versus-pseudo-label comparison is
augmentation-matched: both configurations train with the same dihedral flips and the same
per-channel multiplicative jitter in $[0.80,1.20]$. For Logit-KD and OFA, soft targets are obtained from a live forward pass of the frozen teacher on the correspondingly augmented image; no soft logit map is stored in the pseudo-label pool
(Appendix~A). Because mixed precision and non-deterministic kernels alone
move results by 0.003--0.010~mIoU, each table is trained under a single recipe, stated in
its caption, and numbers are never compared across recipes. Full hyperparameters are in Appendix~A.}
\end{oldtext}
\begin{newtext}
We compare the teacher (an upper reference, not benchmarked on the device), MNDWI with threshold zero (no parameters),
and the students of Section~\ref{sec:students}. Each student is trained on the 2,500 teacher-labeled GDACS
images in up to three ways: with logit distillation (the proposed setup), with labels only, or with the OFA branches. The reference students learn
from the 252 hand labels, without the teacher or with the teacher's probability maps. Every teacher-label configuration uses the same
FPS-curated subset, so architecture and loss comparisons are made on identical data.

All students train at $512\!\times\!512$ with AdamW, weight decay 0.05, a cosine schedule, batch size 16, gradient
clipping at 5 and FP32 arithmetic with deterministic kernels. Fifty epochs over 2,500 images give 7,800 updates. When comparing 252
images with 2,500 at an equal number of updates, the 252-image setting also uses 7,800 updates. The learning
rate is $5\times10^{-5}$ for pretrained students. For each small student, we choose $5\times10^{-4}$ or $5\times10^{-3}$
using the Sen1Floods11 validation split. All experiments use the same dihedral flips and per-channel multiplicative jitter in
$[0.80,1.20]$, and soft targets come from a live forward pass of the frozen teacher on the same augmented image. All
students share this training setup, which differs from the original setup of each architecture. LightFormer's authors, for example, use higher learning
rates and an auxiliary loss. Our comparison therefore evaluates the designs under a common setup rather than their individually optimized
performance. To assess this limitation, we also trained LightFormer with its authors' setup~\cite{chen2026lightformer}. This uses AdamW
with higher learning rates, cross-entropy with Dice and auxiliary losses, multi-scale cropping, photometric
augmentation and early stopping on the Sen1Floods11 validation split. We used this setup without the teacher's maps on the 252 hand labels
and on the 2,500 teacher labels (Section~\ref{sec:analysis}). Full hyperparameters are in Appendix~\ref{app:impl}.
\end{newtext}

\section{Results}
\label{sec:results}

\begin{deltext}
\revst{\textbf{Submitted Section IV-C, Main Results.}
We report two operating points, and keep them in separate tables because they answer
different questions. Table~I fixes the training budget at 252 scenes,
the complete Sen1Floods11 training split, and therefore the largest budget at which the
Sen1Floods11 and teacher-labeled configurations can be compared at all, and asks what
routing the labels through the teacher is worth.
Table~II removes that cap and reports the deployment configuration.}
\end{deltext}

\subsection{\texorpdfstring{\add{Student and Teacher on Four Test Sets}}{Student and Teacher on Four Test Sets}}
\label{sec:main}

\begin{figure*}[t]
\revnewfloat
\centering
\includegraphics[width=\linewidth]{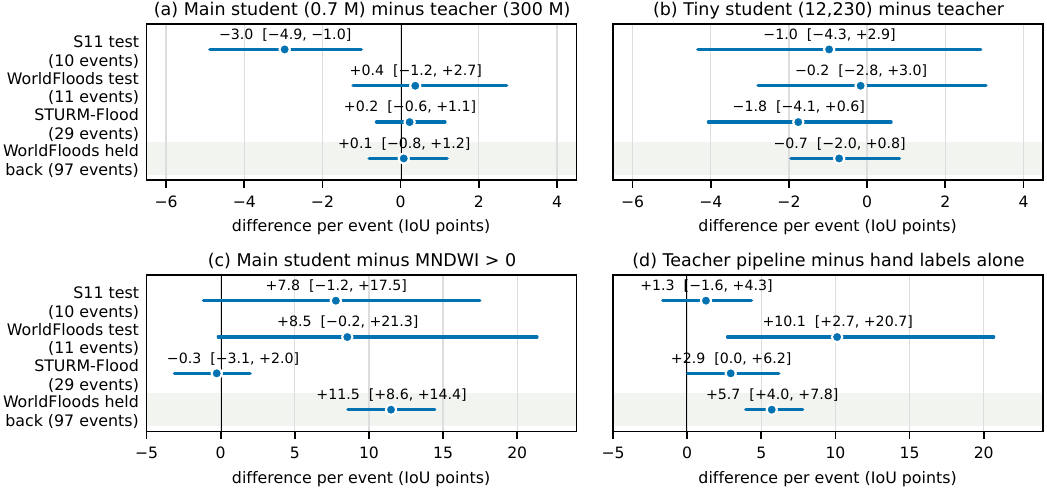}
\caption{\add{\textbf{Paired comparisons on four test sets.} Paired differences in water IoU per flood event, with 95\% intervals from 10,000 bootstrap resamples of whole
events (students: mean of four seeds). (a) Main student minus teacher. (b) Tiny student minus teacher. (c) Main student
minus MNDWI with threshold zero. (d) Main student minus the same design trained on the 252 hand-labeled Sen1Floods11
images without the teacher, both trained for 7,800 updates. The hand-label run decays its learning rate after every update,
the proposed setup after every epoch. Values to the right of zero favor the first-named model. Shading marks
the 97 held-back events. Table~\ref{tab:main} lists both settings of (d) for all six students.}}
\label{fig:four}
\end{figure*}

\begin{oldtext}
\paragraph*{\revst{Student--teacher comparison at $N\!=\!2{,}500$}}
\revst{Removing the 252-scene cap moves the compact OFA student to $0.756 \pm 0.005$ on
STURM against the teacher's $0.755$. Averaging per flood map rather than per pixel
does not change the verdict: the paired difference over the 46 flood maps is
$-0.0012$, 95\% CI $[-0.008, +0.006]$, with the student ahead on 20 of 46 and
$p\!=\!0.46$, so the two are not separated at this unit of replication. Per event (29 groups) it is $+0.0007$, $p\!=\!0.78$.
This holds on one benchmark only. On WorldFloods the student remains below the teacher, by
$0.007$~mIoU pooled and $0.023$ per scene (ahead on 2 of 18 scenes,
$p_{\text{holm}}\!=\!0.002$), and on the in-distribution S11 split the 300\,M model is still
ahead by 0.020~mIoU. The claim is therefore specific: on STURM-Flood, the more shifted
out-of-distribution set under our protocol, a paired per-flood-map test does not separate the
0.7\,M student from the foundation model it was distilled from, and the confidence interval
bounds any remaining difference at well under $0.01$~mIoU.}

\revst{This comparison has three qualifications. It does not hold on WorldFloods. STURM's
processing-level shift (Appendix~A) may contribute to the reduced teacher lead.
The comparison also depends on the evaluation tensor: re-scoring the same weights at STURM-Flood's native $128\!\times\!128$ grid
moves the paired difference to $-0.028$ water IoU per flood map; the teacher is ahead, but the paired test
narrowly misses the $0.05$ threshold (student ahead on 16 of 46, $p\!=\!0.054$). Because STURM's ground extent is
fixed, this control changes effective sampling and network feature-map size jointly and cannot
separate them (Supplementary Section~S3).}
\end{oldtext}
\begin{newtext}
Table~\ref{tab:main} gives the water IoU per event of every model on the four test sets and its false-alarm rate on
water-free ground. Fig.~\ref{fig:four} gives the paired differences with 95\% intervals. Trained on the 2,500
teacher-labeled GDACS images, the main student matches the 300\,M teacher on the three test sets outside the
training distribution: the differences are $+0.4$ $[-1.2, +2.7]$ IoU points on WorldFloods, $+0.2$ $[-0.6, +1.1]$ on
STURM-Flood and $+0.1$ $[-0.8, +1.2]$ on the 97 held-back events. On the Sen1Floods11 test split, whose images come from
the same events as the teacher's fine-tuning data, the teacher stays ahead of the main student by 3.0 $[1.0, 4.9]$ IoU points. The tiny student,
24,500 times smaller than the teacher, stays within 2 IoU points of it on all four sets, and all its intervals include zero.

On STURM-Flood, its
different processing (Appendix~\ref{app:impl}) may contribute to the parity, and scoring at its native
$128\!\times\!128$ grid changes the comparison (Supplementary Section~\ref{S-sec:s_harmonisation}). Larger designs
trained on the same 2,500 teacher-labeled images are not clearly better (Table~\ref{tab:main}): MobileNetV3-Small,
LightFormer-ResNet18 and UNet-ResNet50, the common flood-segmentation baseline with 46 times more parameters, stay within
2.1 IoU points of the main student on every test set. None performs better on all four. The main student is the smallest of
these four designs, so we select it for deployment.
\end{newtext}

\begin{delfloat}
\begin{table}[t]
\revdelfloat{TABLE II}
\centering
{\footnotesize \revst{Submitted Table~II. \textbf{Deployment operating results.} Pool-cap removed: $N\!=\!2{,}500$
  FPS-selected (curated dataset) tiles. The Sen1Floods11 column of
  Table~I has no counterpart here because Sen1Floods11 has no further images
  to add. Micro mIoU; subscripts are the seed standard deviation over 4 seeds.
  EfficientViT-B0 is 0.7\,M parameters, UNet-ResNet50 32.5\,M; $\dagger$ =
  higher-capacity model not tested for deployment. Bold marks the best model \& distillation per column.}\par}
\setlength{\tabcolsep}{4pt}
\resizebox{\columnwidth}{!}{%
\begin{tabular}{lccc}
\toprule
Model ($N\!=\!2{,}500$ teacher-supervised) & S11 & STURM & WF-fo \\
\midrule
Prithvi-EO-2.0, teacher & $0.897$ & $0.755$ & $0.771$ \\
\midrule
EfficientViT-B0, CE & $0.873_{\pm.002}$ & $0.754_{\pm.003}$ & $0.754_{\pm.008}$ \\
EfficientViT-B0, Logit-KD & $0.875_{\pm.003}$ & $\mathbf{0.758}_{\pm.003}$ & $\mathbf{0.767}_{\pm.009}$ \\
\emph{EfficientViT-B0, OFA (ours)} &
  $0.877_{\pm.003}$ & $0.756_{\pm.005}$ & $0.764_{\pm.002}$ \\
UNet-ResNet50, CE$^\dagger$ & $0.881_{\pm.013}$ & $0.730_{\pm.016}$ & $0.755_{\pm.018}$ \\
UNet-ResNet50, Logit-KD$^\dagger$ & $\mathbf{0.889}_{\pm.006}$ & $0.747_{\pm.005}$ & $0.762_{\pm.003}$ \\
\bottomrule
\end{tabular}}
\end{table}
\end{delfloat}

\begin{table*}[t]
\revnewfloat
\centering
\caption{\add{\textbf{Accuracy on four test sets and false alarms.} Water IoU per flood event (in points, mean $\pm$ standard deviation over four seeds) and false alarms, the
share of 200 water-free WorldFloods windows with more than 1\% predicted water (averaged over 9 events). Bold: best
student per column within each block. Ours: the two students we deploy (Section~\ref{sec:deployment}).
$^\ddagger$Its authors' losses, augmentation and early stopping. Pooled values: Supplementary Table~\ref{S-tab:s_pooled}.}}
\label{tab:main}
\footnotesize
\setlength{\tabcolsep}{4.5pt}
\begin{tabular}{llccccc}
\toprule
Model & Parameters & S11 test & WorldFloods test & STURM-Flood & Held back & False alarms \\
 & & (10 events) & (11 events) & (29 events) & (97 events) & (\%) \\
\midrule
\multicolumn{7}{l}{\itshape References} \\
Teacher, Prithvi-EO-2.0 & 300\,M & 79.1 & 79.4 & 49.9 & 60.7 & 1.5 \\
MNDWI $>0$ & none & 68.3 & 71.2 & 50.4 & 49.3 & 57.0 \\
\midrule
\multicolumn{7}{l}{\itshape Students, 2,500 teacher-labeled GDACS images, logit distillation (proposed setup)} \\
EfficientViT-B0 (main student, ours) & 0.7\,M & $76.1 \pm 0.6$ & $\mathbf{79.7} \pm 0.2$ & $50.1 \pm 0.8$ & $60.8 \pm 0.5$ & 5.4 \\
MobileNetV3-Small & 3.55\,M & $77.1 \pm 0.6$ & $79.6 \pm 0.2$ & $\mathbf{51.4} \pm 0.3$ & $60.4 \pm 0.5$ & 2.9 \\
LightFormer-ResNet18 & 11.43\,M & $75.9 \pm 1.3$ & $78.7 \pm 0.6$ & $49.3 \pm 1.1$ & $59.8 \pm 0.5$ & 9.8 \\
UNet-ResNet50 & 32.5\,M & $\mathbf{78.2} \pm 0.8$ & $79.3 \pm 0.5$ & $48.5 \pm 0.8$ & $\mathbf{61.4} \pm 0.4$ & \textbf{2.2} \\
Tiny student (ours) & 12,230 & $78.1 \pm 2.0$ & $79.2 \pm 0.3$ & $48.1 \pm 2.1$ & $60.0 \pm 0.7$ & 7.1 \\
Micro student & 1,266 & $76.0 \pm 1.3$ & $78.1 \pm 0.8$ & $49.8 \pm 1.9$ & $58.7 \pm 1.0$ & 13.0 \\
\midrule
\multicolumn{7}{l}{\itshape Same designs, 252 hand-labeled S11 images, 7,800 updates, no teacher unless stated} \\
EfficientViT-B0 & 0.7\,M & $74.8 \pm 1.7$ & $69.6 \pm 3.9$ & $47.2 \pm 1.5$ & $55.1 \pm 1.2$ & 11.8 \\
\quad with the teacher's maps (main loss) & 0.7\,M & $75.7 \pm 0.9$ & $74.9 \pm 2.8$ & $\mathbf{50.9} \pm 1.1$ & $57.0 \pm 1.1$ & \textbf{4.5} \\
MobileNetV3-Small & 3.55\,M & $76.8 \pm 1.2$ & $75.1 \pm 1.4$ & $48.3 \pm 3.5$ & $55.1 \pm 0.5$ & 15.6 \\
LightFormer-ResNet18 & 11.43\,M & $75.6 \pm 0.7$ & $71.7 \pm 0.4$ & $48.8 \pm 0.7$ & $55.9 \pm 0.7$ & 11.0 \\
\quad with its authors' setup$^\ddagger$ & 11.43\,M & $78.3 \pm 1.0$ & $78.1 \pm 1.9$ & $50.2 \pm 2.6$ & $57.9 \pm 0.2$ & 11.1 \\
UNet-ResNet50 & 32.5\,M & $80.8 \pm 0.3$ & $78.0 \pm 0.9$ & $48.7 \pm 1.5$ & $56.9 \pm 0.5$ & 9.9 \\
Tiny student & 12,230 & $\mathbf{81.2} \pm 1.2$ & $\mathbf{79.2} \pm 0.8$ & $45.8 \pm 5.6$ & $\mathbf{59.3} \pm 0.5$ & 11.2 \\
Micro student & 1,266 & $80.2 \pm 1.0$ & $79.0 \pm 0.7$ & $50.2 \pm 0.7$ & $58.0 \pm 0.6$ & 17.5 \\
\bottomrule
\end{tabular}
\end{table*}

\subsection{\texorpdfstring{\add{One Change at a Time}}{One Change at a Time}}
\label{sec:onechange}

\begin{delfloat}
\begin{table*}[t]
\revdelfloat{TABLE I}
\centering
{\footnotesize \revst{Submitted Table~I. \textbf{Matched training budget, $N\!=\!252$ scenes.} Every row trains on 252 images.
  This is the complete Sen1Floods11 training split and therefore the largest budget
  at which the two label sources can be compared directly:
  the \emph{Sen1Floods11} rows use that split, while the \emph{pseudo} rows use 252
  GDACS tiles labeled by the frozen Prithvi teacher, which was fine-tuned on the same
  split and never on GDACS. Architectures, objectives, hyperparameters, seeds and
  augmentation are identical. What differs is the label source and the imagery each pool
  supplies with it, including its native chip extent (4.63 × 5.12 \,km for Sen1Floods11, 2.60 × 2.62 \,km
  for GDACS; Appendix~A). Mean\,$\pm$\,std
  over 4 seeds, pixel-pooled (micro); teacher single-run. Bold marks the better of each Sen1Floods11/pseudo pair per column.}\par}
\setlength{\tabcolsep}{4pt}
\resizebox{\textwidth}{!}{%
\begin{tabular}{llcccccc}
\toprule
& & \multicolumn{2}{c}{S11 test (in-distribution)} & \multicolumn{2}{c}{STURM-Flood (OOD)} & \multicolumn{2}{c}{WorldFloods-v2 (OOD)} \\
\cmidrule(lr){3-4}\cmidrule(lr){5-6}\cmidrule(lr){7-8}
Model (params) & Labels & mIoU & water IoU & mIoU & water IoU & mIoU & water IoU \\
\midrule
Teacher, Prithvi-EO-2.0 (300\,M) & --- & $0.897$ & $0.822$ & $0.755$ & $0.643$ & $0.771$ & $0.641$ \\
\midrule
\multirow{2}{*}{EfficientViT-B0, CE (0.7\,M)}
  & Sen1Floods11 & $\mathbf{0.844} \pm .002$ & $\mathbf{0.731} \pm .005$ & $0.717 \pm .018$ & $0.597 \pm .017$ & $0.744 \pm .018$ & $0.606 \pm .029$ \\
  & pseudo       & $0.825 \pm .008$ & $0.696 \pm .013$ & $\mathbf{0.735} \pm .003$ & $\mathbf{0.615} \pm .005$ & $\mathbf{0.748} \pm .015$ & $\mathbf{0.607} \pm .024$ \\
\multirow{2}{*}{EfficientViT-B0, Logit-KD (0.7\,M)}
  & Sen1Floods11 & $0.836 \pm .006$ & $0.717 \pm .009$ & $0.733 \pm .009$ & $0.618 \pm .011$ & $0.752 \pm .004$ & $\mathbf{0.617} \pm .005$ \\
  & pseudo       & $\mathbf{0.842} \pm .005$ & $\mathbf{0.727} \pm .009$ & $\mathbf{0.753} \pm .002$ & $\mathbf{0.642} \pm .003$ & $\mathbf{0.753} \pm .014$ & $0.614 \pm .023$ \\
\multirow{2}{*}{\emph{EfficientViT-B0, OFA (0.7\,M)}}
  & Sen1Floods11 & $\mathbf{0.853} \pm .001$ & $\mathbf{0.745} \pm .001$ & $0.736 \pm .006$ & $0.623 \pm .006$ & $\mathbf{0.770} \pm .008$ & $\mathbf{0.645} \pm .012$ \\
  & pseudo       & $0.852 \pm .002$ & $0.743 \pm .003$ & $\mathbf{0.753} \pm .007$ & $\mathbf{0.640} \pm .011$ & $0.760 \pm .010$ & $0.625 \pm .017$ \\
\midrule
\multirow{2}{*}{UNet-ResNet50, CE (32.5\,M)}
  & Sen1Floods11 & $\mathbf{0.862} \pm .018$ & $\mathbf{0.762} \pm .028$ & $0.721 \pm .024$ & $\mathbf{0.616} \pm .026$ & $\mathbf{0.772} \pm .007$ & $\mathbf{0.650} \pm .008$ \\
  & pseudo       & $0.842 \pm .011$ & $0.724 \pm .020$ & $\mathbf{0.736} \pm .015$ & $0.613 \pm .025$ & $0.759 \pm .008$ & $0.625 \pm .012$ \\
\multirow{2}{*}{UNet-ResNet50, Logit-KD (32.5\,M)}
  & Sen1Floods11 & $\mathbf{0.876} \pm .007$ & $\mathbf{0.784} \pm .012$ & $0.739 \pm .010$ & $0.632 \pm .007$ & $\mathbf{0.785} \pm .003$ & $\mathbf{0.667} \pm .005$ \\
  & pseudo       & $0.863 \pm .004$ & $0.761 \pm .007$ & $\mathbf{0.763} \pm .002$ & $\mathbf{0.658} \pm .003$ & $0.780 \pm .006$ & $0.658 \pm .010$ \\
\bottomrule
\end{tabular}}
\end{table*}
\end{delfloat}

\begin{oldtext}
\paragraph*{\revst{Matched-budget comparison}}
\revst{Table~I asks whether a fixed manual annotation budget is more useful when the compact student is trained directly on the 252 Sen1Floods11 scenes or when the same task teacher supervises 252 different flood-event scenes. On STURM-Flood the teacher-supervised configuration is ahead in all five architecture$\times$objective cells, by $+0.017$ to $+0.024$~mIoU per flood map. Four of the five paired Wilcoxon comparisons survive Holm correction within the five-cell block (Holm-corrected $p$: EfficientViT-CE $3.8\times10^{-3}$, Logit-KD $3.3\times10^{-3}$, OFA $8.9\times10^{-3}$, UNet-Logit-KD $3.3\times10^{-4}$); UNet-CE does not ($p\!=\!0.086$). At the coarser event level (29 activations) three of the five survive. This supports teacher supervision as a competitive substitute for direct use of the manual split on this benchmark, not teacher-generated labels being intrinsically better than human annotations.}

\revst{The comparison carries a geometric qualification. Sen1Floods11 and GDACS chips do not cover the same ground extent before the common training resize. Retraining the Sen1Floods11 arm with each chip center-cropped to the GDACS tiles' extent (four seeds, otherwise identical recipe) changes the three EfficientViT cells differently: OFA falls to $+0.006$~mIoU per flood map (27 of 46, $p\!=\!0.071$), cross-entropy to $-0.0004$ (25 of 46, $p\!=\!0.84$), while Logit-KD rises to $+0.025$ (38 of 46, $p\!=\!7.9\times10^{-6}$). The crop is not a clean isolation of extent, since it also removes more than half of the positive area from 86 of the 252 chips and brings training sampling closer to evaluation sampling. It does establish that the matched-budget difference cannot be attributed to label source alone.
The scaling experiment is not bound by the same fixed-data constraint: additional unlabeled scenes can be supervised by the teacher without requiring new manual annotations.}

\revst{On the Sen1Floods11 test split, direct supervision remains favoured in most matched cells, as expected for same-distribution training. WorldFloods-v2 is more sensitive to aggregation because a single AOI carries 57\% of its flood pixels, so the pooled score is close to a one-scene measurement; we therefore report both pooled and per-scene summaries and do not use that benchmark for a general label-source claim.}

\paragraph*{\revst{Effect of the soft-target term at matched budget}}
\revst{In the EfficientViT block, plain cross-entropy on teacher-generated labels is the weakest
of the three pseudo-label configurations on all three benchmarks, and adding the temperature-scaled soft
term to the hard-target loss moves it by $+0.018$~mIoU on STURM, $+0.017$ on S11 and
$+0.005$ on WorldFloods. That step changes three things at once, the soft term, its weight
and the temperature, so we attribute the effect to the step and not to the soft target
alone. Resolved per flood map on STURM the step is $+0.0119$ on 34 of 46 maps
($p\!=\!2.6\times10^{-4}$), so only the CE-to-Logit-KD difference is statistically
resolved at this budget. We state it as a property of this regime and not as a general claim
about soft labels, because the same step on \emph{Sen1Floods11} labels produces a comparable
$+0.016$ on STURM at $N\!=\!252$. What remains specific to the pseudo-label regime is that an argmax discards the
teacher's confidence field. The soft target retains that spatial confidence information,
which can improve sample efficiency even though it introduces no independent label source
(Supplementary Section~S9). We do not use the size of the observed gain as an information
bound.}
\end{oldtext}
\begin{deltext}
\paragraph*{\revst{Objective comparison, resolved per flood map}}
\revst{On pseudo-labels the ordering CE\,$\to$\,Logit-KD\,$\to$\,OFA is monotone on S11 and
WorldFloods, while on STURM the top two objectives tie ($0.735 \to 0.753 \to 0.753$).
With four seeds, no individual step is significant because the exact sign test cannot reach
$p\!<\!0.05$ at $n\!=\!4$. We therefore resolve the two steps per flood map, over the
46 STURM flood maps. The result is the same for both label sources at this budget: the
\emph{Logit-KD} step is statistically resolved and the \emph{OFA} step is not. On Sen1Floods11
labels the Logit-KD step is $+0.0169$ on 37 of 46 maps ($p\!=\!2.4\times10^{-5}$) and the
OFA step is $-0.0023$ on 18 of 46 (n.s.); on pseudo-labels they are $+0.0119$ on 34 of 46
($p\!=\!2.6\times10^{-4}$) and $-0.0018$ on 22 of 46 (n.s.). At the deployment budget
neither survives. This is consistent with Fig.~2: OFA's advantage over logit distillation
occupies a window that closes at $N\!\approx\!251$, and the full Sen1Floods11 budget of 252 sits
just past its right edge. We therefore do not claim OFA is the better objective at a matched
budget, only that it has the strongest seed mean in the intermediate range
$50 \le N < 252$; below that range the sign is not consistent. We deploy it for the reasons
given in Section~IV-C and Section~III-D.}
\end{deltext}
\begin{newtext}
Table~\ref{tab:onechange} separates the three factors that change together when comparing 252 Sen1Floods11 images with 252 GDACS images:
imagery, tile extent and label source. It also reports the effect of each pipeline component. Unless
stated otherwise, comparisons use the main student.

\emph{Label source.} Teacher labels achieve human-level annotation quality. On the same 252 Sen1Floods11 images, they produce the same student accuracy as human labels. All
differences are within 0.8 IoU points, with or without the probability maps. The teacher was fine-tuned on these images. The
finding therefore supports parity with human labels rather than superiority over them.

\emph{Image extent.} The GDACS tiles of the original experimental setup covered 2.6\,km, half the width of a Sen1Floods11 chip
(5.12\,km). Training on such crops of the new images costs 4.4 IoU points on the Sen1Floods11 test split and gains 1.8 on
STURM-Flood, so the extent alone moves the scores. We therefore download every image at the Sen1Floods11 extent, and all
other results use 5.12\,km images.

\emph{Imagery.} With the same teacher labels and 252 images each, the Sen1Floods11 images are better than the GDACS
images on the Sen1Floods11 test split (by 2.7 to 3.2 IoU points) and on STURM-Flood (by 2.6 to 5.8), and slightly worse on
WorldFloods (by 1.1 to 1.5, with all intervals including zero). Teacher labels on GDACS images therefore do not raise
STURM-Flood accuracy at equal image count. GDACS images differ from Sen1Floods11 images in scenes, places and dates at once,
so this comparison does not isolate geographic diversity either.

\emph{Probability maps and number of images.} The teacher's probability maps help most when images are few, mainly outside the
training distribution. In each comparison, both settings give the hard labels the same weight, so the difference measures
the maps alone. At equal training length (7,800 updates), the maps add 5.3 and 3.8 IoU points on WorldFloods and
STURM-Flood with 252 hand-labeled images, but only 0.5 and 1.5 with 2,500 teacher-labeled images; on the Sen1Floods11
test split the gain is similar in both cases. More images help too: with maps and the same number of updates, 2,500
teacher-labeled images give $+3.8$ $[+2.2, +5.6]$ IoU points over 252 hand-labeled ones on the held-back events. There,
the complete teacher pipeline adds $+5.7$ $[+4.0, +7.8]$ IoU points over training on the hand labels alone
(Fig.~\ref{fig:four}d). On the public test sets, neither comparison passes Holm correction ($p\!\geq\!0.111$);
Table~\ref{tab:onechange} lists all values.

\emph{Learning curves.} Twelve more experiments of the main student were evaluated at ten fixed intermediate points of training (Supplementary Section~\ref{S-sec:s_convergence}). We compare the second half of training. From 3,900 updates onward, the 2,500 teacher-labeled images keep the
student within 1 IoU point of the teacher on all three test sets. The 252 hand-labeled images leave it 4.0 to 4.7
IoU points below the teacher on WorldFloods and 3.3 to 3.8 points below on the held-back events. On STURM-Flood, the 252 hand-labeled images instead keep it 0.5 to 1.0 IoU points above the
teacher, and the final difference, 2,500 teacher-labeled minus 252 hand-labeled images, is $-1.4$ $[-3.5, +0.5]$ IoU points. At the end of training, 2,500 against 252 images from the same teacher-labeled pool gives
$+1.9$ $[-0.6, +6.3]$ IoU points on WorldFloods, $+0.2$ $[-0.8, +1.1]$ on STURM-Flood and $+0.6$ $[-0.1, +1.1]$ on the
held-back events.

We therefore attribute the gain to the teacher pipeline as a whole, including its probability maps and additional images. The
number of images, their diversity and the source of their labels change together between the pools, and we do not claim
that any one of them, or the quality of the teacher's labels, causes the gain.

\emph{Labeler.} Replacing the teacher's hard labels with MNDWI labels on the same 2,500 images costs 2.1 to 4.9 IoU points when
the student learns from hard labels alone. Once the teacher's probability maps are added, the hard-label source
has little effect. The teacher's value thus lies in its probability maps as much as in its hard labels, in line with the
``dark knowledge'' that soft targets carry in distillation~\cite{hinton2015distilling}.

\emph{OFA side branches.} OFA gives no reliable gain over logit distillation for EfficientViT-B0, MobileNetV3-Small or
LightFormer, with 252 or 2,500 images: all differences lie between $-0.8$ and $+0.8$ IoU points, and none survives Holm
correction over the 39 OFA comparisons. We therefore use plain logit distillation in the proposed setup.

The quality filter and farthest-point sampling were tested only under the original experimental setup of
this study (training at $224\!\times\!224$ pixels, mIoU). These experiments were not repeated under the current setup, as noted in Table~\ref{tab:onechange}.
\end{newtext}

\begin{table*}[t]
\revnewfloat
\centering
\caption{\add{\textbf{One change at a time.} Each row compares two training settings and gives the
difference in water IoU per flood event, left setting minus right setting (in points, mean of four seeds per setting, with a 95\%
interval obtained by resampling whole events). Positive values favor the left setting. The settings differ in one factor, except in the first number-of-images row, the whole-pipeline row and the QAT row, where several factors change together. Main student model used, unless another model is named. Settings with 252 images use 50 epochs (750 updates) unless 7,800 updates are stated. The first block separates the
three factors that differ between 252 Sen1Floods11 and 252 GDACS images. The second tests each pipeline component. The last lists components tested only under the original experimental setup. $^\ast$Final point
of the learning-curve runs (Supplementary Section~\ref{S-sec:s_convergence}). This row changes only the number of images. Held-back results of these comparisons are
given in the text. The first number-of-images row uses a second training of the proposed setup whose learning rate decays after every update, like its 252-image setting, so on the public test sets it differs from Table~\ref{tab:main} by up to 0.3 IoU points.}}
\label{tab:onechange}
\scriptsize
\setlength{\tabcolsep}{3pt}
\renewcommand{\arraystretch}{1.12}
\begin{tabular}{@{}p{0.13\textwidth}p{0.40\textwidth}r@{\ }lr@{\ }lr@{\ }l@{}}
\toprule
Factor & Left setting $-$ right setting & \multicolumn{2}{c}{S11 test} & \multicolumn{2}{c}{WorldFloods test} & \multicolumn{2}{c}{STURM-Flood} \\
\midrule
\multicolumn{8}{@{}l}{\itshape What differs between 252 Sen1Floods11 and 252 GDACS training images} \\
\textbf{Label source} & teacher labels $-$ hand labels, on the same 252 S11 images & $+0.6$ & $[-0.1, +1.4]$ & $+0.7$ & $[-0.3, +1.7]$ & $0.0$ & $[-0.6, +0.7]$ \\
 & the same, both settings also learning from the teacher's maps & $-0.5$ & $[-1.7, +0.3]$ & $-0.4$ & $[-1.3, +0.3]$ & $+0.3$ & $[-0.1, +0.7]$ \\
\addlinespace
\textbf{Imagery} & 252 GDACS $-$ 252 S11 images, both with teacher labels & $-2.7$ & $[-7.1, +1.4]$ & $+1.5$ & $[-3.1, +8.2]$ & $-5.8$ & $[-9.5, -2.6]$ \\
 & the same, both settings also learning from the teacher's maps & $-3.2$ & $[-6.7, -0.2]$ & $+1.1$ & $[-0.3, +2.9]$ & $-2.6$ & $[-5.7, +0.1]$ \\
\addlinespace
\textbf{Image extent} & 2.6\,km $-$ 5.12\,km GDACS images, teacher labels & $-4.4$ & $[-8.1, +0.3]$ & $-2.1$ & $[-6.8, +1.8]$ & $+1.8$ & $[+0.1, +3.7]$ \\
\midrule
\multicolumn{8}{@{}l}{\itshape Components of the pipeline} \\
\textbf{Probability maps} & with $-$ without the teacher's maps, 252 hand-labeled images & $+2.7$ & $[+1.1, +5.0]$ & $+7.7$ & $[+3.0, +14.5]$ & $+1.0$ & $[-0.9, +2.9]$ \\
 & with $-$ without, 2,500 teacher-labeled images & $+1.2$ & $[-0.3, +3.0]$ & $+0.5$ & $[-0.9, +2.2]$ & $+1.5$ & $[+0.1, +3.1]$ \\
 & with $-$ without, 252 hand-labeled images, 7,800 updates & $+0.9$ & $[0.0, +2.0]$ & $+5.3$ & $[+2.1, +8.5]$ & $+3.8$ & $[+2.0, +5.8]$ \\
\addlinespace
\textbf{Number of images} & 2,500 teacher-labeled $-$ 252 hand-labeled, both with maps, 7,800 updates & $+0.7$ & $[-1.5, +3.0]$ & $+4.9$ & $[+0.2, +13.7]$ & $-0.6$ & $[-2.4, +1.2]$ \\
 & 2,500 $-$ 252 teacher-labeled GDACS images (the 252 are a subset of the 2,500), with maps$^\ast$ & $+0.8$ & $[-1.3, +2.8]$ & $+1.9$ & $[-0.6, +6.3]$ & $+0.2$ & $[-0.8, +1.1]$ \\
\addlinespace
\textbf{Whole pipeline} & teacher pipeline $-$ 252 hand labels without teacher, 7,800 updates & $+1.3$ & $[-1.6, +4.3]$ & $+10.1$ & $[+2.7, +20.7]$ & $+2.9$ & $[0.0, +6.2]$ \\
\addlinespace
\textbf{Labeler} & MNDWI labels $-$ teacher labels, 2,500 images & $-2.1$ & $[-7.4, +3.5]$ & $-4.9$ & $[-15.3, +2.5]$ & $-4.4$ & $[-7.9, -1.4]$ \\
 & the same, both settings also learning from the teacher's maps & $+1.2$ & $[+0.1, +2.6]$ & $+0.3$ & $[-0.7, +1.4]$ & $-0.1$ & $[-0.6, +0.3]$ \\
\addlinespace
\textbf{OFA branches} & OFA $-$ logit distillation, 2,500 images & $+0.2$ & $[-1.1, +1.2]$ & $+0.2$ & $[-0.2, +0.6]$ & $-0.6$ & $[-1.7, +0.1]$ \\
 & the same, MobileNetV3-Small & $-0.4$ & $[-1.1, +0.2]$ & $+0.2$ & $[-0.1, +0.5]$ & $+0.2$ & $[-0.1, +0.5]$ \\
 & the same, LightFormer-ResNet18 & $+0.8$ & $[+0.2, +1.8]$ & $+0.2$ & $[-0.3, +0.9]$ & $-0.3$ & $[-0.7, 0.0]$ \\
 & the same, 252 GDACS or 252 hand-labeled images & \multicolumn{6}{c}{all within $-0.8$ and $+0.8$} \\
\addlinespace
\textbf{Quantization format} & INT8 or FP16 $-$ FP32, on the edge device & \multicolumn{6}{c}{Table~\ref{tab:device}} \\
\addlinespace
\textbf{QAT} & QAT $-$ PTQ, not separated from calibration and export changes & \multicolumn{6}{p{0.30\textwidth}}{main student: PTQ fails in all four seeds, QAT passes. Tiny and micro students: PTQ suffices for 13 of 16 engines (Suppl.~Sec.~\ref{S-sec:s_deploy})} \\
\addlinespace
\textbf{Activation} & ReLU6 $-$ Hardswish, main student, FP32 on the edge device & $-2.6$ & $[-6.8, +0.1]$ & $-0.7$ & $[-1.9, +0.2]$ & $-0.6$ & $[-1.1, 0.0]$ \\
\midrule
\multicolumn{8}{@{}l}{\itshape Original experimental setup of this study ($224\!\times\!224$ training, 2.6\,km GDACS tiles, mIoU), not repeated} \\
\textbf{Quality filter} & with $-$ without the pre-filter, 2,500 images & \multicolumn{6}{p{0.30\textwidth}}{removing it costs 0.1 to 31 mIoU points, depending on the selector (Suppl.~Sec.~\ref{S-sec:s_curation})} \\
\textbf{Selection rule} & FPS $-$ random and three other selectors & \multicolumn{6}{p{0.30\textwidth}}{spread of 1.2 mIoU points on STURM-Flood, within the seed spread. FPS retained without claiming a gain} \\
\bottomrule
\end{tabular}
\end{table*}

\subsection{\texorpdfstring{\add{What the Teacher Adds to Each Student}}{What the Teacher Adds to Each Student}}
\label{sec:analysis}

\begin{oldtext}
\revst{\textbf{Submitted Section V, Effect of Training-Set Size and Distillation Objective.}
The experiments separate two effects. At small training-set sizes, access to soft teacher targets and the OFA auxiliary heads can improve sample efficiency. As the teacher-supervised pool grows, however, objective differences shrink while all objectives benefit from additional scene coverage. At the $N=2{,}500$ operating point, OFA and Logit-KD are not reliably separated on STURM-Flood, so we do not present OFA as the source of the final OOD performance.}

\revst{This is consistent with the role of the teacher in our pipeline: the student approximates the
same teacher output over an increasingly broad set of inputs. At larger training-set sizes,
adding supervised scenes has a larger observed effect than changing among the evaluated
output-space objectives. Supplementary diagnostics retain the gradient-path,
representation-alignment, and auxiliary-supervision analyses that support this interpretation,
but they are not required for the central claim.}
\end{oldtext}
\begin{deltext}
\paragraph*{\revst{Objective differences at the deployment budget}}
\revst{The gap that the objective opens below 252 scenes has closed by 2{,}500: the three
EfficientViT rows of Table~II span 0.004~mIoU on STURM and their
ordering is no longer monotone, with Logit-KD and OFA separated by less than either
one's seed spread. OFA stays above CE on every benchmark at both budgets, but the strict
CE\,$<$\,Logit-KD\,$<$\,OFA ordering does not hold throughout. Read together with
Table~I, where the same objective comparison spans 0.018, this suggests that the choice of distillation objective matters most when training data are limited, while its effect becomes smaller at the larger deployment budget
(Section~V). At $N=2{,}500$, the choice among the three objectives has little effect
on clean accuracy. We deploy the OFA configuration because it is the strongest one in the
scarce regime a new collection starts in, and because it is the one we carried through
quantization-aware training, not because it is measurably better here.}
\end{deltext}

\begin{delfloat}
\begin{figure*}[t]
\revdelfloat{FIG. 2}
\centering
\includegraphics[width=0.75\linewidth]{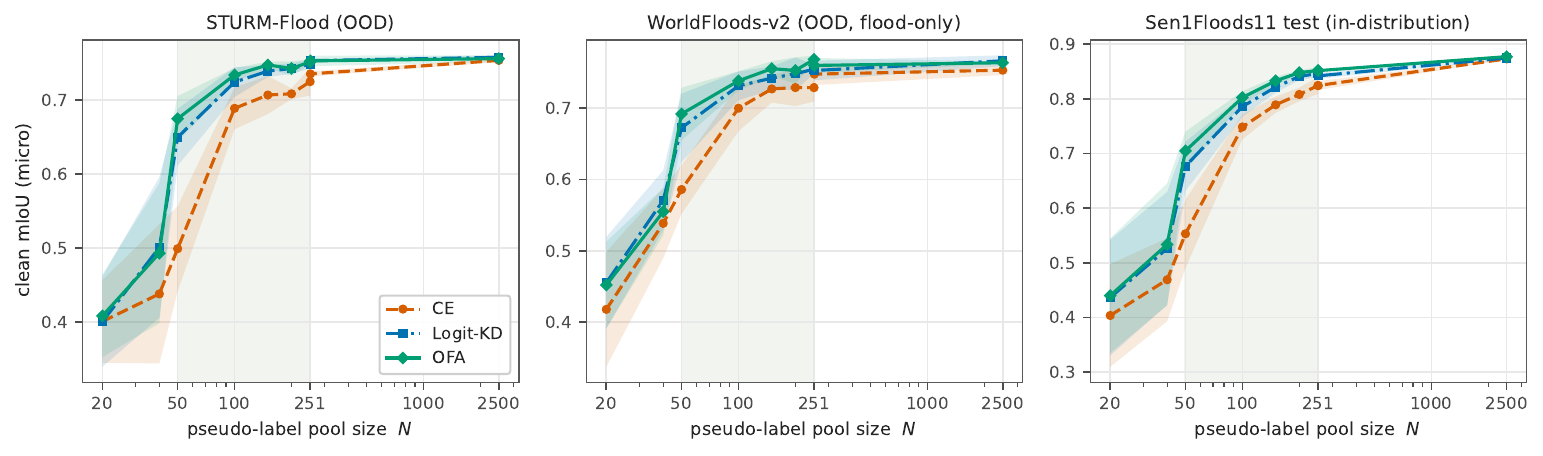}
{\footnotesize \revst{Submitted Fig.~2. \textbf{Effect of pool size and distillation objective.} Clean micro mIoU
  against pseudo-label pool size for the three objectives, on all three benchmarks.
  EfficientViT-B0, one recipe, one pool, four seeds ($\pm$1\,sd). The shaded band marks
  $50\!\leq\!N\!\leq\!251$, the range in which OFA's seed mean is ahead of Logit-KD on
  all three benchmarks; outside it the sign is not consistent. Per-seed win
  rates and the per-$N$ spans are in the released CSV.}\par}
\end{figure*}
\end{delfloat}

\begin{newtext}
Table~\ref{tab:main} lists all six students, from 1,266 parameters to 32.5\,M, with the proposed setup and with the 252
hand labels alone, each trained for 7,800 updates. Comparing its second and third blocks shows what the teacher adds to each
design. With the teacher pipeline, every student performs within 2 IoU points of the teacher on WorldFloods and STURM-Flood,
so compact students match the larger ones. The gain follows the shortfall on hand labels alone: on WorldFloods, the main student gains 10.1 IoU points and UNet-ResNet50 gains 1.2, the tiny student gains 2.3 on STURM-Flood, and the micro student,
already at the teacher's level on WorldFloods and STURM-Flood, gains nothing there. On the held-back events, the four pretrained students gain 3.9 to 5.7 IoU
points. All six students raise fewer false alarms with the proposed setup than without the teacher; the tiny student falls
from 11.2\% to 7.1\% of water-free windows. Below about 1,000 parameters, accuracy fails where the scenes differ most: on
STURM-Flood the 418-parameter per-pixel network is 17.9 IoU points below the teacher (Section~\ref{sec:index}). Trained on
the hand labels with its authors' setup instead of ours, LightFormer gains up to 6.3 IoU points. It reaches the main
student's level on STURM-Flood but remains 1.7 and 2.9 IoU points below it on WorldFloods and the held-back events
(Table~\ref{tab:main}). Supplementary Section~\ref{S-sec:s_backbones} repeats both comparisons with the same hard-label loss
in all setups, which gives similar gains, and separates the teacher's maps, which help every pretrained student but not the
two trained from scratch.
\end{newtext}

\begin{oldtext}
\revst{The 32.5\,M UNet is the more accurate model in distribution, with the best S11 score in
both tables, and the less accurate one on STURM. This is consistent with the
higher-capacity model fitting the teacher's output distribution more closely at this pool size
and transferring less well; the disagreement audit of Supplementary Section~S8 points the same
way. We do not claim the compact student is universally superior: on WorldFloods under pixel
pooling at a matched budget, the \emph{Sen1Floods11} UNet-Logit-KD configuration is the
strongest model in this paper (Table~I).}
\end{oldtext}

\subsection{\chghead{External Benchmark Interpretation}{Learned Models, a Water Index and Water-Free Ground}}
\label{sec:index}
\label{sec:twobench}

\begin{oldtext}
\revst{STURM-Flood and WorldFloods-v2 are used as external evaluation sets, not as independent proof of a general robustness mechanism. They differ in scene composition, aggregation sensitivity, product preprocessing and label conventions, and they partially overlap in Copernicus Emergency Management Service activations. We therefore report both pixel-pooled and per-scene summaries where appropriate and keep benchmark-specific conclusions separate.}

\revst{A fixed, parameter-free baseline provides a reference for interpreting external benchmark
performance. We evaluate MNDWI~\cite{xu2006modification},
$\mathrm{MNDWI}=(\mathrm{B3}-\mathrm{B11})/(\mathrm{B3}+\mathrm{B11})$, thresholded at zero,
through the same dataloaders, resizing, class mapping, and ignore masks as the learned models.
Under pixel pooling it reaches 0.747~mIoU on STURM-Flood and 0.763 on WorldFloods-v2, against
0.755 and 0.771 for the teacher and 0.756 and 0.764 for the N=2,500 student. Paired group-level
tests do not resolve either learned model from the rule on either clean external benchmark.}

\revst{This does \emph{not} establish equivalence, and pooled mIoU is only one aggregation. The index is
substantially weaker in distribution (0.789~mIoU and 0.650 water IoU on the Sen1Floods11 test
split, against 0.877 and 0.787 for the N=2,500 student and 0.897 and 0.822 for the teacher). Its
per-scene WorldFloods water IoU is 0.418 against the teacher's 0.504, while on STURM under
per-scene averaging it is ahead of both models. It is exactly invariant to a scene-wide
multiplicative gain because it is a normalized difference, and it loses 44--66\% of its mIoU
under the additive haze envelope of Section~IV-F, where the teacher loses 4--5\%.
Neither the choice of index nor the threshold transfers across datasets: the normalized difference water index (NDWI) is the stronger
rule in distribution and collapses on WorldFloods, and the optimum threshold differs between the
benchmarks (Supplementary Section~S4).}

\revst{Strong test-time performance of MNDWI does not make the foundation model redundant as a
supervision source. In the teacher-free hard-label control, replacing teacher-generated masks
with MNDWI masks reduces mIoU by approximately 0.038 on both Sen1Floods11 and STURM-Flood,
although WorldFloods-v2 does not show the same ordering (Supplementary Section~S4).}

\revst{Taken together, these results limit the interpretation of clean external accuracy. Under our
group-level tests, neither learned model is statistically distinguished from the fixed MNDWI
baseline on the two clean external datasets. We therefore use STURM-Flood and WorldFloods-v2 to
assess external generalization of the compact model, while treating synthetic perturbation and
spectral-index analyses as diagnostics.}
\end{oldtext}
\begin{newtext}
We evaluate MNDWI~\cite{xu2006modification}, $\mathrm{MNDWI}=(\mathrm{B3}-\mathrm{B11})/(\mathrm{B3}+\mathrm{B11})$,
thresholded at zero, through the same loaders, resizing and reference masks as the learned models with no training required.
Supplementary Section~\ref{S-sec:s_mndwi} adds nine further index variants: NDWI~\cite{mcfeeters1996use}, AWEI in its
shadow and no-shadow forms~\cite{feyisa2014automated} and WI2015~\cite{fisher2016comparing} with their standard thresholds,
and all five indices with a threshold tuned on the Sen1Floods11 training and validation images.
On the three test sets, MNDWI is competitive in water IoU. However, the main student is ahead of the index by 7.8 and 8.5 IoU
points on the Sen1Floods11 and WorldFloods test splits and level with it on STURM-Flood ($-0.3$). All three intervals include zero
(Fig.~\ref{fig:four}c).

The learned models perform reliably better in settings relevant to unattended use. On the 97 held-back events, the main student leads
by 11.5 $[8.6, 14.4]$ IoU points. Of the pixels it classifies as water, 84.5\% are water, compared with 57.6\% for the index. On the 200
water-free windows the index raises a false alarm on 57.0\% of windows, the teacher on 1.5\%, and the main and tiny
students on 5.4\% and 7.1\% (Table~\ref{tab:main}). The windows come from the official WorldFloods test split and the
held-back events come from its training and validation scenes. The two results therefore use separate samples. Both show the
index's main failure: predicting water where there is none. On 28 ordinary dry tiles, every learned model raises an alarm on about
one tile and the index on six. On 21 tiles with shadows or snow, every method frequently raises alarms (the teacher on 10, the main
student on 12 and the index on 16). MNDWI with threshold zero is not the strongest index: AWEI$_\text{sh}$ and WI2015 with
threshold zero are 1.4 to 8.5 IoU points better on all four test sets. Neither closes the gap: the best variant on the
held-back events, WI2015 with a tuned threshold, remains 4.2 $[2.5, 6.1]$ IoU points behind the main student. Every
variant raises a false alarm on 14.8\% to 57.0\% of the water-free windows
(Supplementary Table~\ref{S-tab:s_index}). Tuned thresholds do not transfer reliably between datasets (Supplementary
Section~\ref{S-sec:s_mndwi}), as earlier studies found for water-index thresholds~\cite{ji2009analysis,mateogarcia2021towards}.
Spatial context matters most where scenes differ. Trained on the same images with the same loss, the 418-parameter
per-pixel network and the 1,058-parameter CNN perform similarly on the Sen1Floods11 test split. The CNN uses a $5\!\times\!5$-pixel neighborhood and performs 16.3 IoU points better on
STURM-Flood. The tiny student is 2.3 IoU points below the index there and 8 to 11 points above it
on the other three sets. An onboard flood-mapping study found the same order on WorldFloods: per-pixel linear models matched
the best NDWI threshold, and small convolutional networks gained 6 to 8 IoU points~\cite{mateogarcia2021towards}.
\end{newtext}

\begin{deltext}
\revst{\textbf{Submitted Section IV-E, Label Efficiency and Scaling.}
The matched-budget experiment is intentionally conservative because both arms are capped at 252 training scenes. The practical advantage of teacher supervision is that the cap can be removed: once the task teacher has been trained, additional Sentinel-2 scenes can be labeled automatically and added without acquiring new manual annotations. Figure~4 and Fig.~5 show this scaling behavior under a fixed student architecture and training recipe.}

\revst{On STURM-Flood, teacher-supervised training is ahead of the direct Sen1Floods11 arm at the matched budgets and continues improving after the manually labeled split is exhausted. At $N=2{,}500$ the EfficientViT-B0 OFA student reaches 0.756~mIoU against the teacher's 0.755. The same operating point remains below the teacher on WorldFloods-v2, so scaling reduces but does not eliminate the teacher--student gap across domains. In distribution, water IoU rises from 0.743 at the matched budget to 0.787 at $N=2{,}500$, against 0.822 for the teacher. We therefore describe scaling as narrowing the gap rather than closing it. Figure~3 illustrates the same scaling effect qualitatively at $N=50$ and $N=2{,}500$.}

\revst{The choice of subset selector is secondary to this result. The pre-filter prevents pathological selections, whereas differences among reasonable ranking rules are small at the deployment budget and benchmark-dependent below it. The selector comparison is reported in Supplementary Section~S2.}
\end{deltext}

\begin{delfloat}
\begin{figure*}[t]
\revdelfloat{FIG. 3}
\centering
\includegraphics[width=0.95\linewidth]{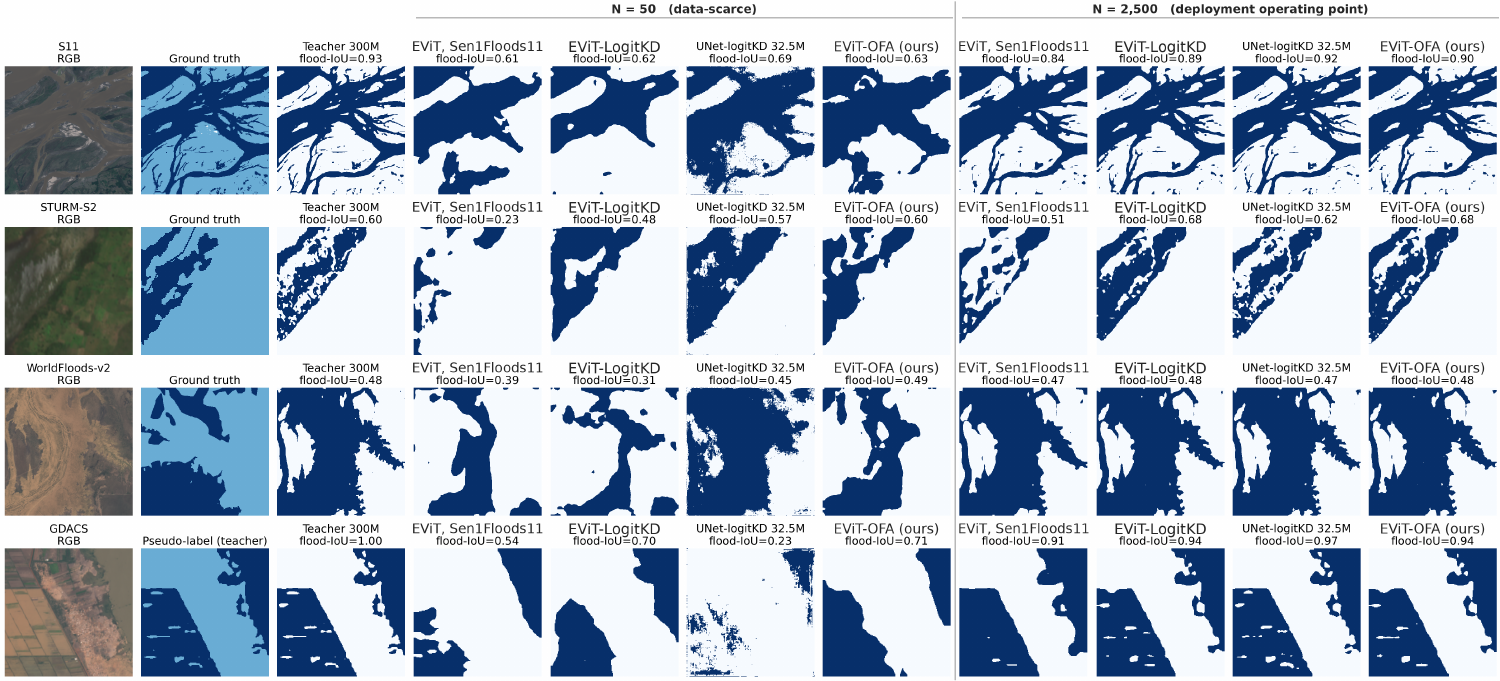}
{\footnotesize \revst{Submitted Fig.~3. \textbf{Four qualitative scenes at two training budgets.} Rows are the four
  evaluation sources (Sen1Floods11, STURM-Flood, WorldFloods-v2, and a held-out GDACS tile
  whose ``ground truth'' is the teacher's pseudo-label); RGB, reference and teacher are
  budget-independent and shown once. The four model columns repeat at $N\!=\!50$ and
  $N\!=\!2{,}500$, with per-panel water IoU; the Sen1Floods11 column caps at the split
  size, so its right-hand entry is the 251-image ladder point. At $N\!=\!50$ the
  failures differ by configuration; at the operating point all four converge onto the
  teacher's answer. Display scenes were selected for low cloud cover.}\par}
\end{figure*}

\begin{figure}[t]
\revdelfloat{FIG. 4}
\centering
\includegraphics[width=\linewidth]{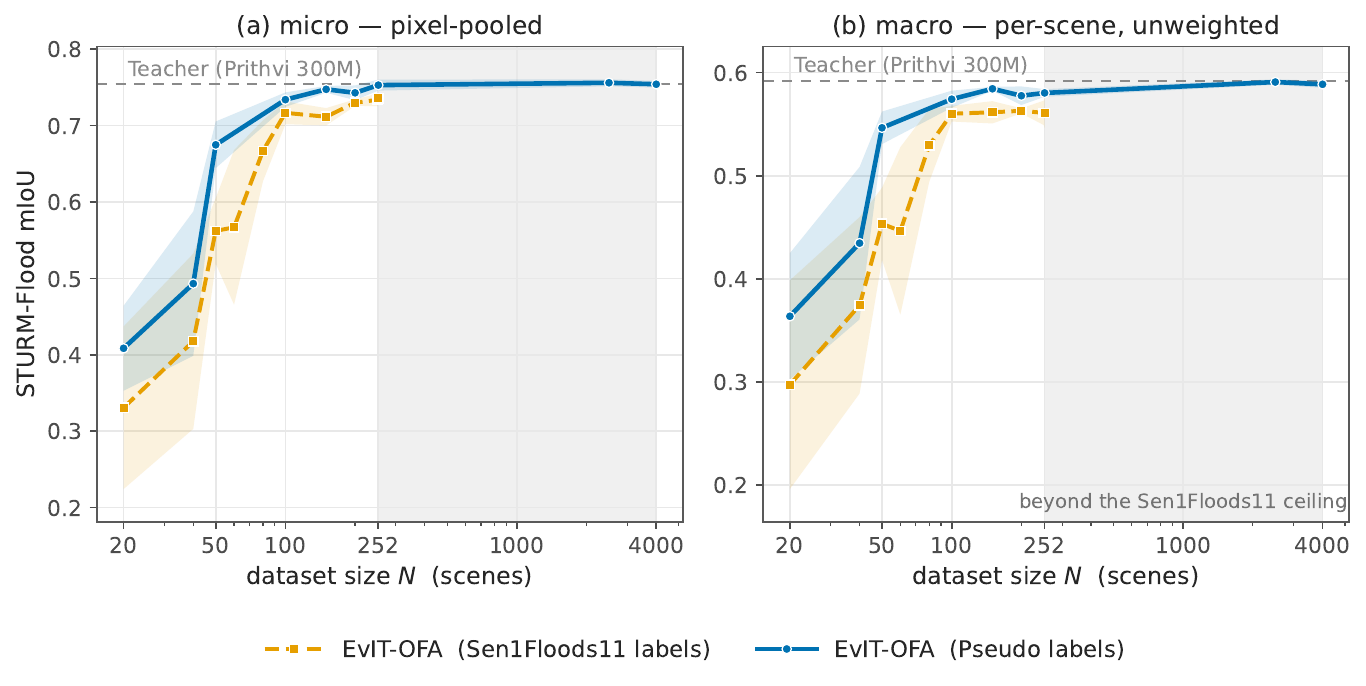}
{\footnotesize \revst{Submitted Fig.~4. \textbf{Matched-budget and scaling comparison on STURM-Flood.}
EfficientViT-B0 with OFA, 4 seeds, $\pm$1\,sd. \emph{Both} curves distill from the same
frozen teacher and differ only in the training scenes and in the source of the hard target:
$N$ Sen1Floods11 images with the human masks, or $N$ GDACS tiles with the teacher's
(Section~III-C). That teacher was fine-tuned on the same split and no GDACS imagery,
so both rest on the same annotation budget; the $x$-axis is training-set size, not
annotation cost. The pseudo-label configuration is ahead at every matched $N$ under both
aggregation rules and continues into the shaded region, where the Sen1Floods11
configuration cannot go because there are no more labeled images. The $N=4{,}000$ run is included only as a scaling diagnostic and shows saturation beyond the $N=2{,}500$ deployment operating point.}\par}
\end{figure}

\begin{figure}[t]
\revdelfloat{FIG. 5}
\centering
\includegraphics[width=\linewidth]{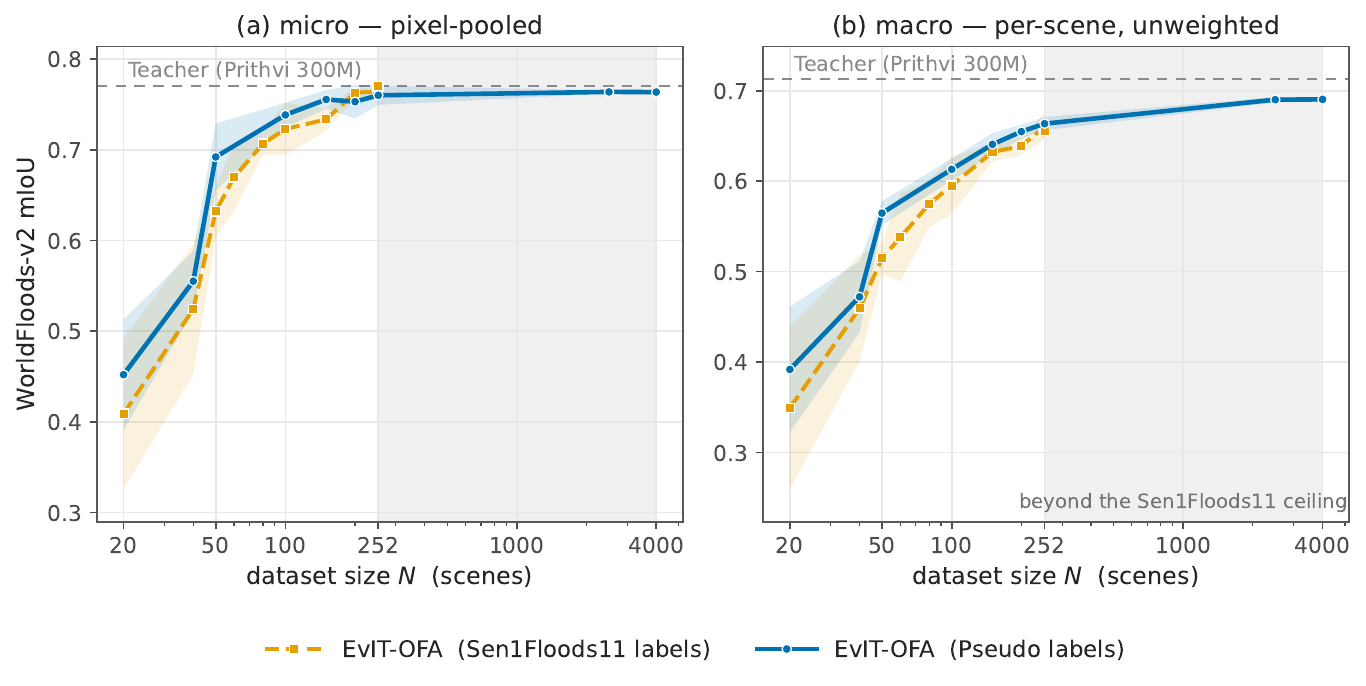}
{\footnotesize \revst{Submitted Fig.~5. \textbf{Matched-budget and scaling comparison as in Fig.~4 on WorldFloods-v2.}
  The shape replicates with one visible exception: the
  pseudo-label configuration leads at every matched $N$ under per-scene averaging, while
  under pixel pooling the Sen1Floods11 arm is $0.010$ ahead at $N\!=\!200$ and $252$
  (Section~IV-D). The architecture control is Supplementary Fig.~S4. The $N=4{,}000$ run is included only as a scaling diagnostic and shows saturation beyond the $N=2{,}500$ deployment operating point.}\par}
\end{figure}
\end{delfloat}

\begin{deltext}
\revst{\textbf{Submitted Section IV-F, Acquisition-Shift Diagnostics} (moved to Supplementary Section~S3 as diagnostics of the original study).
Synthetic perturbations are used here to diagnose model sensitivity, not to support a general robustness claim. Three axes are applied in reflectance space before the frozen normalization: a scene-wide multiplicative gain $g\in[0.8,1.3]$; an additive, wavelength-dependent haze offset of strength $h\in[0,0.20]$~\cite{makarau2014haze}; and a per-band gain that perturbs the visible and near-/short-wave-infrared groups differentially. Sensitivity is summarized as the worst-case mIoU drop over an axis, computed per seed and then averaged. Definitions and full results are in Supplementary Section~S3.}

\revst{Under the scalar-gain axis the teacher-supervised student is the flatter of the two arms, and this survives the geometry-matched control (all three objective pairs, $p\le2.5\times10^{-8}$ per flood map). The robustness test is a weak one, because the normalized-difference index cancels a scene-wide scalar. Therefore its worst-case drop on this axis is exactly zero, with no parameters, and the tested range overlaps the per-channel jitter both arms train on. The supported statement is brightness invariance under this diagnostic, not general radiometric robustness.}

\revst{The ordering does not carry over to the other two axes. Under per-band gain, with the visible and infrared groups perturbed differentially at magnitudes spanning documented Sentinel-2 Level-2A uncertainty ranges~\cite{gorrono2024framework}, the advantage disappears or reverses across objectives and benchmarks. Under haze the matched-budget advantage does not survive the geometry-matched control for any of the three objectives. Within the teacher-supervised arm, the worst-case STURM-Flood haze drop decreases from $0.081$ at $N\!=\!252$ to $0.028$ at $N\!=\!2{,}500$; because both budgets use the same training source, this comparison is not exposed to the geometry confound above. Training-pool diversity is not isolated from other differences between the datasets, including label source, imagery, geography, and acquisition geometry. We therefore do not attribute the observed effects causally to diversity.}
\end{deltext}

\subsection{Teacher Annotation Semantics}
\label{sec:inherit}

The student is trained from teacher supervision and therefore is not an independent observer of the teacher's systematic label errors. Sen1Floods11 defines surface water, and the fine-tuned teacher carries that convention into the \chg{pseudo-labeled}{teacher-labeled} pool. A pre-event audit shows that many tiles labelled as water by the teacher contain permanent or seasonal water rather than flooding from the target event. The student largely reproduces this behaviour: across four seeds, it agrees with the teacher on $96.1 \pm 1.0\%$ of these flagged baselines. \delx{At the same time, it predicts water on only $1.9 \pm 0.8\%$ of the baselines that the teacher left empty.}\add{These numbers come from students trained under the original 224-pixel experimental setup. We did not repeat this audit. Instead, we measure false alarms against independent references (Section~\ref{sec:index}).}

This agreement is not explained by overlap with the training set. Of the 2,942 flagged baselines, 961 are part of the \add{earlier} training subset and 1,981 are held out. Agreement is almost identical for both groups: $96.0 \pm 0.6\%$ for training tiles and $96.2 \pm 1.2\%$ for held-out tiles.

The result shows that the student also learns the teacher's definition of water. For an operational flood-increment product, permanent or seasonal water must therefore be removed using an additional reference-water or change-detection step.

\del{We report this audit mainly as a limitation and reproducibility check rather than as a separate contribution. }Increasing the amount of teacher-generated supervision also increases the amount of the teacher's annotation semantics transferred to the student.
The full audit is in Supplementary Section~\chg{S5}{\ref{S-sec:s_semantics}}.

\section{Deployment}
\label{sec:deployment}

\begin{deltext}
\revst{\textbf{Submitted Section VI-A, INT8 Deployment Accuracy.}
Table~III follows the deployment chain at $N=2{,}500$. EfficientViT-B0 is retrained with ReLU6 for TensorRT compatibility, then quantization-aware trained, exported and compiled. Every row is the mean over the same four training seeds. The first three rows use the workstation evaluator and the last two use the on-device inference path. The higher WorldFloods-v2 mIoU after QAT is consistent with a regularization effect from quantization-aware fine-tuning and is not used as a primary accuracy claim. The like-for-like quantization comparison is device float versus device INT8. Clean accuracy is the deployment criterion; efficiency is reported separately in Table~IV.}

\revst{Activation calibration and checkpoint selection use the Sen1Floods11 validation split. Thus the pipeline requires no \emph{new manual annotation}, but existing ground-truth validation data still enters model selection and calibration. No ground-truth label contributes a training gradient to the pseudo-labeled student.}
\end{deltext}

\begin{delfloat}
\begin{table}[t]
\revdelfloat{TABLE III}
\centering
{\footnotesize \revst{Submitted Table~III. \textbf{Accuracy through the deployment chain.} EfficientViT-B0 with OFA at the
  $N\!=\!2{,}500$ operating point; mean\,$\pm$\,std over the four training seeds
  (42, 64, 91, 107). Micro mIoU. The first three rows are scored with the workstation
  evaluator, the last two through the on-device inference path. The like-for-like
  quantization cost is device float $\rightarrow$ device INT8.}\par}
\small
\setlength{\tabcolsep}{4pt}
\resizebox{\columnwidth}{!}{%
\begin{tabular}{llcc}
\toprule
Stage & Labels & STURM-Flood & WorldFloods-v2 \\
\midrule
EfficientViT-B0, OFA        & Sen1Floods11, 252 & $0.7357 \pm .0063$ & $0.7701 \pm .0077$ \\
EfficientViT-B0, OFA        & pseudo, 2{,}500   & $0.7561 \pm .0046$ & $0.7639 \pm .0018$ \\
\quad $+$ ReLU6             & pseudo, 2{,}500   & $0.7481 \pm .0055$ & $0.7714 \pm .0119$ \\
\midrule
\quad $+$ QAT, float export (device) & pseudo, 2{,}500 & $0.7557 \pm .0062$ & $0.7861 \pm .0105$ \\
\quad \emph{$+$ QAT, INT8 engine (deployed)} & pseudo, 2{,}500 & $0.7448 \pm .0236$ & $0.7702 \pm .0214$ \\
\midrule
\multicolumn{2}{l}{\itshape Quantization cost (device float $\rightarrow$ INT8)} & $-0.0110$ & $-0.0158$ \\
\bottomrule
\end{tabular}}
\end{table}

\begin{table*}[t]
\revdelfloat{TABLE IV}
\centering
{\footnotesize \revst{Submitted Table~IV. On-device efficiency on Jetson Xavier NX: GPU kernel time (\texttt{trtexec},
  200 iterations, 20 warm-up; no I/O overhead), for \emph{one representative compiled engine
  per precision}. EfficientViT-B0 OFA, ReLU6+QAT. Engine size = compiled TensorRT binary on disk.
  Power = GPU mean/max over the run. FPS is the throughput reported by
  \texttt{trtexec} over total wall time; because successive inferences can overlap, it need not
  equal $1000/$mean GPU latency. Italics mark the deployed configuration; bold marks the best
  value in each column. FP16 is the fastest precision here, INT8 the smallest. These are
  hardware-build measurements, not training-seed statistics; the accuracy of the chain over the
  four training seeds is Table~III. (The FP32 mean power should read 3.25\,W.)}\par}
\small
\setlength{\tabcolsep}{4pt}
\begin{tabular}{lrrrrrr}
\toprule
Precision & Latency\,mean & Latency\,p99 & FPS & Engine & GPU\,power & Speedup \\
 & (ms\,$\downarrow$) & (ms\,$\downarrow$) & ($\uparrow$) & (MB\,$\downarrow$) & (W\,mean/max) & vs FP32 \\
\midrule
\multicolumn{7}{l}{\itshape Teacher (300\,M) --- not benchmarked on device} \\
\midrule
FP32 & 7.91 & 8.05 & 134.1 & 3.4 & 3.95\,/\,7.67 & 1.00$\times$ \\
FP16 & \textbf{5.20} & \textbf{5.28} & \textbf{214.6} & 2.4 & 4.21\,/\,7.26 & \textbf{1.52}$\times$ \\
\emph{INT8 (QAT, deployed)} & 5.57 & 5.60 & 179.2 & \textbf{1.5} & 3.42\,/\,\textbf{5.83} & 1.42$\times$ \\
\bottomrule
\end{tabular}
\end{table*}
\end{delfloat}

\subsection{On-Device Benchmark (Jetson Xavier NX)}
\label{sec:jetson}

\begin{oldtext}
\revst{We use the NVIDIA Jetson Xavier NX (JetPack~5.0, TensorRT~8.4) as an accessible embedded GPU
testbed. Table~IV reports timing, engine size and power for the OFA,
ReLU6-from-scratch, 2{,}500-scene QAT model. These are properties of one compiled engine rather
than of the trained weights, so they are hardware-build measurements and not four-seed training
statistics; accuracy over the four training seeds is in Table~III. Latency is
measured with \texttt{trtexec} on one representative compiled engine per precision: 200 engine
executions with 20 warm-up iterations at $512\!\times\!512$, batch size 1, timing GPU kernel
time only (no Python, host$\leftrightarrow$device transfers, or I/O).}

\paragraph*{\revst{Accuracy and export validation}}
\revst{Each compiled engine is checked against its corresponding float export before deployment.
Evaluated through the same on-device inference path, the INT8 engine costs $0.011$~mIoU on
STURM-Flood and $0.016$ on WorldFloods-v2 relative to the float export, over the four training
seeds (Table~III). TensorRT build settings, calibration choices and the
export checks are in Supplementary Section~S7.}

\paragraph*{\revst{Deployment characteristics}}
\revst{FP16 is the fastest precision on this platform at $5.20$\,ms and $214.6$\,FPS; the deployed
INT8 engine reaches $5.57$\,ms and $179.2$\,FPS. INT8's advantage is footprint and peak power:
a 1.5\,MB engine against $2.4$\,MB for FP16 and $3.4$\,MB for FP32, and $5.83$\,W peak GPU
power against $7.26$ and $7.67$\,W. The profiled INT8 engines use 13.9--14.2\,MB of runtime
device memory (1.8\,MB persistent weights, 4.0\,MB scratch, 8.1--8.4\,MB activations at
$1\!\times\!6\!\times\!512\!\times\!512$). Host-to-device and device-to-host transfers add
0.36--0.39\,ms on top of the kernel time.}
\end{oldtext}
\begin{newtext}
We use the NVIDIA Jetson Xavier NX edge device (TensorRT~8.4, batch size one, $512\!\times\!512$ input) to compare the
main, tiny and micro students in their number formats. Each is evaluated over four training seeds through the device
inference path on the three public test sets and the water-free windows. Table~\ref{tab:device} gives mean and p99
inference time, throughput, engine size, power and speedup against FP32, together with the change in water IoU against
the FP32 engine of the same seed. Inference runs on the device's integrated accelerator, and the measured time excludes host--device copies.
The combined power use of the processing units is measured during steady-state inference.

\emph{Main student.} Its mixed INT8/FP16 engine, built from the ReLU6 student (Section~\ref{sec:qat}), needs
4.94\,ms per image, giving a $1.46\times$ speedup over its FP32 engine. The engine is 1.59\,MB and consumes 5.7 rather than 7.8\,W. Against
that FP32 engine it changes water IoU by $+2.0$ $[+0.3, +3.8]$ points on the Sen1Floods11 test split, $+0.9$
$[-0.2, +2.3]$ on WorldFloods, $-1.7$ $[-4.3, +0.3]$ on STURM-Flood and $+0.4$ $[-0.4, +1.2]$ on the held-back events.
It raises more false alarms, on 9.6\% rather than 4.8\% of the water-free windows. Against the FP32 engine of the Hardswish
student of Table~\ref{tab:main}, the deployed engine is $-2.2$ $[-5.0, -0.2]$ IoU points on STURM-Flood and within 0.6 points
on the other two public test sets. The Hardswish student of Table~\ref{tab:main} gains nothing from INT8 on this device:
its mixed engine needs 12.3\,ms, more than its FP32 engine (9.4\,ms), because of format conversions around the quantized
layers (Supplementary Section~\ref{S-sec:s_deploy}).

\emph{Students without attention.} FP16 costs no accuracy (at most 0.1 IoU points). Mixed INT8 runs at the speed of FP16 with
12 to 18\% less power and loses at most 1.2 IoU points. False alarms change by less than 1 percentage point. Full INT8, in which every tensor between layers is INT8, meets the format requirement of
integer-only accelerators. The tiny student loses 0.0, 0.1 and 0.5 IoU points on the three public test sets and runs
at its FP32 speed (1.76 against 1.75\,ms). The micro student is 2.3 times slower than in FP32. Full INT8
provides no speed gain on this device because the export pads narrow layers to wider channels and upsamples with dense transposed
convolutions. These costs are specific to the export and the device rather than to integer arithmetic (Supplementary Section~\ref{S-sec:s_deploy}).

\emph{Choice of number format.} On this device the mixed INT8/FP16 engine is the fastest and smallest engine we measured for the main student. For the
students without attention, mixed INT8 saves power at the speed of FP16, and full INT8 retains the tiny student's accuracy
on the tested device and provides a candidate export for integer-only targets. Three of the 16 INT8 engines of the tiny and micro students needed corrected
quantizer placement or additional QAT before passing the acceptance rule on the Sen1Floods11 validation split. Acceptance
on this split does not guarantee the same accuracy-loss limit for every test set and seed.
\end{newtext}

\begin{table*}[t]
\revnewfloat
\centering
\caption{\add{\textbf{Number formats on the edge device.} The deployed student models on the edge device (batch size one, $512\!\times\!512$ input, mean of four seeds). Latency: inference time per image on the device's integrated accelerator,
excluding host--device copies. FPS: images per second. Engine: size of the TensorRT engine file, including kernels and
metadata. Power:
mean and largest sample on the combined processing units  during steady inference. Speedup: relative to the FP32
engine of the same seed. Water IoU per event: the FP32 rows (italics) give the value on the device, the other rows the
change against it, with 95\% intervals in Supplementary Table~\ref{S-tab:s_device}. False alarms: 200 water-free
WorldFloods windows with more than 1\% predicted water. The main student is deployed with ReLU6 (Section~\ref{sec:qat}). The
Hardswish rows correspond to the student in Table~\ref{tab:main}. Shaded: the two deployed configurations, \colorbox{useGPU}{for edge devices that support mixed-precision computation} and \colorbox{useINT}{for integer-only hardware}.}}
\label{tab:device}
\footnotesize
\setlength{\tabcolsep}{3.4pt}
\begin{tabular}{llrrrrcrrrrr}
\toprule
 & & \multicolumn{6}{c}{Efficiency} & \multicolumn{3}{c}{Water IoU per event (points)} & False \\
\cmidrule(lr){3-8}\cmidrule(lr){9-11}
Student & Format & Latency & p99 & FPS & Engine & Power, mean/max & Speedup & S11 test & WorldFloods & STURM-Flood & alarms \\
 & & (ms) & (ms) & & (MB) & (W) & & & test & & (\%) \\
\midrule
\csname @@input\endcsname tables/device_benchmark.tex 
\bottomrule
\end{tabular}
\end{table*}

\del{We did not benchmark the 300\,M teacher on the Xavier NX, so the premise that it does not
fit an embedded budget is argued from parameter count rather than measured here.
UNet-ResNet50 would also run on this device. }The compact \chg{student's}{students'} advantage on this
platform is \chg{therefore engine size, latency and power characteristics rather than basic execution
feasibility}{latency, engine size and power}, and the \chg{stronger memory argument concerns}{memory argument is strongest on} tighter on-board and NPU/FPGA
targets\del{ that we did not evaluate} (Section~\ref{sec:limitations})\add{. For integer-only targets, the full-INT8 tiny student is
the relevant result}.

\begin{figure*}[t]
\revnewfloat
\centering
\includegraphics[width=0.81\linewidth]{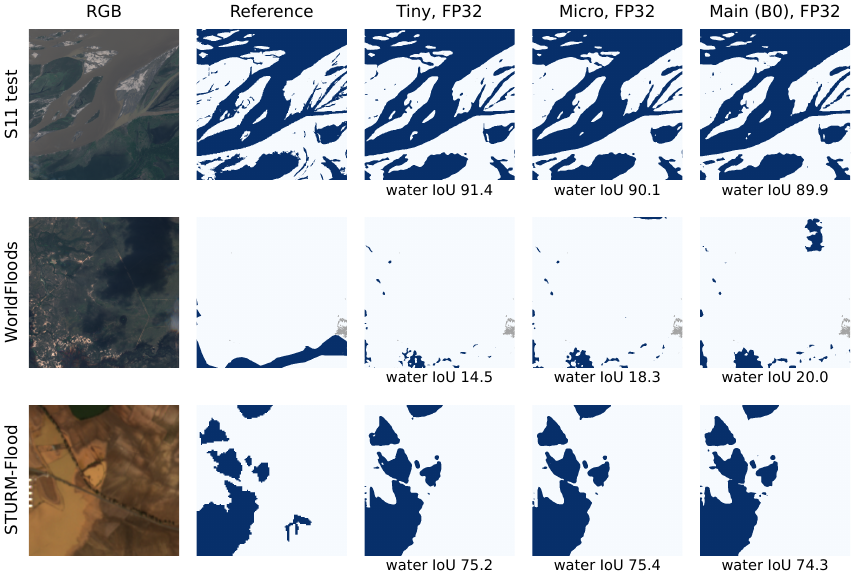}
\caption{\add{\textbf{Example predictions.} Predictions of the main, tiny and micro students of Table~\ref{tab:main} (one training seed, FP32 engines on the Jetson) on one scene from each public
test set, selected before prediction. Dark blue indicates water, and gray marks pixels excluded from scoring. Each score is the water IoU of that
scene. The WorldFloods scene is a hard case, far below that set's average.}}
\label{fig:qual_models}
\end{figure*}

\begin{newtext}
\paragraph*{Examples}
Fig.~\ref{fig:qual_models} shows the main, tiny and micro students of Table~\ref{tab:main} on one scene from each public test set, with the FP32 engines
on the device. The scenes were selected before these predictions were made. They illustrate
error types rather than estimate accuracy. Supplementary Figs.~\ref{S-fig:qual_tiny_precision}--\ref{S-fig:qual_b0_precision}
compare number formats on the same scenes using the same prepared inputs. These differences include the complete
deployment chain, such as QAT, rather than rounding alone.
\end{newtext}

\section{\chghead{Limitations}{Discussion}}
\label{sec:limitations}

\begin{newtext}
\paragraph*{What decides the teacher's value}
With teacher labels and probability maps on 2,500 images, every student from 1,266 to 32.5\,M parameters reaches the
teacher's level on WorldFloods and STURM-Flood. The teacher therefore adds most to a student that falls short when trained on
252 hand labels alone, such as EfficientViT-B0. For the tiny student, it mainly reduces false alarms. The benefit of the larger
pool depends on student capacity. Compared with 252 hand-labeled images with probability maps, the 2,500 teacher-labeled images improve
every student with at least 12,230 parameters by 1.5 to 3.7 IoU points on the held-back events. They do not improve the 1,266-parameter
micro student. 

\paragraph*{Which student to deploy}
Compact students match larger ones outside the training distribution, so small ones can be deployed. On the edge
device, the main student runs as a mixed INT8/FP16 engine at 4.94\,ms per $512\!\times\!512$ image, at a cost of 2.2
IoU points on STURM-Flood against the FP32 student of Table~\ref{tab:main}. On
the tested Jetson, the tiny student in full INT8 loses at most 0.5 IoU points against its FP32 version, which stays within
2 points of the teacher.

\paragraph*{When is a learned model worth using instead of a water index?}
A fixed MNDWI threshold costs nothing and matches the learned models in water IoU on small, curated test sets. Its
weakness becomes apparent across many varied events and on dry land, where it falsely predicts water. A learned model
justifies its cost when an unattended system must limit false alarms or work across many events without
event-specific threshold tuning. Tuned thresholds do not transfer reliably between datasets. A student with 12,230 parameters already
provides this benefit.

\paragraph*{Water-free ground}
The pre-filter removes images that the teacher predicts as fully dry from training. Still, all six students trained with
the proposed setup raise fewer false alarms on water-free ground than the same design trained on the hand labels alone,
although 89 of the 252 hand-labeled images contain less than 1\% water.
All methods, including the teacher, fail on shadows and snow. MNDWI has the form of a snow index, and shadows darken the
infrared bands as water does. Unattended use needs a shadow and snow mask or broader negative training data.

\paragraph*{Hardware dependence}
The speed results apply to this device and toolchain. INT8 provides a speed benefit only when the model and its export
support efficient execution. This holds for the main student with ReLU6 but not with Hardswish or the padded full-INT8 exports. An INT8
tensor format also does not establish that every internal operation uses integer arithmetic. We did not
benchmark integer-only accelerators, FPGAs or microcontrollers, so those targets need their own measurements of speed and
memory.

\paragraph*{Surface water is not flood change}
Sen1Floods11 defines surface water, and the teacher and its students reproduce that convention (Section~\ref{sec:inherit}). An
operational flood product needs a reference-water or change-detection step, with its own error analysis, which this paper
does not measure.

\end{newtext}
\paragraph*{\add{Limitations}}
\textbf{Scope of teacher, task and hardware.} The study uses one geospatial foundation-model teacher, \chg{one primary compact student}{one training setup for all students (only LightFormer was also trained with its authors' setup)}, binary Sentinel-2 segmentation and one embedded deployment platform. The 300\,M teacher was not benchmarked on the Xavier NX\chg{, and UNet-ResNet50 would also execute on that device. The deployment result therefore establishes a small measured engine footprint, latency and power profile, not universal infeasibility of larger models}{. Its encoder alone takes about 1.2\,GB in FP32~\cite{du2025onorbit}, against 1.59\,MB for the deployed engine. UNet-ResNet50 would also execute on that device, so the result is a small measured footprint, latency and power profile}.

\textbf{Teacher agreement is not independent validation.} The student is trained from the teacher's output and reproduces its systematic annotation conventions (Section~\ref{sec:inherit}), so agreement between them cannot be used as evidence that either is correct. All accuracy claims here rest on held-out ground truth, and the teacher-independent checks we do report are spectral proxies rather than annotations.

\begin{oldtext}
\revst{\textbf{Teacher semantics and dry-scene operation.} Sen1Floods11 targets surface water, and the student largely reproduces this convention. An operational flood-increment product therefore needs a reference-water or temporal change component. The pseudo-label pre-filter also removes fully dry scenes, while both external benchmarks are centered on flood extents; false alarms on ordinary dry imagery are consequently not characterized well enough for unattended event detection.}

\revst{\textbf{Benchmark-guided development.} Method choices were refined using the same public benchmarks reported in the paper. Four training seeds, event-level tests, contamination auditing and multiplicity correction reduce but do not eliminate this risk. A future evaluation should reserve a genuinely untouched event set.}
\end{oldtext}
\begin{newtext}
\textbf{Limits of the claims.} We restrict the interpretation in two ways. The downloaded images differ from the Sen1Floods11 images in scene content, location and date. We do not claim that a teacher-labeled image is worth more than a hand-labeled one. No experiment isolates geographic diversity as a cause. Because a fixed MNDWI threshold is competitive on the small curated test sets, we state where a learned model is better than this spectral rule and where it is not.

\textbf{Test sets.} The official test splits of Sen1Floods11 and WorldFloods hold only 10 and 11 events, so their
intervals are wide. The 97 held-back events provide narrower intervals. Model selection used only the Sen1Floods11 validation split,
whose images share events with the training and test splits.
\end{newtext}

\section{Conclusion}

\begin{oldtext}
\revst{We presented a foundation-model-to-edge distillation pipeline for Sentinel-2 flood segmentation. A 300\,M-parameter Prithvi-EO-2.0 teacher, adapted from 252 manually labeled Sen1Floods11 training scenes, supervises additional unlabeled flood-event imagery and trains a 0.7\,M-parameter EfficientViT-B0 student. At the same 252-scene budget, teacher-supervised training is competitive with direct use of the human-labeled split and raises STURM-Flood performance in all tested configurations, although a geometry-matched control shows that the difference cannot be attributed to label source alone.}

\revst{The main benefit of the teacher route is therefore scale. Once the task teacher exists, the student training set can grow beyond the fixed manual annotation set without acquiring new manual labels. At 2,500 teacher-supervised scenes the student narrows the in-distribution water-IoU gap to the teacher (0.787 against 0.822) and matches its STURM-Flood mIoU under our evaluation protocol, while still trailing it on WorldFloods-v2. A competitive fixed-threshold MNDWI baseline bounds the interpretation of clean external accuracy: these benchmarks support external generalization of the compact model, not a general claim that learned models dominate spectral indices.}

\revst{After ReLU6 retraining and quantization-aware training, the student compiles to a 1.5\,MB
INT8 TensorRT engine and runs on a Jetson Xavier NX at 5.57\,ms of GPU compute per
$512\!\times\!512$ image, with approximately 14\,MB of runtime device memory. These measurements demonstrate a route from a large
geospatial foundation model to a compact embedded segmenter: a fixed manual annotation budget is
extended through teacher supervision to a larger training set, while most of the teacher's task
performance is retained in a model roughly $430\!\times$ smaller by parameter count.}
\end{oldtext}
\begin{newtext}
We distilled a 300\,M-parameter Prithvi-EO-2.0 teacher, fine-tuned on 252 hand-labeled Sen1Floods11 images, into compact
surface-water segmenters for flood mapping. On the same images, teacher labels achieve human-level annotation quality for student training. The
teacher's probability maps and additional images labeled at no annotation cost enable a 0.7\,M student model to perform similarly to the teacher in IoU on WorldFloods, STURM-Flood and 97 held-back events. On the held-back events, performance is 5.7 IoU points above training on hand labels alone. Every
student from 1,266 parameters to 32.5\,M performs within 2 IoU points of the teacher on WorldFloods and STURM-Flood. Learned
models perform reliably better than a fixed water index across many varied events and on water-free ground. Every model was tested on four
test sets from three public datasets and on two dry-scene sets introduced here. These claims therefore rest on more than one benchmark.

On a Jetson Xavier NX edge device, the main student runs as a mixed INT8/FP16 engine at 4.94\,ms per
$512\!\times\!512$ image. A 12,230-parameter student, about $24{,}500\!\times$ smaller than the teacher, runs in full
INT8 with accuracy within 0.5 IoU points of its FP32 version. A fixed manual annotation budget can thus be extended through a foundation-model teacher into compact
models that run on embedded hardware. Future work should reduce false alarms on shadows and snow with a dedicated mask or broader negative
training data, and add a reference-water or change-detection step for operational flood products. The full-INT8 students also
remain to be measured on integer-only accelerators.
\end{newtext}

\section*{Data, Code, and Ethics Statement}

All satellite imagery used in this work is Copernicus Sentinel-2 data, freely
available under the Copernicus open data license. The GDACS event catalog is
publicly accessible, and imagery was retrieved via Google Earth Engine in
accordance with its terms of service. Sen1Floods11, STURM-Flood, and
WorldFloods-v2 are published open datasets and were used under their respective
licenses.

Code for dataset construction and curation, training, evaluation, \chg{robustness
experiments}{the held-back and water-free test sets}, and deployment is available in the
repository at \url{https://github.com/sycz00/FloodDistill}. The repository
also provides the curated \chg{pseudo-label}{teacher-label} pool identifiers and curation index,
\add{the list of the 200 water-free windows, the 49 dry tiles with their screening record,}
experiment configurations, exported ONNX models, and \chg{TensorRT engine-build}{engine-build}
configuration required to reproduce the reported pipeline. The teacher is the
unmodified public Prithvi-EO-2.0-300M-TL Sen1Floods11 checkpoint. The
teacher-semantics audit in Section~\ref{sec:inherit} and Supplementary
Section~\chg{S5}{\ref{S-sec:s_semantics}} documents which systematic annotation conventions are retained by
the distilled student.

\section*{Acknowledgment}
Generative AI tools were used for language editing and grammatical refinement. Anthropic Claude was also used to assist with code review and minor modifications to the plotting and experimental result-collection code. All modifications and experimental results were manually reviewed and verified by the authors.

\appendices

\section{Implementation Details}
\label{app:impl}

This appendix gives the preprocessing and normalization pipeline that maps every
dataset into a common reference space, and the training hyperparameters.

\subsection*{Data Preprocessing and Cross-Dataset Normalization}

All training data (Sen1Floods11 and the GDACS \chg{pseudo-label}{teacher-label} pool) and all evaluation
data are Sentinel-2, but they do not arrive in a common numerical range, and the single
most important preprocessing principle in this work is that \emph{the normalization
applied at evaluation time must reproduce the normalization seen during training}. A
model trained on Sen1Floods11 learns decision boundaries at specific positions in a
normalized feature space; if an OOD dataset is mapped into a different space, those
boundaries no longer apply and the apparent ``domain gap'' is partly an artifact of
mismatched scaling.

\paragraph*{Anchor: Sen1Floods11 (S11)}
Sen1Floods11 is Level-1C (top-of-atmosphere reflectance). We use six bands
(B2, B3, B4, B8A, B11, B12); note B8A (narrow NIR, 865\,nm), not B8. Raw digital numbers (DNs) are converted to reflectance via $r = \mathrm{DN}\times 10^{-4}$ and then
channel-standardized, $\hat r = (r - \mu)/\sigma$, with per-band $\mu,\sigma$ computed
on the S11 training split:
\[
\begin{aligned}
\mu    &= [0.1413,\, 0.1380,\, 0.1235,\, 0.3090,\, 0.2045,\, 0.1191],\\
\sigma &= [0.0741,\, 0.0737,\, 0.0869,\, 0.1180,\, 0.0977,\, 0.0766].
\end{aligned}
\]
These constants define the reference space; every other dataset is mapped into it.

\paragraph*{Input resolution}\ifrevmarked\leavevmode\fi 
\begin{oldtext}
\revst{The native chip sizes differ across datasets (S11 $512\!\times\!512$, GDACS
$264\!\times\!264$, STURM $128\!\times\!128$, WorldFloods $512\!\times\!512$), so we standardise
on two working sizes. \emph{Training} resizes every chip to $224\!\times\!224$
(bilinear for images, nearest-neighbor for masks; a resize, \emph{not} a random
crop). Unless otherwise stated, the main experiments use the same augmentation: dihedral
flips and rotations plus multiplicative jitter in $[0.80,1.20]$, applied consistently to the images. For Logit-KD and OFA, the frozen teacher is evaluated on the transformed $224\!\times\!224$ image during training to produce the corresponding soft target (Section~IV-B). All students are optimized at this size. \emph{Benchmark evaluation} is performed at $512\!\times\!512$ unless otherwise stated: S11 is used
at its native $512$, WorldFloods at its native $512$, and STURM's native
$128\!\times\!128$ chips are upsampled to $512$ (images bilinearly, label masks with
nearest-neighbor interpolation) so every model sees a
consistent evaluation resolution (both EfficientViT-B0 and UNet require the side length
to be a multiple of their patch/stride, which $512$ satisfies).}
\end{oldtext}
\begin{newtext}
All training and evaluation use $512\!\times\!512$ inputs. Sen1Floods11, the GDACS images and WorldFloods are used on their
native 10\,m grid. STURM's native $128\!\times\!128$ chips are upsampled to $512$, using bilinear interpolation for images and
nearest-neighbor interpolation for label masks. Apart from LightFormer with its authors' setup, no image is resized for training and every setting uses the same augmentation: dihedral flips and rotations plus multiplicative jitter in
$[0.80,1.20]$. The frozen teacher is evaluated on the transformed image to produce the soft target.
\end{newtext}

\paragraph*{Effect of the input size}\ifrevmarked\leavevmode\fi 
\begin{oldtext}
\revst{Training tensors are resized to $224\!\times\!224$. Because the native chip extents differ
across datasets, this implies dataset-dependent effective spatial sampling and extent: a
Sen1Floods11 chip covers $4.63\times5.12$\,km and reaches $\approx$23\,m effective sampling,
whereas a GDACS tile covers $\approx$2.60\,km and reaches $\approx$11.7\,m. At evaluation,
$512$ is native sampling for Sen1Floods11 and WorldFloods-v2 (10.0\,m) but a $4\times$ upsample
for STURM, whose $128^2$ chips cover 1.28\,km and therefore reach 2.50\,m effective sampling.
The two OOD benchmarks are consequently not scored at a common physical scale, and no result
that pools them should be read as one; Section~IV-C reports the STURM evaluation-scale
control. The two training arms of Table~I likewise differ in geometry as well as
in label source; Section~IV-C reports the geometry-matched control.}

\revst{Because the students are fully convolutional~\cite{long2015fully} and
EfficientViT's linear attention is likewise resolution-agnostic, inference is not
restricted to the $224\times224$ training resolution, and the models produce dense
predictions for $512\times512$ inputs without architectural modification; evaluation
uses frozen BN statistics, and the receptive fields cover both scales. We nevertheless
treat the change in spatial dimensions as a potential distribution shift rather than
assuming exact resolution invariance. Training at $224$ keeps the
$\sim$80-configuration ablation program computationally tractable. We fix evaluation at
$512\!\times\!512$ as a common study protocol and did not sweep the evaluation size
systematically.}
\end{oldtext}
\begin{newtext}
Because the GDACS images now cover the same 5.12\,km as a Sen1Floods11 chip and every model trains on native 10\,m
pixels, the training sources no longer differ in extent or effective sampling. STURM-Flood remains a $4\times$ upsample
(2.50\,m effective sampling), so it is not scored at the same physical scale as the other sets. Supplementary
Section~\ref{S-sec:s_harmonisation} reports a control on the native $128\!\times\!128$ grid.
\end{newtext}

\paragraph*{RGB visualization}
All qualitative figures show the true-color image B4/B3/B2 (red/green/blue)
of the six-band input. Because the tensors are channel-standardized, each displayed
image is rescaled with a per-image min--max stretch to $[0,1]$; this stretch is for
display only and plays no role in training or evaluation.

\paragraph*{On-device evaluation}
The evaluation on Jetson reuses these
loaders, normalization constants, and benchmark masks at $512^2$, so the \chg{TRT}{device} engines see the same
preprocessing as the workstation evaluation. \chgx{Differences between the pre-QAT workstation rows and
the QAT device rows can therefore reflect the QAT stage and export precision rather than a
preprocessing change. The higher WorldFloods-v2 mIoU of the QAT float export is consistent with
a regularization effect from QAT and is not interpreted as a general accuracy gain. The
device-float-to-INT8 comparison in Table~III isolates quantization within the
on-device path.}{The FP32 engines reproduce the workstation scores within 0.1 IoU points, and every INT8 or FP16 engine is
compared with the FP32 engine of the same seed on the device.}

\paragraph*{GDACS \chghead{pseudo-label}{teacher-label} pool and WorldFloods-v2}
Both are Level-1C and fall naturally into the S11 space, so the S11
$\mu,\sigma$ are applied directly after the $\times 10^{-4}$ scaling, with no
dataset-specific adjustment. For GDACS this also means the S11-fine-tuned teacher is applied in the same spectral regime it was fine-tuned in, so no
dataset-specific rescaling is required. The pool carries no ground truth, so label accuracy on
it is not directly measurable; Section~\ref{sec:inherit} reports what the paired pre-event
acquisitions establish.

\paragraph*{Training-pool composition}
A side effect of the pre-filter of Section~\ref{sec:curation} is that every training \chg{tile}{image}
contains at least 5\% predicted \chg{flood}{water}, so the student never sees a fully dry scene
during distillation. \chgx{False-positive behavior on entirely water-free landscapes is
therefore not exercised by the training data, and neither OOD benchmark can score it
because both are cut from delineated flood extents. The pre-event baselines of
Section~IV-G are the closest measurement we have: the student raises a false alarm
on $1.9\pm0.8\%$ of the 8{,}883 baselines the teacher left empty, over four seeds.}{False-positive behavior on
water-free landscapes is therefore not covered by the training data. Section~\ref{sec:index} measures it on 200 water-free
WorldFloods windows and 49 dry tiles with independent references.}

\paragraph*{STURM normalization}
STURM needs one extra step, because its chips are not distributed in reflectance
units. The dataset paper documents the forward transform used to produce them: a
min--max scaling with a fixed digital-number range of 400--1600 applied identically to
every Sentinel-2 channel. We invert it exactly. Writing $v$ for the stored pixel value,
$\mathrm{DN} = 400 + 1200\,v$ and reflectance $= \mathrm{DN}\times10^{-4} = 0.04 +
0.12\,v$. We apply this inverse without clipping; stored STURM values can fall below the
nominal $[0,1]$ interval, so recovered reflectance can fall below 0.04. The six S11-matching bands are then selected from STURM's nine (indices
$[0,1,2,6,7,8]$ of B2, B3, B4, B5, B6, B7, B8, B11, B12) and the S11 $\mu,\sigma$ applied
unchanged. Note that STURM's near-infrared band is B8, not the narrower B8A the other
datasets supply: its authors state that they excluded B8a. We feed B8 into the model's B8A
slot, a 32\,nm center-wavelength mismatch that we did not correct. \chgx{A guard removes the $\approx$0.6\% of chips containing NaN pixels.}{About 0.6\% of the chips contain NaN pixels, which are ignored in scoring, so every chip stays in the evaluation.}

The transform is dataset-wide and image-independent, so nothing about the test set enters the
normalization: no test-time adaptation and no per-image statistics. It also follows the same
rule as every other dataset here, namely recover reflectance and then apply the training-split
statistics. We prefer a documented inverse to an empirical rescaling of the evaluation set,
since the latter fits the evaluation data to the model\del{, and model rankings on this benchmark are
sensitive to that choice}.

STURM's label legend has five water classes: non-flood (0), flooded (1), river-related
features (2), open water (3), reservoirs (4) and lakes (5), plus a no-data class. We
score \chgx{$\{1,2\}$ as flood, set $\{3,4,5\}$ to ignore}{all five water classes $\{1,\ldots,5\}$ as water}, and treat cloud and no-data as
ignore (Section~\ref{sec:datasets}).

\paragraph*{Relevance to the OOD claim}
Applying the same fixed normalization removes simple scale differences, but STURM retains a
processing-level shift and a B8/B8A band mismatch, so its domain gap mixes geographic
generalization with product-level shift. The blue band provides a diagnostic for the processing-level difference.
Sen1Floods11, the GDACS pool and WorldFloods-v2 are Level-1C top-of-atmosphere;
the STURM chips appear not to follow the same distribution. At
top-of-atmosphere the air itself scatters light into the sensor, most strongly at short
wavelengths, so the blue band has a floor below which no surface can take it. Pooled over
the valid pixels of 200 Sen1Floods11 tiles the 1st percentile of B2 is $0.081$, and the
percentiles run blue $0.081 >$ green $0.069 >$ red $0.039$, which is the Rayleigh ordering.
WorldFloods gives $0.080$. STURM, pooled over 600 tiles, gives $0.013$, with blue
\emph{below} green.
These statistics are inconsistent with the Level-1C distributions observed in Sen1Floods11 and
WorldFloods-v2 and indicate that an atmospheric or radiometric correction was applied. The
dataset paper does not state a processing level, so this is an inference from the data and it
identifies only that a correction was applied, not which one.

We do not correct for it, since atmospherically correcting the training data would change the
space every model in this paper was fitted in. Two consequences follow: part of the STURM domain
gap is a processing-level gap rather than a purely geographic one, and on STURM the haze axis of
\chg{Section~IV-F}{the earlier diagnostics of Supplementary Section~\ref{S-sec:s_harmonisation}} adds an atmosphere back onto a corrected image rather than deepening one
that is already present.

\subsection*{Hyperparameters}

Table~\ref{tab:hparams} summarizes the training hyperparameters used across all experiments.
All \chg{pseudo-label}{teacher-label} experiments use the FPS-selected pool unless the selector is named.
\chg{The QAT phase uses the ReLU6-from-scratch checkpoint as starting point.}{QAT starts from the trained FP32 student of the same seed.}

\begin{delfloat}
\begin{table}[h]
\revdelfloat{TABLE V}
\centering
{\footnotesize \revst{Submitted Table~V. Default training hyperparameters for the main
comparisons; deviations are stated in the relevant table captions and sections.}\par}
\small
\setlength{\tabcolsep}{5pt}
\resizebox{\columnwidth}{!}{%
\begin{tabular}{lll}
\toprule
Hyperparameter & EfficientViT-B0 (all objectives) & UNet-ResNet50 \\
\midrule
Optimizer            & AdamW & AdamW \\
Learning rate        & 5e-5 & 5e-5 \\
Weight decay         & 0.05 & 0.05 \\
Learning-rate schedule          & Cosine & Cosine \\
Batch size           & 16 & 16 \\
Epochs               & 50 & 50 \\
Training input (resize) & 224$\times$224 & 224$\times$224 \\
Evaluation resolution & 512$\times$512 & 512$\times$512 \\
Input bands          & B2, B3, B4, B8A, B11, B12 & B2, B3, B4, B8A, B11, B12 \\
Data augmentation    & dihedral $+$ jitter & dihedral $+$ jitter \\
Head mixture $\alpha$ (focal/KD) & 0.5 & 0.5 \\
OFA branch weight $\alpha_\text{OFA}$ & 0.3 & N/A \\
OFA modulation $\gamma$ & 1.2 & N/A \\
KD temperature       & 4.0 & 4.0 \\
Pseudo-label pool       & 23,720 $\to$ 5,987 $\to$ 2,500 & idem \\
Seeds (reported)     & 4 & 4 \\
\midrule
\multicolumn{3}{l}{\textit{QAT fine-tuning (ReLU6 model only)}} \\
\midrule
QAT optimizer        & AdamW & N/A \\
QAT learning rate    & 1e-5 & N/A \\
QAT epochs           & 10 & N/A \\
Quantization bits    & W8A8 & N/A \\
BatchNorm $\gamma$/$\beta$  & frozen & N/A \\
MatMul precision     & FP16 (LiteMLA) & N/A \\
\bottomrule
\end{tabular}}
\end{table}
\end{delfloat}

\begin{table}[h]
\revnewfloat
\centering
\caption{\add{\textbf{Training hyperparameters.} Pretrained students: EfficientViT-B0, MobileNetV3-Small, LightFormer-ResNet18
and UNet-ResNet50. Small students: the 418- to 12,230-parameter designs, trained from scratch. LightFormer with its
authors' setup: Section~\ref{sec:baselines}.}}
\label{tab:hparams}
\footnotesize
\setlength{\tabcolsep}{3pt}
\begin{tabular}{@{}p{0.27\columnwidth}p{0.69\columnwidth}@{}}
\toprule
Setting & Value \\
\midrule
Optimizer & AdamW, cosine schedule, weight decay 0.05, gradient clipping 5, batch size 16 \\
Learning rate & $5\times10^{-5}$ pretrained, $5\times10^{-4}$ or $5\times10^{-3}$ small (chosen on S11 validation) \\
Training length & 50 epochs (7,800 updates with 2,500 images), also 7,800 updates in the equal-update 252-image settings \\
Input & $512\!\times\!512$, bands B2, B3, B4, B8A, B11, B12, dihedral flips and multiplicative jitter \\
Loss & 0.5 focal ($\gamma\!=\!2$) $+$ 0.5 logit distillation ($T\!=\!4$), cross-entropy without teacher \\
OFA variant & branch weight 0.3 (mean over branches), modulation 1.2 \\
Checkpoint & best water IoU per event on S11 validation \\
Quantization & main student: W8A8 QAT, 10 epochs, frozen batch normalization, attention in FP16. Small students: post-training (QAT for three engines). Acceptance within 2 IoU points of FP32 on S11 validation \\
\bottomrule
\end{tabular}
\end{table}

\bibliographystyle{IEEEtran}
\bibliography{references}

\makeatletter
\let\BIOdepth\@IEEEBIOphotodepth\let\BIOwidth\@IEEEBIOphotowidth
\def\@IEEEBIOskipN{1.5\baselineskip}
\def\@textbottom{\vskip\z@ plus 1fill}
\makeatother
\begin{IEEEbiography}[{\parbox[t][\BIOdepth][t]{\BIOwidth}{%
  \includegraphics[width=\BIOwidth,clip,keepaspectratio]{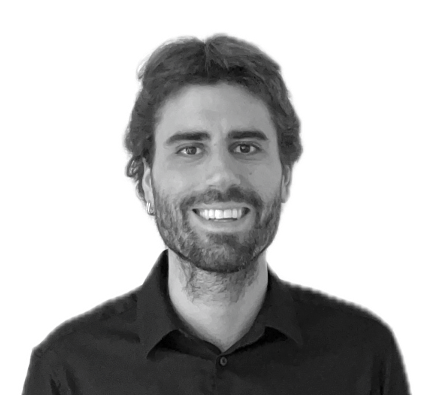}}}]{Fabian Schmalstieg}
received the M.S. degree in computer science from the University of Freiburg, Freiburg, Germany, in 2023. Since 2024, he has been with the Fraunhofer Institute for Telecommunications, Heinrich-Hertz-Institut, Berlin, where he works as a research scientist in the Efficient Deep Learning Group. His research interests include climate-focused AI and making geospatial foundation models deployable on resource-constrained hardware. He has contributed to international standardization activities within the ISO/IEC Moving Picture Experts Group.
\end{IEEEbiography}

\begin{IEEEbiography}[{\includegraphics[width=1in,height=1.25in,clip,keepaspectratio]{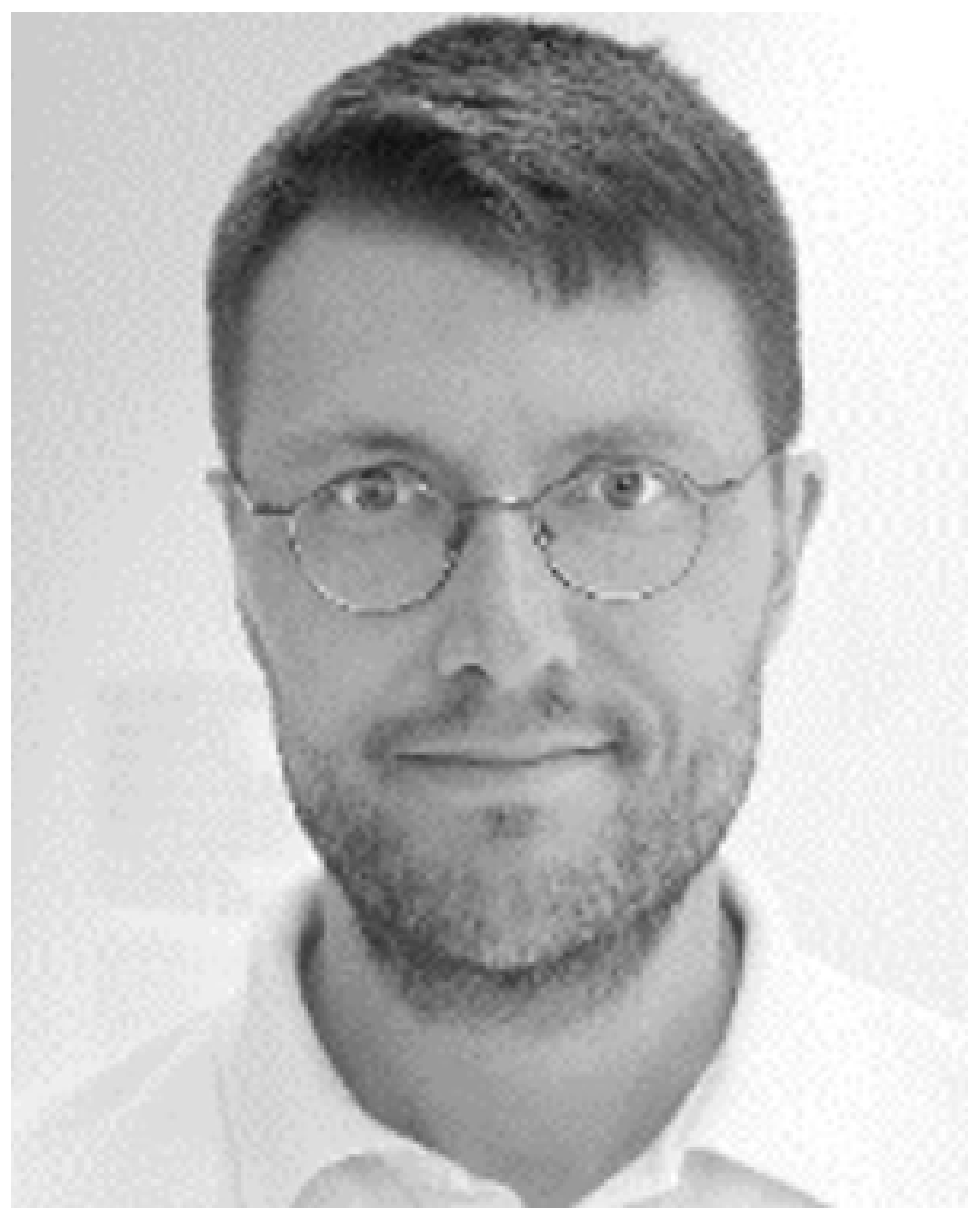}}]{Karsten Müller}
(Senior Member, IEEE) received the Dr.~Ing.\ degree in electrical engineering and the Dipl.~Ing.\ degree from the Technical University of Berlin, Berlin, Germany, in 2006 and 1997, respectively. Since 1997, he has been with the Fraunhofer Institute for Telecommunications, Heinrich-Hertz-Institut, Berlin, where he is currently the Head of the Efficient Deep Learning Group. His research interests include multi-dimensional video coding, efficient federated learning, and compression of neural networks. He has been involved in international standardization activities, successfully contributing to the ISO/IEC Moving Picture Experts Group for work items on visual media content description, multiview, multi-texture, 3D video coding, and neural network representation. He co-chaired an adHoc Group on 3D Video Coding from 2003 to 2012. He was the Chair and Editor of IEEE conferences and the Senior Area Editor of IEEE Transactions on Image Processing.
\end{IEEEbiography}

\begin{IEEEbiography}[{\includegraphics[width=1in,height=1.25in,clip,keepaspectratio]{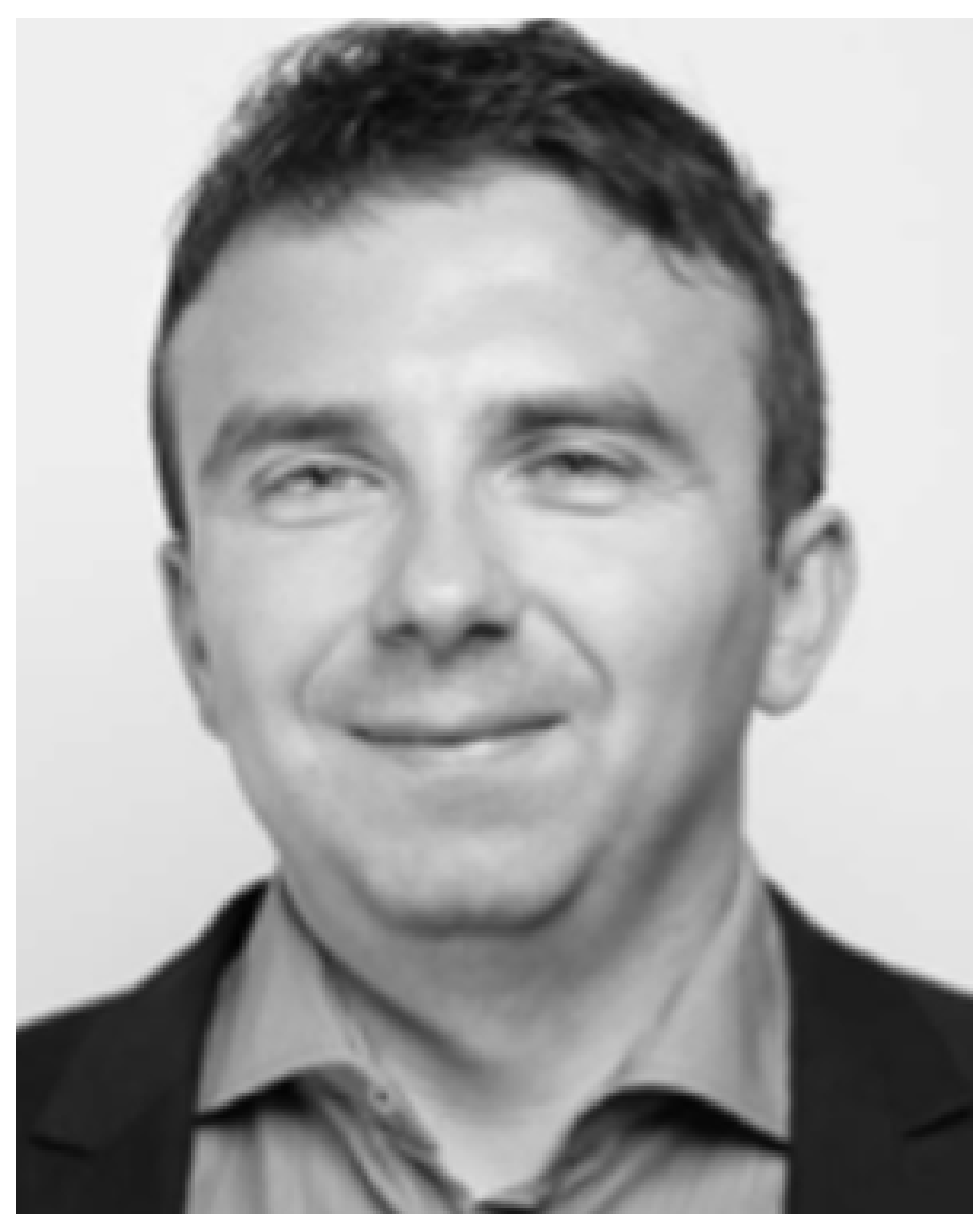}}]{Wojciech Samek}
(Member, IEEE) received the Dr.~rer.~nat.\ degree with distinction (\emph{summa cum laude}) from the Technical University of Berlin, Berlin, Germany, in 2014. He is currently a Professor with the Department of Electrical Engineering and Computer Science, Technical University of Berlin, Berlin, Germany, and is jointly heading the Department of Artificial Intelligence and the Explainable AI Group, Fraunhofer Heinrich Hertz Institute (HHI), Berlin, Germany. He studied computer science with the Humboldt University of Berlin, Berlin, Germany, Heriot-Watt University, Edinburgh, U.K., and The University of Edinburgh, Edinburgh, U.K. He has coauthored more than 150 peer-reviewed journal and conference papers, some of them listed by Thomson Reuters as ``Highly Cited Papers'' (i.e., top 1\%) in the field of Engineering. During his studies, he was awarded scholarships from the German Academic Scholarship Foundation and the DFG Research Training Group GRK 1589/1, and was a visiting Researcher with NASA Ames Research Center, Mountain View, USA. Dr.~Samek is associated faculty, BIFOLD -- Berlin Institute for the Foundation of Learning and Data, ELLIS Unit Berlin and DFG Graduate School BIOQIC, and a Member of the scientific advisory board of IDEAS NCBR. Furthermore, he is the Senior Editor of IEEE Transactions on Neural Networks and Learning Systems, an Editorial Board Member of PLoS ONE and Pattern Recognition, and an elected Member of the IEEE MLSP Technical Committee. He was the recipient of multiple best paper awards, including the 2020 Pattern Recognition Best Paper Award, and part of the expert group developing the ISO/IEC MPEG-17 NNC standard. He is the leading Editor of the Springer book ``Explainable AI: Interpreting, Explaining and Visualizing Deep Learning'' in 2019, the Co-Editor of the open access Springer book ``xxAI -- Beyond Explainable AI'' in 2022, and organizer of various special sessions, workshops and tutorials on topics such as explainable AI, neural network compression, and federated learning.
\end{IEEEbiography}

\ifarxivversion
\ifdefined\ARXIVCOMBINED
\else
\def\revmode{clean}
\documentclass[journal]{IEEEtran}
\usepackage{amsmath,amssymb}
\usepackage{amsthm}
\newtheorem{proposition}{Proposition}
\usepackage{graphicx}
\graphicspath{{./}{figures/}{figures/author_photos/}}
\usepackage{booktabs}
\usepackage{multirow}
\usepackage{subcaption}
\usepackage{stfloats}
\usepackage{xcolor}
\usepackage{colortbl}
\definecolor{useGPU}{HTML}{DCEBF7}\definecolor{useINT}{HTML}{FBE4D5}\definecolor{rowhl}{HTML}{E6F0F8}\definecolor{colhl}{HTML}{EFEFEF}
\usepackage{url}
\usepackage{revmarks}
\usepackage[hidelinks]{hyperref}
\usepackage{placeins}
\extrafloats{40}
\setcounter{topnumber}{3}\setcounter{bottomnumber}{2}\setcounter{totalnumber}{4}\setcounter{dbltopnumber}{4}
\renewcommand{\topfraction}{0.92}\renewcommand{\bottomfraction}{0.75}\renewcommand{\textfraction}{0.06}
\renewcommand{\floatpagefraction}{0.80}\renewcommand{\dbltopfraction}{0.92}\renewcommand{\dblfloatpagefraction}{0.80}
\fi

\renewcommand{\thesection}{S\arabic{section}}
\renewcommand{\thesubsection}{\thesection.\arabic{subsection}}
\renewcommand{\thefigure}{S\arabic{figure}}
\renewcommand{\thetable}{S\arabic{table}}
\renewcommand{\theequation}{S\arabic{equation}}

\ifdefined\ARXIVCOMBINED
  \clearpage
  \makeatletter
  \let\section\@IEEEappendixsavesection
  \gdef\theHsection{Supplement.S\arabic{section}}
  \gdef\thesectiondis{S\arabic{section}}
  \gdef\@IEEEthmcounterinsection#1{S\arabic{#1}}
  \xdef\Hy@chapapp{section}
  \let\SUPP@origlabel\label
  \renewcommand{\label}[1]{\SUPP@origlabel{#1}\SUPP@origlabel{S-#1}}
  \makeatother
  \setcounter{section}{0}
  \setcounter{subsection}{0}
  \setcounter{figure}{0}
  \setcounter{table}{0}
  \setcounter{equation}{0}
  \onecolumn
  \twocolumn[
    \begin{center}
      {\LARGE\bfseries Supplementary Material for\\[0.35em]
      ``Distilling a Geospatial Foundation Model into Compact Surface-Water Segmenters for Edge Flood Mapping''\par}
      \vspace{1.0em}
      {\large Fabian Schmalstieg, Karsten Müller, and Wojciech Samek\par}
      \vspace{1.5em}
    \end{center}
  ]
\else
\begin{document}

\title{Supplementary Material for ``\chg{Cross-Architecture Foundation-Model Distillation for Edge Flood Segmentation}{Distilling a Geospatial Foundation Model into Compact Surface-Water Segmenters for Edge Flood Mapping}''}
\author{Fabian Schmalstieg, Karsten Müller, and Wojciech Samek}
\maketitle
\revlegend
\fi

\begin{oldtext}
\revst{The main article contains the minimum evidence needed for the central argument: the matched-budget comparison, scaling beyond the manual annotation set, external clean evaluation, and measured edge deployment. This supplement retains the longer audit trails, curation ablations, acquisition-shift diagnostics, teacher-semantics checks, optimization analyses and deployment implementation details needed for reproducibility. Robustness experiments are presented here as diagnostics rather than as a headline claim. Table, figure, section and equation numbers written without an S prefix refer to the main article; numbers carrying an S prefix are in this supplement. The abbreviations cross-entropy (CE), knowledge distillation (KD), and One-for-All (OFA) are used as in the main article.}
\end{oldtext}
\begin{newtext}
The main article presents the evidence for the central argument through controlled comparisons, four test sets,
tests on water-free ground and device measurements. This supplement describes the data audit, the held-back and
water-free test sets, the water-index variants and the deployment details. It also reports results for larger students,
LightFormer trained with its authors' setup, learning curves, evaluation measures and additional diagnostics.
Some diagnostics used the original experimental setup, which trained at $224\!\times\!224$ pixels on 2.6\,km GDACS
tiles and scored mIoU with permanent water removed. These diagnostics are labeled ``original experimental setup''.
Table, figure, section and equation numbers without an S prefix refer to the main article. Numbers with an S prefix
refer to this supplement. The abbreviations cross-entropy (CE), knowledge distillation (KD), and One-for-All (OFA)
follow the main article. The primary measure is water IoU per event, expressed in points from 0 to 100.
\end{newtext}

\section{\chghead{Data-Overlap and Contamination Audit}{Data Audit, Held-Back Events and Water-Free Ground}}
\label{sec:s_audit}

\begin{newtext}
\paragraph*{Footprint audit}
Every GDACS image was downloaded again at $512\!\times\!512$ pixels using the same product and center as the earlier
2.6\,km tile. Its central $264\!\times\!264$ pixels are identical to that tile. We compared the projected native
footprints of all training images with those of every test image. No training image overlaps a test image on the ground.
STURM-Flood and WorldFloods share three EMS activations (EMSR419, EMSR422 and EMSR438), so we treat them as complementary
rather than independent.

\paragraph*{Event-identity screen}
A stricter screen based on event identity and location flags 31 of the 2,500 training images. These images belong to,
or lie within 25\,km and 30 days of, an event in the held-back set or Sen1Floods11. None overlaps a held-back or
Sen1Floods11 image on the ground.

\paragraph*{Held-back WorldFloods events}
The training and validation splits of WorldFloods-v2 contain 116 events with 466 flood maps. Before inspecting any labels,
we removed every map whose outline and date overlap those of a GDACS image, a Sen1Floods11 image or a test image from
another set. This leaves 97 events with 349 maps. We use the WorldFloods test-split protocol, with native
$512\!\times\!512$ windows, all reference water scored and clouds ignored. Each finished model was scored once.
No score on this set was used to select a model, threshold or comparison.

\paragraph*{Water-free ground}
The 200 water-free windows are the valid $512\!\times\!512$ windows in the WorldFloods test split that contain no
reference water. They come from 9 events, and three events account for 159 windows. We therefore average the alarm rate
over events. The dry tiles are random land locations of nine land-cover types from ESA WorldCover~\cite{zanaga2022worldcover}:
cropland, grass and shrubs, forest, bare ground, urban, dry floodplain, and terrain with mountain shadow, cloud shadow or
snow. The JRC Global Surface Water layer~\cite{pekel2016high} has never recorded water in the tile or within 100\,m of
it, and no training image lies within 25\,km. These Sentinel-2 Level-1C images were downloaded with the same code as the GDACS images.
Of 159 tiles, one author retained 49 after inspecting each full tile in true-color and short-wave-infrared images. Water appears dark in the latter.
Every tile with the slightest sign of water, or any doubt, was removed. The retained set contains 28 ordinary tiles and 21
with shadows or snow. Alarms are counted in the central $264\!\times\!264$ pixels, where the maps also show no water.
We report the ordinary and shadow-or-snow groups separately.
\end{newtext}

\begin{oldtext}
\revst{\textbf{Submitted audit of the original pool.} The main article reports overlap thresholds and the sensitivity of the label-source result to removing nearby inputs. The full audit is retained here.}

\revst{The pseudo-pool is assembled from a global event catalog, while both out-of-distribution (OOD) datasets
draw on Copernicus Emergency Management Service (EMS) flood events. The question is whether any OOD benchmark scene is effectively also a training scene. Every pool tile has a geolocation and an acquisition date, while every
STURM chip and WorldFloods area of interest (AOI) has corresponding location metadata, so we answer it by comparing coordinates and
dates rather than by
asserting that two catalogs must be disjoint. At the matched budget no STURM chip and no WorldFloods test AOI lies within
25\,km of any of the 252 training tiles the audit was run against, at any date. At the
deployment budget of 2{,}500
tiles, 54 of the 2{,}675 STURM chips (2.0\%, three of the 29 activations) lie within
25\,km \emph{and} 30 days of a training tile; no WorldFloods AOI does, and tightening the
test to 5\,km and two days, the scale at which two acquisitions could share ground,
leaves nothing on either benchmark. Dropping those three activations \emph{raises} STURM
mean intersection over union (mIoU) by 0.003--0.007 for every model in the study, including the teacher, which never saw
the pool, and the Sen1Floods11 configurations, which never saw GDACS: it is a difficulty property
of those maps rather than an advantage that training created. Those three activations
account for 4 of the 46 flood maps, and the quantity the main article reports is unmoved by
dropping them: at the matched budget the pseudo-minus-Sen1Floods11 gap for our student is
$+0.0193$~mIoU per flood map over all 46 and $+0.0189$ over the remaining 42. Distances are
computed between chip and AOI centroids rather than footprint geometry; for WorldFloods, whose
AOIs are full scenes, this is the looser of the two tests.
Separately, STURM and WorldFloods themselves share three EMS activations, amounting to
coincident ground at two of the eighteen WorldFloods AOIs, so we call the two benchmarks
complementary rather than independent throughout; excluding those AOIs moves every
WorldFloods number by less than $0.001$~mIoU. The audit is one script over the pool manifest and the two benchmark metadata tables, released
with the code.}
\end{oldtext}

\begin{deltext}
\paragraph*{\revst{Secondary statistics for the matched-budget block}}
\revst{The main paper reports the paired Wilcoxon test, with Holm correction across the five model comparisons. As a more conservative check, we also apply the exact sign test. After the same correction, two of the five comparisons remain significant: EfficientViT-CE ($p=4.5\times10^{-2}$) and UNet-Logit-KD ($p=7.8\times10^{-4}$).}

\revst{Across the 46 flood maps, the pseudo-label model performs better on 32, 30, 30, 27, and 36 maps for EfficientViT-CE, EfficientViT-Logit-KD, EfficientViT-OFA, UNet-CE, and UNet-Logit-KD, respectively. When the paired Wilcoxon analysis is repeated at the event level using 29 flood events, three of the five comparisons remain significant after Holm correction.}
\end{deltext}

\section{Subset Selection\add{ (original experimental setup)}}
\label{sec:s_curation}
The main article deploys farthest-point sampling (FPS) over the 15-dimensional descriptor of
Eq.~(2) and states that the pre-filter, not the ranking rule, is the component whose effect is
measurable at the deployment budget. This section records the comparison behind that statement\add{. It used the
original experimental setup and was not repeated at $512\!\times\!512$ pixels}.

\paragraph*{Selectors compared}
All selectors draw $N$ tiles from the same 5{,}987-tile filtered candidate pool and are
followed by the identical training \chg{recipe}{setup}. \emph{FPS} is the greedy $k$-center rule of Eq.~(3).
\emph{Confidence ranking} takes the $N$ tiles with the highest mean teacher confidence
$\mu_T$. \emph{GLISTER}~\cite{killamsetty2021glister} selects a subset by greedy maximization of
a validation-set log-likelihood surrogate, evaluated here on the Sen1Floods11 validation split.
\emph{R2-Core} (Reliability $\times$ Robustness Core-set) uses a lazily greedy weighted
facility-location objective over the pool embedding, with tile weights combining reliability,
informativeness, and measured cloud probability. \emph{Uniform filtered} draws $N$ tiles at random from the filtered pool, and \emph{uniform
unfiltered} draws them at random from the full 23{,}720-tile pool, which isolates the
contribution of the pre-filter itself.

\paragraph*{Result}
At $N\!=\!2{,}500$ the clean STURM-Flood spread across all selectors is $0.012$~mIoU, of the
order of the seed spread, so the ranking rule is not resolved at the deployment budget. \chg{The pre-filter is:}{The effect of the pre-filter depends on the selector:} \chg{removing it costs up to $0.048$~mIoU clean}{removing it costs 0.1 mIoU points for FPS, 0.3 for random selection, 4.0 for GLISTER and about 31 for confidence ranking}, and confidence ranking without the
pre-filter degrades in the manner the active-learning literature predicts, because it
concentrates the selection in the region of high $\mu_T$ and low $\sigma_T$. FPS is retained as
the deployed selector on that basis rather than on a measured advantage over the alternatives.

\paragraph*{Perturbation behavior}
On the scalar-gain axis the curated configurations are uniformly flat, losing $0.003$--$0.010$
worst-case mIoU at $N\!=\!2{,}500$ and $0.004$--$0.011$ in a \chg{recipe-matched}{setup-matched} replication at
$N\!=\!252$, against $0.037$ for the Sen1Floods11 reference at the same budget. Because the
$N\!=\!2{,}500$ curation runs use no augmentation while the Sen1Floods11 reference trains with
exactly the per-channel jitter that the gain axis tests, we repeated the comparison with the
\chg{recipe}{setup} matched; the difference is not an artifact of the augmentation. On the haze axis the
selectors do spread, but the ordering does not survive a change of pool size, so we report no
selector-level haze result. Worst-case drop is also not scale-free, and a configuration with
lower clean accuracy has less to lose. These controls associate the scalar-gain difference with
the training source under the tested \chg{recipes}{setups}; they do not isolate scene diversity as a causal
variable, and we make no diversity-causality claim.

\begin{figure*}[t]
\centering
\includegraphics[width=0.70\linewidth]{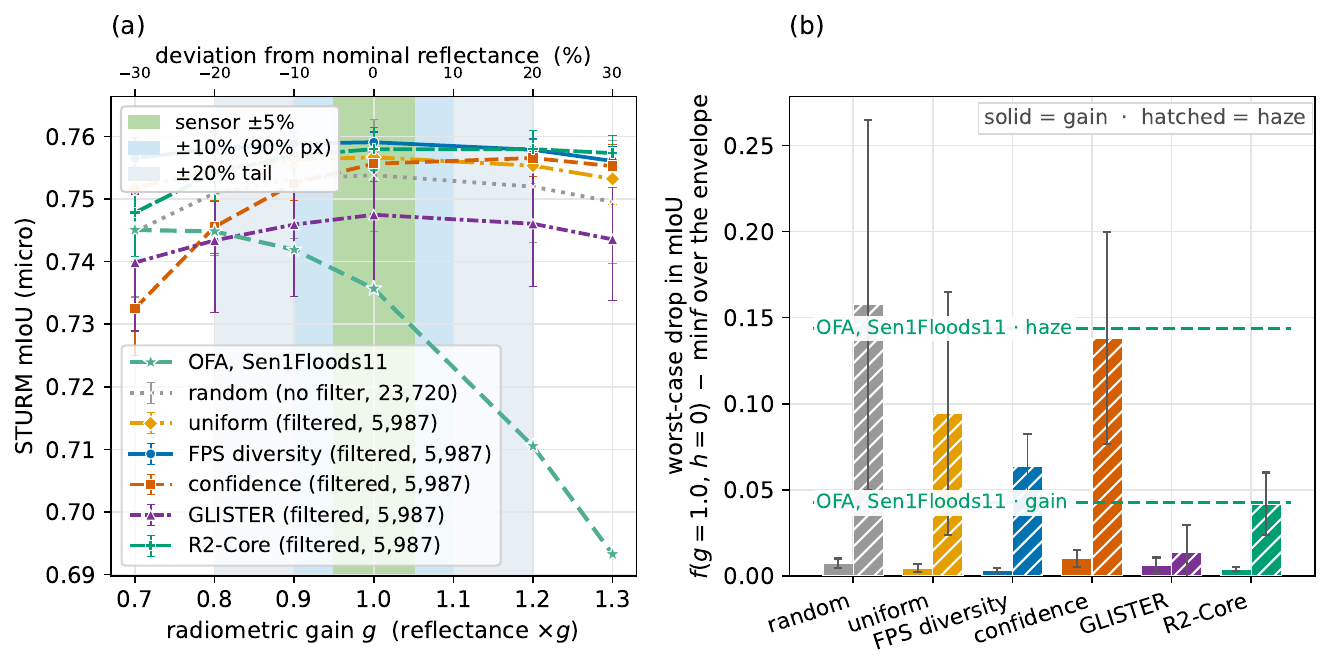}
\caption{%
  \textbf{Subset-selection robustness at $N\!=\!2{,}500$} (\add{original experimental setup, }four seeds, micro mIoU).
  (a) Clean accuracy across the scene-wide gain envelope, with the documented
  Sentinel-2 uncertainty bands shaded. (b) Worst-case mIoU drop under gain and haze,
  with the Sen1Floods11 reference at its $N\!=\!252$ cap shown as dashed lines. R2-Core is included as an
  additional selector for context; no conclusion depends on this individual variant. All selector runs use the unaugmented N=2,500 curation \chg{recipe}{setup} and are not comparable with \chg{Table II or \S IV-F}{the main article}; \chg{While}{the} Sen1Floods11 OFA model at N=252 matches the augmented \chg{recipe}{setup}.
}
\label{fig:s_curation_robustness}
\end{figure*}

Figure~\ref{fig:s_curation_robustness} shows the same selector family on the two perturbation
axes. The filtered selectors remain closely grouped under global gain, whereas their haze
sensitivity is more dispersed. Because the haze ordering changes in the \chg{recipe-matched}{setup-matched}
smaller-budget replication, we treat that spread descriptively rather than as a stable selector
ranking.

\section{Spatial Harmonization, Evaluation Scale, and Radiometric Diagnostics\add{ (original experimental setup)}}
\label{sec:s_harmonisation}

\add{This section reports the original experimental setup, with 224-pixel training on 2.6\,km tiles and mIoU scoring with permanent water removed.}
The two training sources differ in physical geometry before the common $224\!\times\!224$ student-training resize. Sen1Floods11 chips are approximately $4.63\times5.12$\,km, whereas the downloaded GDACS tiles are approximately $2.60\times2.62$\,km. A deterministic center crop of Sen1Floods11 to 264 native pixels was therefore used as a geometry-matched control. The crop is not a perfect isolation of geometry because it also removes flood content: more than half of the positive area is lost in many chips and some become fully dry. Nevertheless, it was done for asking whether the original training-field-of-view difference can explain the matched-budget result. \add{The main article removes this difference by downloading every GDACS image at the Sen1Floods11 extent. Its Table~III directly reports the effect of tile extent.}

For the primary EfficientViT One-for-All (OFA) comparison on STURM-Flood, the clean pseudo-minus-cropped-Sen1Floods11 difference is about $+0.006$ mIoU per flood map and is not resolved at $p<0.05$. Cross-entropy similarly loses a clear separation, while Logit-KD retains a stronger positive margin. This does not mean that teacher supervision has no effect, but that the original matched-budget results (pseudo labels ahead) cannot be accounted as exclusive label-source effect.

\paragraph*{Evaluation-scale control}
STURM-Flood chips are distributed at $128\!\times\!128$ over a fixed 1.28\,km extent, so the
$512$ evaluation tensor used throughout is a $4\times$ upsample to 2.50\,m effective sampling,
against WorldFloods-v2's native 10.0\,m. We re-scored the same weights at 128, 224 and 512
while holding the ground extent fixed. Because the extent is fixed, this changes effective
sampling and the network's deepest feature-map size together and the two cannot be separated
on this benchmark; 128 and 512 are exact integer resamplings of the native grid, whereas 224 is
not, so the 128-versus-512 comparison is the clean one.

The modified normalized difference water index (MNDWI) reference uses the same evaluation
resampling as the learned models.

\begin{table}[t]
\centering
\caption{STURM-Flood clean micro mIoU for identical weights at three evaluation tensors
  (fixed 1.28\,km ground extent\add{, original experimental setup}). Four training seeds where seeds exist; the teacher and the
  index are single deterministic models.}
\label{tab:supp_scale}
\small
\setlength{\tabcolsep}{3pt}
\begin{tabular}{lccc}
\toprule
Model & 128\,px & 224\,px & 512\,px \\
      & (10.0\,m) & (5.71\,m) & (2.50\,m) \\
\midrule
MNDWI $>0$                       & 0.7443 & 0.7449 & 0.7467 \\
Teacher, Prithvi-EO-2.0          & 0.7619 & 0.7569 & 0.7550 \\
EfficientViT-B0 OFA, $N\!=\!2{,}500$ & 0.7464 & 0.7590 & 0.7561 \\
EfficientViT-B0 Logit-KD, $N\!=\!2{,}500$ & 0.7436 & 0.7596 & 0.7578 \\
EfficientViT-B0 CE, $N\!=\!2{,}500$ & 0.7422 & 0.7558 & 0.7537 \\
\bottomrule
\end{tabular}
\end{table}

\begin{figure*}[t]
\centering
\includegraphics[width=0.9\linewidth]{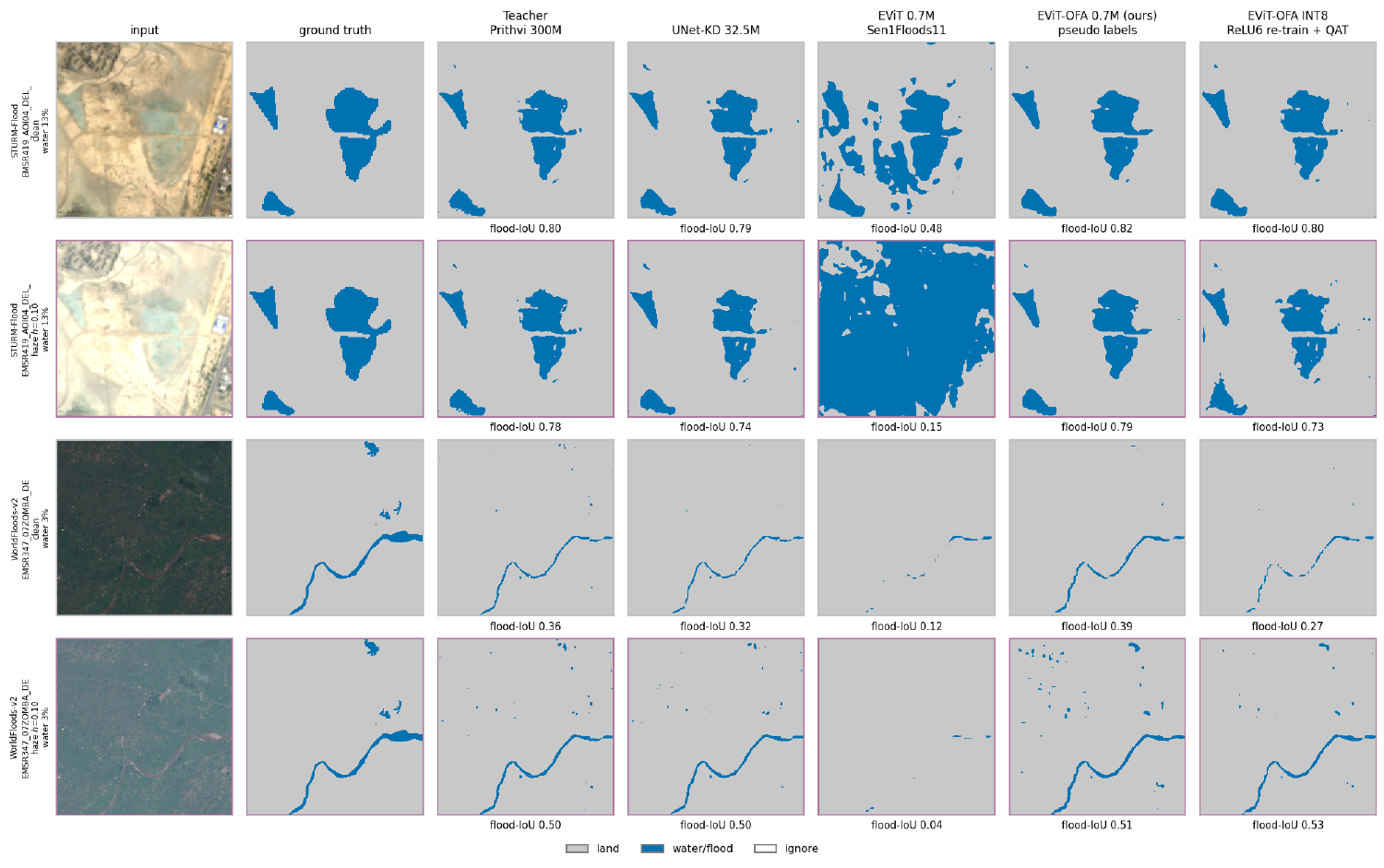}
\caption{Qualitative illustration of the haze perturbation for input, GT, teacher, UNet-Logit KD, EViT-S11, EViT-OFA pseudo, EViT-OFA INT8 with corresponding flood/water IoU\add{ (original experimental setup)}. One STURM-Flood scene (rows 1--2) and one WorldFloods-v2 scene (rows 3--4) are shown clean and with additive haze at $h=0.10$. The true-color image uses a fixed zero-referenced stretch so that the applied perturbation remains visually comparable between rows. The panel is included to illustrate the appearance of the synthetic perturbation rather than to support the quantitative robustness results. }
\label{fig:s_qual_haze}
\end{figure*}

The MNDWI index is nearly scale-invariant, moving by $\le0.003$ across $128 - 512$. The data driven models move by $0.010$--$0.014$ and
the teacher by $+0.007$ in the opposite direction. And the paired student--teacher comparison
of the main article is not scale-invariant: at native 128\,px the deployed model's paired
difference is $-0.028$ water IoU per flood map; the teacher is ahead, but the paired test narrowly
misses the $0.05$ threshold (student ahead on 16 of 46, $p\!=\!0.054$). The main article reports the 512 protocol
throughout and discloses this sensitivity rather than claiming sampling invariance.

\paragraph*{Perturbation models}
All three axes are applied in reflectance space, before the Sen1Floods11 normalization statistics are applied. Writing $r_b$ for the reflectance of band
$b$: the \emph{scalar gain} axis maps $r_b \mapsto g\,r_b$ for a single factor
$g\in\{0.80,0.90,1.00,1.10,1.20,1.30\}$ applied to all six bands; The \emph{haze} perturbation adds a wavelength-dependent offset to each band,

$$
r_b' = r_b + h\,w_b,
$$

where $w_b$ decreases from blue to short-wave infrared~\cite{makarau2014haze} and
$h\in\{0,0.05,0.10,0.20\}$ controls the perturbation strength.

For the \emph{per-band gain} perturbation, we use two variants. In the first, only the visible bands (B2, B3, B4) are increased or decreased by a factor of $1+d$ or $1-d$, while the near- and short-wave-infrared bands (B8A, B11, B12) remain unchanged. In the second, the two spectral groups are changed in opposite directions: when the visible bands are multiplied by $1+d$, the near- and short-wave-infrared bands are multiplied by $1-d$, and vice versa. We use
$d\in\{0.05,0.10,0.20,0.30\}$. The first three severities span the order of band-dependent uncertainties reported for Sentinel-2 Level-2A products, while $d=0.30$ is included as an explicit stress level~\cite{gorrono2024framework}.
Sensitivity is summarized as
the \emph{worst-case drop}: clean mIoU minus the lowest mIoU over the axis, computed per seed
and then averaged over seeds.

Figure~\ref{fig:s_qual_haze} provides a qualitative illustration of the additive haze perturbation used in this diagnostic. The scalar-gain perturbation is visually simpler: it multiplies all six reflectances by the same factor and therefore changes the overall scene brightness without changing band ratios. We include the example only to make the synthetic perturbations concrete; the qualitative panel is not used as evidence for the robustness comparisons.
Two properties of these axes matter for interpretation. A scalar gain changes neither
within-band contrast nor band ratios, so a normalized-difference index (MNDWI) is exactly invariant to
it, so its worst-case drop on that axis is exactly zero with no parameters; the range
$g\in[0.8,1.2]$ also lies inside the per-channel jitter both training \chg{arms}{setups} use for augmentation, so only
$g=1.3$ is an extrapolation. Per-band gain preserves within-band contrast and moves band
ratios, and haze does both. Under the per-band axis the teacher-supervised advantage does not
persist and reverses on WorldFloods-v2 across objectives and severity tiers.
We therefore interpret the scalar-gain result as \emph{brightness invariance}, not as evidence of general radiometric robustness.

The haze axis is additive and wavelength-dependent. Its matched-budget advantage is removed by
the geometry-matched control, so it cannot be attributed to the label source. Within the teacher-supervised \chg{arm}{runs}, the worst-case STURM-Flood haze drop decreases from $0.081$
at $N\!=\!252$ to $0.028$ at $N\!=\!2{,}500$; because both budgets use the same training
source, this comparison is not exposed to the geometry confound above.

\paragraph*{Paired group-level results}
Under the uncropped matched-budget comparison, all six tests (three configurations $\times$ two
axes) remain significant after Holm correction; the weakest is EfficientViT-CE under haze at
$p=1.2\times10^{-2}$. Under the geometry-matched control, only the scalar-gain result survives:
all three objective pairs separate per flood map at $p\le2.5\times10^{-8}$ under both the
Wilcoxon signed-rank and the exact sign test.

Under the haze axis the matched-budget difference does \emph{not} survive the geometry-matched
control: against the cropped Sen1Floods11 experiment none of the three objective pairs reaches
$p<0.05$ (cross-entropy $-0.0112$, 17 of 46; Logit-KD $+0.0025$, 24 of 46; OFA $+0.0170$, 27 of
46). Part of the haze contrast is therefore attributable to acquisition geometry rather than to
the label source. We consequently make no general robustness claim to the pseudo-labelled pool being more robust due to sample diversity.

\section{Spectral-Index Reference Baseline}
\label{sec:s_mndwi}

A fixed MNDWI threshold is evaluated through the same dataloaders, resizing, class mapping and ignore masks as the learned models. The rule is
\[
\mathrm{MNDWI}=\frac{B3-B11}{B3+B11},\qquad \hat y=\mathbf{1}[\mathrm{MNDWI}>0].
\]
It has no learned parameters and loads no checkpoint. \chgx{Under the main clean protocol it reaches
0.747~mIoU on STURM-Flood and 0.763 on WorldFloods-v2, and paired group-level tests separate
neither the 300\,M teacher nor the deployed student from it on either benchmark.}{Table~\ref{tab:s_index} compares it with
nine further variants. These include MNDWI~\cite{xu2006modification} with a tuned threshold, the normalized difference
water index (NDWI)~\cite{mcfeeters1996use}, the automated water extraction indices AWEI$_\text{nsh}$ and
AWEI$_\text{sh}$~\cite{feyisa2014automated}, and the water index 2015 (WI2015)~\cite{fisher2016comparing}.
The latter four indices are each evaluated at threshold zero and at a threshold tuned on the Sen1Floods11 training
and validation labels. Suitable thresholds can vary between scenes, regions and datasets~\cite{ji2009analysis,mateogarcia2021towards}.}

\begin{table*}[t]
\revnewfloat
\centering
\caption{\add{\textbf{Water-index variants.} Ten water-index variants compared with the learned models. Scores are water IoU per
flood event (points) and false alarms on 200 water-free WorldFloods windows (\%, averaged over 9 events).
Tuned thresholds are selected on the Sen1Floods11 training and validation labels. Student scores are means over four
seeds. Bold marks the best index variant in each column, with lower values preferred for false alarms.}}
\label{tab:s_index}
\footnotesize
\begin{tabular}{lccccc}
\toprule
Rule or model & S11 test & WorldFloods test & STURM-Flood & Held back & False alarms \\
\midrule
MNDWI, threshold 0 & 68.3 & 71.2 & 50.4 & 49.3 & 57.0 \\
MNDWI, tuned on S11 & 76.4 & 73.9 & 42.8 & 53.3 & 38.3 \\
NDWI, threshold 0 & 67.4 & 66.0 & 45.8 & 53.7 & \textbf{14.8} \\
NDWI, tuned on S11 & 76.3 & 70.1 & 49.8 & 54.3 & 31.9 \\
AWEI$_\text{nsh}$, threshold 0 & 76.1 & 73.5 & 30.9 & 52.4 & 21.6 \\
AWEI$_\text{nsh}$, tuned on S11 & 74.2 & 73.8 & 29.6 & 52.5 & 20.4 \\
AWEI$_\text{sh}$, threshold 0 & 76.8 & 73.9 & 51.8 & 53.6 & 44.2 \\
AWEI$_\text{sh}$, tuned on S11 & 79.0 & 76.6 & 45.5 & 55.8 & 28.8 \\
WI2015, threshold 0 & 75.1 & 73.2 & \textbf{52.0} & 52.4 & 55.5 \\
WI2015, tuned on S11 & \textbf{79.5} & \textbf{77.4} & 45.3 & \textbf{56.6} & 28.0 \\
\midrule
Teacher & 79.1 & 79.4 & 49.9 & 60.7 & 1.5 \\
Main student & 76.1 & 79.7 & 50.1 & 60.8 & 5.4 \\
Tiny student & 78.1 & 79.2 & 48.1 & 60.0 & 7.1 \\
\bottomrule
\end{tabular}
\end{table*}

\begin{figure*}[t]
\revnewfloat
\centering
\includegraphics[width=\linewidth]{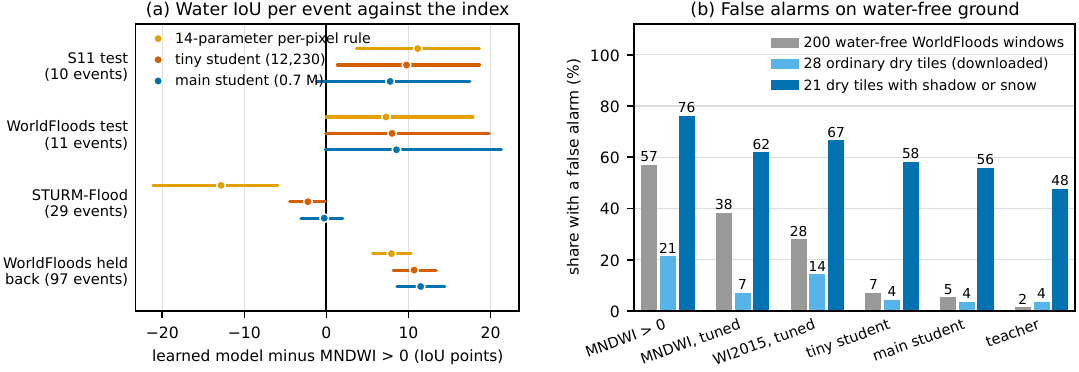}
\caption{\add{\textbf{Learned models compared with water indices.} (a) Difference in water IoU per flood event between three
learned models of increasing size and MNDWI at threshold zero (points, 95\% intervals from resampling whole events).
The 14-parameter rule learns from the 252 hand labels, and the two students learn from the 2,500 teacher-labeled images.
(b) False alarms for the fixed index, two tuned indices and learned models. An alarm means that more than 1\% of a
region is predicted as water. Results cover the 200 water-free WorldFloods windows, averaged over 9 events, and the
28 ordinary and 21 shadow-or-snow dry tiles. Student results are means over four seeds.}}
\label{fig:s_index_ladder}
\end{figure*}

\begin{newtext}
The fixed MNDWI rule is not the strongest index. AWEI$_\text{sh}$ and WI2015 with threshold zero perform better on all
four test sets and raise fewer false alarms. On the Sen1Floods11 test split and STURM-Flood, the best index variants
also score above the main student. Threshold tuning does not transfer consistently across datasets. A threshold tuned
on Sen1Floods11 usually improves performance on its test split but often reduces performance on STURM-Flood.
For example, WI2015 rises from 75.1 to 79.5 on the Sen1Floods11 test split and falls from 52.0 to 45.3 on STURM-Flood.
The main article assesses transfer on the diverse held-back events. There, the best variant, WI2015 with a tuned
threshold, remains 4.2 IoU points below the main student. Every variant raises far more false alarms on water-free ground
than the learned models, except the fixed NDWI rule, which instead misses much of the water.
\end{newtext}

\begin{oldtext}
\paragraph*{\revst{Comparison Across Datasets and Aggregations}}
\revst{The result depends on how performance is averaged. On WorldFloods-v2, the learned models are clearly better when each scene is weighted equally: the MNDWI rule reaches only $0.418$ macro water IoU, compared with $0.504$ for the teacher. This shows that MNDWI performs poorly on several individual scenes, even though its pooled score remains competitive.}

\revst{On STURM-Flood, the opposite is true. When performance is averaged over individual scenes, MNDWI slightly outperforms both learned models. It reaches $0.5945$ mIoU and $0.3891$ water IoU, compared with $0.5921/0.3723$ for the teacher and $0.5909/0.3664$ for the N=2,500 student. We therefore report these per-scene results rather than only the pooled scores.}

\revst{On the in-distribution Sen1Floods11 test set, however, the learned models are clearly better. MNDWI reaches $0.789$ mIoU and $0.650$ water IoU, compared with $0.877/0.787$ for the N=2,500 student and $0.897/0.822$ for the teacher. Here, the teacher performs best, followed by the N=2,500 student, with MNDWI clearly behind both.}
\end{oldtext}

\paragraph*{Further properties\add{ (original experimental setup)}}
Three further properties separate the rule from the learned models. It is exactly invariant to
a scene-wide multiplicative gain, so its worst-case drop on that diagnostic axis is exactly zero
with no parameters. It is far more sensitive to additive haze, losing 44\% of its STURM mIoU and
66\% of its WorldFloods mIoU at the strongest tested level against 4--5\% for the teacher, and
the teacher separates from it per flood map at every tested severity. And neither the choice of
index nor the threshold transfers: NDWI is the stronger rule in distribution (0.721 against
0.650 water IoU) and collapses to 0.249 on WorldFloods-v2, while the optimal threshold sits at
$+0.05$ on STURM-Flood and $+0.15$ on the other two sets, so tuning it on the in-distribution
split yields 0.7379 on STURM, below the untuned zero rule. Setting a threshold per domain
requires labels from that domain, which is what the pipeline exists to avoid.

\begin{oldtext}
\paragraph*{\revst{Hard-labeler control}}
\revst{Replacing the teacher's stored hard masks by $\mathrm{MNDWI}>0$ masks on the same 2{,}500
tiles, with everything else held fixed, separates two roles. As a \emph{labeler} the
foundation model is not replaceable: under plain cross-entropy, which involves no teacher
forward pass, the index-labeled student loses 0.038~mIoU in distribution and on STURM-Flood
(7 of 46 flood maps, $p\!=\!1.8\times10^{-6}$) and is markedly more seed-variable. As a source
of \emph{hard targets} it is: under OFA, where the live frozen teacher still supplies the soft
targets, the two label sources are indistinguishable in distribution and on STURM-Flood, and
the index-labeled arm is ahead on WorldFloods-v2 (16 of 18 scenes,
$p\!=\!1.3\times10^{-3}$). The component the pipeline depends on is therefore teacher
supervision as a whole rather than the stored hard mask specifically.}
\end{oldtext}
\begin{newtext}
\paragraph*{Hard-labeler control}
Replacing the teacher's hard labels with $\mathrm{MNDWI}>0$ labels on the same 2,500 images, with all other settings
fixed, reduces performance by 2.1 to 4.9 IoU points when the student learns from hard labels alone. The difference on
STURM-Flood is $-4.4$ $[-7.9, -1.4]$ IoU points. With the teacher's probability maps added, the two label sources yield
nearly the same accuracy (Table~III of the main article). The pipeline therefore depends on teacher supervision as a
whole rather than on the stored hard label.

\paragraph*{A learned per-pixel rule (descriptive)}
After the size study, we added a learned per-pixel linear rule with 14 parameters, comprising six band weights and one
bias per class. This is the per-pixel linear baseline from an onboard flood-mapping study~\cite{mateogarcia2021towards}.
On WorldFloods, that study found that its linear model matched the best NDWI threshold and scored 6 to 8 IoU points
below small convolutional networks. Our comparison of learned models shows the same ordering under our protocol.
A later in-orbit demonstration ran such flood models on board a satellite~\cite{mateogarcia2023inorbit}.
Table~\ref{tab:s_r1} reports our linear rule's scores. When trained on the hand labels, it outperforms the fixed index
on the Sen1Floods11 test split and held-back events but scores substantially lower on STURM-Flood. It raises false
alarms on 19.1\% of the water-free windows. When trained on the 2,500 teacher-labeled images, it identifies only the
clearest water, a pattern already observed on the validation images. On the Sen1Floods11 test split, 96.8\% of the
pixels it classifies as water are water, but it detects only 58.9\% of all water pixels. The 418-parameter per-pixel
network, trained on the same images with the same loss, reaches 78.6 IoU points on that test split. This suggests that a
single linear boundary in band space cannot fit the teacher's labels across 1,370 flood events, whereas a small nonlinear
model can. Access to each pixel's neighborhood then allows the small students to maintain their accuracy on STURM-Flood
(Section~V-D of the main article).
\end{newtext}
\begin{table}[t]
\revnewfloat
\centering
\caption{\add{\textbf{A learned per-pixel rule.} Water IoU per event (points) for a learned 14-parameter per-pixel rule, added
after the size study as a descriptive comparison. The last row gives the difference from MNDWI with 95\% intervals.
Bold marks intervals that exclude zero.}}
\label{tab:s_r1}
\scriptsize
\setlength{\tabcolsep}{2.5pt}
\resizebox{\columnwidth}{!}{%
\begin{tabular}{lcccc}
\toprule
 & S11 test & WorldFloods test & STURM-Flood & Held back \\
\midrule
Per-pixel rule, 252 hand labels & 79.5 & 78.5 & 37.5 & 57.2 \\
Per-pixel rule, 2,500 teacher labels & 58.1 & 51.1 & 30.6 & 47.5 \\
MNDWI $>0$ & 68.3 & 71.2 & 50.4 & 49.3 \\
Rule (hand labels) $-$ MNDWI & \textbf{+11.1 [+3.6, +18.7]} & +7.3 [0.0, +17.8] & \textbf{$-$12.8 [$-$21.1, $-$5.9]} & \textbf{+7.9 [+5.6, +10.3]} \\
\bottomrule
\end{tabular}}
\end{table}

\begin{newtext}
Fig.~\ref{fig:s_index_ladder} compares learned models of increasing size with the index in terms of water IoU and false alarms.
\end{newtext}
\section{Teacher Semantics and Follow-Up Tests\add{ (original experimental setup)}}
\label{sec:s_semantics}
The main article states that the student reproduces the teacher's
surface-water convention, so target-event flood products still require a reference-water
step. This section gives the proxy construction, tile-level measurements, and follow-up
self-training and teacher-assistant tests\add{. These tests used the students and pool from the original experimental setup}.

Training on a model's output raises a question that accuracy tables cannot answer: when the
student and the teacher agree, is that because the student learned the task or because it
learned the teacher? The pre-event half of the GDACS pool (Section~IV-A) lets us ask
it without any annotation: nothing a model marks on the 11{,}825 pre-event baselines
(Section~IV-A) can be inundation caused by the target GDACS event. The teacher marks water on \textbf{2{,}942} of
them (24.9\%).

\paragraph*{Spectral corroboration of the flagged baselines}
The teacher was fine-tuned on Sen1Floods11's \emph{surface-water} annotations
(Section~IV-A), and a permanent river is surface water in January as much as in
July. So before calling any of these predictions an invention we ask a question the teacher
cannot influence: \emph{is there water there at all?} We use MNDWI as an independent check of what the teacher detects on the pre-event baselines. We denote by $\rho_{\text{water}}$ the IoU between a model's predicted water mask and the corresponding binary $\mathrm{MNDWI}>0$ water mask. Because these images were acquired before the target flood event, water detected in them cannot represent inundation caused by that event. MNDWI confirms that most flagged baselines do contain substantial surface water: in $77.3\%$ of the 2{,}942 tiles, at least $10\%$ of the image is classified as water, while only $9.3\%$ contain almost no MNDWI water (water fraction below $0.02$).

The teacher predictions also largely coincide with these spectrally identified water areas. The median $\rho_{\text{water}}$ is $0.65$ on flagged baselines, compared with only $0.14$ on baselines where the teacher predicts no water. This suggests that the teacher is mostly detecting real surface-water features rather than producing arbitrary false positives. The problem is therefore mainly one of annotation semantics: permanent or seasonal water is a valid target for surface-water segmentation, but it should not be interpreted as new inundation caused by the target flood event.

\paragraph*{Student--teacher agreement against teacher-independent references}
The two teacher-independent checks give complementary results. Across the full pool, the
student's masks agree with the teacher's at IoU $0.624\pm0.024$ over four seeds, corresponding
to disagreement over approximately 38\% of the mask area. Scoring both models against two
teacher-independent references shows where that difference
helps from where it hurts. The first reference is $\rho_{\text{water}}$, defined above.
The second, $\rho_{\text{new}}$, targets the flood increment. From the same
$\mathrm{MNDWI}>0$ masks computed at the two dates, we take the water present at the event
date and absent in the pre-event acquisition of the same location, and score a model's mask
against exactly that difference. No annotation is involved, and the pre-event acquisition
acts as the reference-water layer. Two confounds limit how far this proxy can be pushed and
are the reason we use it only for a directional reading. Baselines were retrieved under a
stricter 20\% cloud gate than event-window tiles (50\%, falling back to 80\%) and no cloud
masking is applied at any stage, so unmasked cloud and cloud shadow enter the difference
asymmetrically. The paired acquisitions are 25--502 days apart (the baseline search window
is 30--395 days before the event start), so seasonal water change
enters it as well. Both push in the direction of over-counting ``new'' water, and neither
affects $\rho_{\text{water}}$, which is measured on a single date.
Relative to the teacher, the
student moves \emph{toward} both references on the 11{,}895 flood-date tiles ($+0.013$ and
$+0.019$) and \emph{away} from both on the 11{,}825 baselines ($-0.011$, $-0.012$). On the
2{,}942 baselines (teacher predicts flood) specifically it agrees with the teacher's call on
$\mathbf{96.1\pm1.0\%}$ over the four training seeds, at mask IoU $0.833\pm0.009$, while
raising a false alarm on only $1.9\pm0.8\%$ of the 8{,}883 baselines \chg{the teacher left empty}{on which the teacher
predicts less than 5\% water}, so
this is a targeted reproduction of one specific failure rather than a generally water-prone
model. \add{This false-alarm rate uses the teacher's own maps as the reference. Section~V-D of the main article measures
false alarms against independent references.}

The agreement is not explained by overlap with the training set. Of the 2{,}942 pre-event baselines on which the teacher predicts water, about one third are part of the $N\!=\!2{,}500$ training subset and the remaining two thirds are held out. The student's agreement with the teacher is almost identical in both groups: $96.0\pm0.6\%$ on training tiles and $96.2\pm1.2\%$ on held-out tiles. The student therefore reproduces the teacher's surface-water convention just as strongly on images it never saw during training.

The false-alarm analysis is even more clearly held out. None of the 8{,}883 baselines \chg{that the teacher leaves empty}{with less than 5\% predicted water} can enter the training subset, because the pre-filter requires a predicted flood fraction above $0.05$. The reported false-alarm rate is therefore measured entirely on unseen tiles.

A compact student need not reproduce every teacher error exactly: architecture,
regularization, and averaging across many scenes can make some deviations helpful on an
independent reference. The flood-date proxy comparison is consistent with that possibility,
but it does not establish a general zero-mean-noise theorem. The baseline result shows the
opposite case clearly: when the teacher's error is tied to a systematic annotation
convention, the student retains it almost intact.

\paragraph*{Consequences}
The audit rules out two simple fixes in this experiment. Relabelling the pool with the
student would reproduce some 96\% of the flagged baseline labels, so iterative self-training
does not address this semantic failure mode. A teacher-assistant experiment using a
four-seed UNet-ResNet50 Logit-KD ensemble to relabel the same pool was also null across the sweep;
Section~\chg{S8}{\ref{app:diag}} shows that the foundation model is the more accurate of the two on the pixels where
the labelers disagree. A direct way to address the semantic mismatch
is instead to use information that is independent of the teacher's current-event mask. The
paired pre-event acquisition provides exactly that reference, and the same construction
that defines $\rho_{\text{new}}$ could be used as a reference-water veto at inference. We
did not test that, so it is a suggestion rather than a result.

The dominant systematic error a distilled flood segmenter carries over here is therefore
not imagined water but the annotation semantics of the dataset the teacher was fine-tuned
on.
This is a semantic mismatch rather than an arbitrary false-positive mechanism. It can be
addressed at the labeling stage or with a reference-water layer at inference, rather than by
changing the distillation objective. None of our current benchmarks directly scores this failure
mode (Section~\chg{VII}{VII of the main article}). Any pipeline
that distills a surface-water foundation model into a flood product carries it, whether or
not it measures it.

\begin{deltext}
\revst{\textbf{Submitted Section S6, OFA Gradient-Path Analysis} (removed: the controlled re-run finds no OFA gain to explain).
The main article retains the empirical OFA window and the interpretation that its
auxiliary heads alter the optimization path without adding an independent label source.
The branch-level analysis is kept here.}

\paragraph*{\revst{OFA Gradient-Path Diagnostic}}
\revst{OFA adds branch losses at each encoder stage $i$:}
\[
  \mathcal{L}_\text{OFA} = \mathcal{L}_\text{KD}
    + \alpha_\text{OFA}\sum_i
      \mathbb{E}_x\,\ell_\text{br}(b_i(x), p_T(x)).
\]
\revst{Every branch is derived from the same frozen teacher output. The branches can therefore
change how gradients reach intermediate student stages, but they do not add an independent
annotation source. We do not assume that OFA and Logit-KD have identical parameter-space
minima; the claim tested here is narrower: OFA's advantage should appear as an optimization
effect whose size depends on the training regime.}

\revst{The additional gradient paths are}
\[
  \nabla_\theta \mathcal{L}_\text{OFA}
  = \nabla_\theta \mathcal{L}_\text{KD}
  + \alpha_\text{OFA}\sum_{i=1}^{L}
    \left(\frac{\partial F_i^\theta}{\partial \theta}\right)^\top g_i,
  \tag*{(old S1)}
\]
\revst{where $g_i=\partial \ell_\text{br}(b_i,p_T)/\partial F_i^\theta$. These terms provide
shorter supervision paths to intermediate stages instead of routing every update through
the full encoder and main head. This is the mechanism we examine below; the measurements,
not an assumption of a shared optimum, are the evidence for the interpretation.}

\paragraph*{\revst{Location of the OFA Difference}}
\revst{The gradient-conditioning argument makes one clean prediction: deep supervision can
only help once the branch targets themselves are stable. With too few scenes, each
branch fits its own unstable local flood signal and the shared backbone receives
contradictory updates, so in this regime OFA should bring no advantage. Figure~2 (main paper) is consistent
with this, and it locates the effect more precisely than ``scarce data'' does. At
$N\!=\!20$ the three objectives are tied on STURM within a span of $0.008$~mIoU (CE
$0.401$, Logit-KD $0.401$, OFA $0.409$); at $N\!=\!40$ OFA is \emph{behind} Logit-KD, by
$0.008$ on STURM and $0.016$ on WorldFloods, on three of four seeds in both cases. OFA's
advantage exists in a window rather than at the scarce end in general: from $N\!=\!50$ to
$N\!=\!251$ its seed mean is ahead on all three benchmarks, widest at $N\!=\!50$, and by
$N\!=\!2{,}500$ the three-objective spread has narrowed to $0.004$. The per-branch diagnostic
(submitted Fig.~S5, Section~S8) shows the branch gradients in
active conflict below that window. If the observed differences mainly reflect finite-sample optimization, they should
shrink as coverage grows. The data show that: the three-objective comparison spans 0.018~mIoU on
STURM at 252 scenes and 0.004 at 2{,}500 (Tables~I
and~II), with the ordering no longer resolvable at the larger pool.
This pattern is consistent with a finite-sample optimization effect rather than a
persistent advantage of the final supervisory target.}

\revst{Gradient surgery (PCGrad~\cite{yu2020gradient}) applied to the conflicting branches does not
help, which is consistent with the problem being broader than removable directional
conflict alone.
As the exact cosine magnitudes are sensitive to the averaging window, we report the
full diagnostic in Section~S8.}

\revst{In summary, the CE-to-OFA endpoint gap at the matched budget is driven mainly by the
CE-to-Logit-KD step; OFA's distinct advantage appears over the intermediate data window
and is gone at 2{,}500 scenes. The branch diagnostics are consistent with an optimization
effect rather than with access to a new supervision source. The practical ways to improve beyond this regime are therefore different from simply
adding more teacher-derived loss terms: expand scene coverage, improve the labeler, or
change the student's capacity or inductive bias. A stronger labeler only helps if its
advantage is expressed on the transfer pool itself; Section~S5 and
Section~S8 examine that distinction.}
\end{deltext}

\section{Deployment Implementation\add{, Acceptance} and Export Validation}
\label{sec:s_deploy}

\begin{newtext}
\paragraph*{Engines}
We compiled the Hardswish main student in FP32, FP16 and mixed INT8, and the tiny and micro students in FP32, FP16, mixed INT8 and
full INT8. With four training seeds per configuration, this comparison includes 44 engines. A full-INT8 main student cannot be built
(main article, Section~III-E). Every engine was evaluated on the device using the Sen1Floods11 validation and test
splits, the WorldFloods test split, STURM-Flood and the water-free windows. The additional engines of the deployed ReLU6 main student,
described below, were also evaluated on the held-back events. Table~IV of the main article reports each engine's efficiency.
Table~\ref{tab:s_device} reports its change in water IoU relative to the FP32 engine of the same seed, with a 95\%
interval obtained by resampling whole events. Figs.~\ref{fig:qual_tiny_precision} to~\ref{fig:qual_b0_precision} show,
for seed 42, the predictions in every number format on the three scenes of Fig.~3 of the main article.

\paragraph*{Main student (deployed, ReLU6)}
The deployed main student uses ReLU6 activations from its first training update, with the proposed setup and the same
initialization. QAT runs for ten epochs on the 2,500 teacher-labeled images. It uses W8A8 quantization, quantize--dequantize
operations on convolution inputs and weights, frozen batch normalization, a floating-point classifier and one warm-up
epoch. The checkpoint is selected on the Sen1Floods11 validation images. The engine runs 56 of its 57 convolutions in INT8 and its eight attention matrix products in FP16. All four seeds pass the
acceptance rule. On the validation images, their pooled mIoU changes by $-0.4$ to $+1.0$ points and their water IoU by
$-0.7$ to $+1.8$ points relative to the FP32 model of the same seed. Their files occupy 1.57 to 1.61\,MB.
False alarms on the water-free windows increase from 4.8\% to 9.6\% (Table~IV of the main article). This increase
arises during quantization-aware fine-tuning rather than from 8-bit arithmetic. When run in FP32 without quantizers,
the fine-tuned weights raise alarms on 12.5\% of the windows, ranging from 1.3\% to 22.2\% across the four seeds.
An FP16 engine of this student failed the output checks for every seed, so Table~IV contains no FP16 row for it.

QAT was not isolated as a single factor. For both main students, it changes together with calibration, rescaling and
export. It served as a recovery mechanism for models that failed PTQ alone.

\paragraph*{Main student with Hardswish}
The Hardswish student in Table~II of the main article is harder to quantize. Post-training quantization fails the
acceptance rule for all four seeds, so its INT8 engines use QAT. This required channel-balancing rescaling of convolution
inputs and weights, followed by seed-specific continuations of 384 to 1,664 updates. These engines retain attention,
normalization and the classifier in FP32. On the device, the mixed engine is slower than its FP32 counterpart (Table~IV
of the main article). Profiling attributes the extra time to scaling and format conversions around the quantized layers.

\paragraph*{Tiny and micro students}
Thirteen of their 16 INT8 engines use post-training quantization. For the other three, one tiny and two micro engines,
post-training quantization exceeded the acceptance limit, so they use the student's QAT model. The tiny engine required
only a correction to quantizer placement to explicitly quantize the outputs of residual additions. Its weights, previously
fine-tuned once by QAT, are unchanged. The micro model received 128 further QAT updates.
The micro student's full-INT8 engines embed its narrow layers in channels padded with zeros because the INT8 kernels
did not accept the narrow layers. This adds no learned capacity but makes the engines slower and larger.

\paragraph*{Acceptance}
An INT8 engine is accepted if it loses at most 2 IoU points relative to the FP32 engine of the same seed on the 89
Sen1Floods11 validation images. The limit applies to pooled mIoU, per-event mIoU and pooled water IoU. This is the accuracy
gap typically observed after 8-bit post-training quantization of convolutional networks~\cite{krishnamoorthi2018quantizing}.
All 20 INT8 engines of the Hardswish, tiny and micro students pass, and 17 also meet the stricter 1-point limit set
before the first device run. No test set was used to select or repair an engine.

\paragraph*{Timing, size and power}
Latency and its 99th percentile are measured with TensorRT's \texttt{trtexec} tool at batch size one. A 20\,ms warm-up
is followed by at least 200 timed runs over at least 3\,s. We report the compute time on the integrated accelerator, without host--device copies, and
the throughput in FPS from the same run. We average each engine's mean latency and p99 across seeds. Speedup is the ratio
of the FP32 latency to the engine's latency for the same seed. Engine MB is the compiled file size. Power is sampled approximately every 100\,ms on the module's combined rail of the CPU,
the integrated accelerator and the vision accelerators (\texttt{VDD\_CPU\_GPU\_CV}) during 30\,s of steady inference. The table reports the
average of the four per-seed means and the largest sample across the four runs. Clocks and power mode were unchanged.

\paragraph*{Graph and implementation costs}
The full-INT8 exports pad narrow layers to wider channels and upsample using dense transposed convolutions. Padding
converts some depthwise operations in the micro student into dense operations with up to 64 times the nominal
multiply--accumulate count. These costs arise from export rather than integer arithmetic. A prototype replaces the tiny
student's padded decoder with an integer decoder. For one trained model, its complete full-INT8 pipeline, from the
normalized image to the mask, runs in 1.26\,ms, compared with 1.32\,ms in FP16.

\end{newtext}
\begin{table*}[t]
\revnewfloat
\centering
\caption{\add{\textbf{Device accuracy relative to FP32.} Change in water IoU per event (points), averaged over four seeds,
with 95\% intervals from resampling whole events. Each engine is compared with the FP32 engine of the same student
and seed. Shading identifies the two deployed configurations, as in Table~IV of the main article. The last two rows
of the ReLU6 block instead use the Hardswish student's FP32 engine as the reference. On the held-back events, the
deployed engine changes water IoU by $+0.4$ $[-0.4, +1.2]$ points relative to its FP32 engine.}}
\label{tab:s_device}
\footnotesize
\begin{tabular}{llccc}
\toprule
Student & Format & S11 test & WorldFloods test & STURM-Flood \\
\midrule
\rowcolor{useGPU}Main, ReLU6 & mixed INT8/FP16 & +2.0 [+0.3, +3.8] & +0.9 [$-$0.2, +2.3] & $-$1.7 [$-$4.3, +0.3] \\
(deployed) & FP32 $-$ Hardswish FP32 & $-$2.6 [$-$6.8, +0.1] & $-$0.7 [$-$1.9, +0.2] & $-$0.6 [$-$1.1, 0.0] \\
 & mixed INT8/FP16 $-$ Hardswish FP32 & $-$0.6 [$-$4.2, +2.7] & +0.2 [$-$0.7, +1.1] & $-$2.2 [$-$5.0, $-$0.2] \\
\midrule
Main, Hardswish & FP16 & +0.1 [0.0, +0.2] & 0.0 [0.0, +0.1] & 0.0 [0.0, +0.1] \\
 & mixed INT8 & $-$0.6 [$-$2.1, +0.8] & $-$0.5 [$-$1.5, +0.1] & $-$3.3 [$-$5.8, $-$1.2] \\
 & full INT8 & \multicolumn{3}{l}{not possible: the attention needs floating point} \\
\midrule
Tiny student & FP16 & 0.0 [0.0, 0.0] & 0.0 [0.0, 0.0] & 0.0 [0.0, 0.0] \\
 & mixed INT8 & $-$0.3 [$-$0.4, $-$0.1] & $-$0.2 [$-$0.3, $-$0.1] & $-$1.2 [$-$2.1, $-$0.5] \\
\rowcolor{useINT} & full INT8 & 0.0 [$-$0.4, +0.3] & $-$0.1 [$-$0.3, +0.2] & $-$0.5 [$-$1.0, 0.0] \\
\midrule
Micro student & FP16 & 0.0 [0.0, 0.0] & 0.0 [0.0, 0.0] & 0.0 [0.0, 0.0] \\
 & mixed INT8 & +1.2 [+0.1, +2.9] & $-$0.3 [$-$1.0, +0.2] & $-$1.2 [$-$2.2, $-$0.3] \\
 & full INT8 & +1.1 [0.0, +2.8] & $-$0.3 [$-$0.9, +0.1] & $-$1.3 [$-$2.4, $-$0.4] \\
\bottomrule
\end{tabular}
\end{table*}

\subsection*{Quantization and export details}
The main article states the 8-bit weights-and-activations (W8A8) quantization-aware training
(QAT) \chg{recipe}{setup}, the quantize--dequantize (QDQ) placement, the frozen batch-normalization (BN)
parameters, the static export shape, and the mixed \chg{8-bit integer (INT8)/16-bit floating-point (FP16)}{INT8/floating-point}
engine. The details below
record the graph mechanics and numerical failure modes.

\paragraph*{Quantization-aware training}
We apply QAT on the \chg{pseudo-label dataset}{2,500 teacher-labeled images} using W8A8 quantization. During QAT, fake-quantize nodes are inserted at the input of
each quantizable operation: a \emph{QuantizeLinear}/\emph{DequantizeLinear} (QDQ)
pair simulates the quantization error during the forward pass using the formula
$\hat{x} = \mathrm{clamp}(\mathrm{round}(x / s),\,-128,\,127) \times s$, where
$s$ is the per-tensor calibration scale. The backward pass computes gradients
through the straight-through estimator; the model learns to minimise both
segmentation loss and the representation error introduced by the fixed-scale
quantization.

Two non-obvious decisions had to be made to make this work without accuracy collapse:

\textbf{(1) QDQ only on convolution inputs and weights, never on outputs.}
On an intermediate layer
a second quantization of the same tensor is redundant, while on the output head it is
actively harmful, because there the int32 accumulator really is rescaled into int8
with a background class-dominated scale. By
omitting output QDQ we leave the final logits in float (the int32 accumulator is
dequantized via the input$\times$weight scales), so the flood range survives; on
intermediate layers it simply avoids a redundant node.

\textbf{(2) Frozen BatchNorm affine parameters during QAT.}
TensorRT folds each BatchNorm into its preceding convolution, producing an effective
weight $w' = w\,\gamma/\sqrt{\sigma^2+\epsilon}$ and a folded bias
$b' = \beta - \mu\,\gamma/\sqrt{\sigma^2+\epsilon}$. The per-tensor activation scale of
that fused Conv is set by the largest-magnitude channel of its output. If the BN
affine parameters $\gamma,\beta$ keep training during QAT, a few channels' folded bias
$b'$ can drift to large negative values, inflating the layer's amax and hence its INT8
scale by up to $9\times$; every other channel in that tensor is then quantized against
an over-large scale, reducing the effective resolution of the remaining channels. Freezing
$\gamma$ and $\beta$ and the BatchNorm (BN) running statistics during
quantization-aware training (QAT) keeps the folded bias stable, so only the INT8 scales adapt.

\paragraph*{Export and compilation}
Export follows the standard PyTorch route: \texttt{torch.onnx.export} on the
QAT model, followed by \texttt{onnx-simplifier} for constant folding. This is the
common practice for QDQ deployment to TensorRT (NVIDIA's \texttt{pytorch-quantization}
toolkit and the TensorRT developer guide follow the same pattern). We emit three
Open Neural Network Exchange (ONNX) files: a BN-folded float baseline, a simplified static-shape graph for 32-bit floating-point (FP32)/FP16
compilation, and the QDQ graph for INT8. One non-standard step is needed. When
\texttt{torch.onnx.export} is run with dynamic batch/spatial axes, the graph carries
the input shape symbolically through \texttt{Shape}$\to$\texttt{Gather}$\to$%
\texttt{Concat}$\to$\texttt{Reshape} subgraphs operating on INT64 shape tensors
(``INT64 shape chains''). TensorRT 8.4 either rejects these or builds an unnecessary
dynamic-shape engine. Our shape-fixing pass pins the input to a static
$1\!\times\!6\!\times\!512\!\times\!512$ and constant-folds those INT64 subgraphs into
fixed dimensions, leaving a clean static graph for the builder.

The LiteMLA attention MatMul layers in EfficientViT remain in FP16 throughout the
INT8 engine. These layers compute a linear-attention accumulator whose dynamic range
is 200--400, far too large for INT8 at per-tensor granularity: at that range each INT8
bin spans $\approx$2.4, leaving only 1--5 distinct values for the attention normalizer
channel and degrading the attention computation. INT16$\times$INT16$\to$INT32 accumulation
would comfortably hold that range, but it is not available: TensorRT 8.4 exposes only FP32,
FP16 and INT8 as builder precisions
and has no INT16 GEMM tensor type, so there is no INT16 kernel for the builder to
select on this hardware. INT8 being numerically unsafe and INT16 unavailable, FP16 is
the only viable precision for these eight MatMuls; the resulting engine is a
mixed INT8/FP16 graph.
\begin{newtext}
For the deployed ReLU6 student, the layer inspector confirms that attention runs in FP16. The Hardswish engines retain
these matrix products, their normalization and the classifier in FP32 because that student's attention numerator
overflowed in FP16.
\end{newtext}

Engine compilation must be performed on the target device itself: using \chg{TRT}{TensorRT}\,8.4 on the Jetson Xavier NX (JetPack~5.0) cannot transfer its engine to a different architecture target. The compiled engine is not transferable between device types.

\begin{oldtext}
\subsection*{\revst{Deployment accuracy and calibration}}
\revst{All float and quantized accuracy values use the same four training seeds (42, 64, 91, 107) and
the same teacher-supervised pool. ReLU6 retraining is the architecture change required for the
TensorRT path; quantization-aware training is then applied to that model. Activation
calibration and checkpoint selection use the existing Sen1Floods11 validation split, so no
additional manual annotation is acquired and no ground-truth label contributes a training
gradient to the student.}

\revst{Table~III reports the deployment stages scored with the workstation evaluator and the two exported models scored through the on-device inference pipeline. Both paths use the same benchmark definitions and ignore masks. The higher WorldFloods-v2 mIoU of the QAT float export is consistent with a regularization effect from QAT and is not used as a primary accuracy claim.}

\revst{The quantization effect is evaluated within the device pipeline itself. INT8 changes mIoU from
$0.7557\pm0.0062$ to $0.7448\pm0.0236$ on STURM-Flood and from
$0.7861\pm0.0105$ to $0.7702\pm0.0214$ on WorldFloods-v2, corresponding to drops of
$0.0110$ and $0.0158$ mIoU, respectively. The larger seed spread of the INT8 engines is reported as measured.
The deployment claim is restricted to clean accuracy plus measured engine characteristics. No
corruption-robustness recovery is attributed to quantization.}

\subsection*{\revst{Export validation and profiling}}
\revst{Each TensorRT export is validated against its corresponding float model before device
reporting. The profiling protocol uses batch size one with input
$1\times6\times512\times512$, 20 warm-up iterations and 200 timed iterations. Latency, engine
size and power are properties of a compiled engine rather than of the trained weights, so they
are hardware-build measurements and must not be read as training-seed statistics: the main
article reports one representative compiled engine per precision. For the deployed INT8 engine
these are a 1.5\,MB engine, 5.57\,ms of graphics processing unit (GPU) compute, and 13.9--14.2\,MB of device memory
(1.8\,MB persistent, 4.0\,MB scratch, 8.1--8.4\,MB activations); host-to-device and
device-to-host transfers add a further 0.36--0.39\,ms. FP16 is the faster precision on this
Xavier NX configuration, while INT8 gives the smallest engine and the lowest peak GPU power.}
\end{oldtext}

\section{Additional Diagnostics and \chghead{Ablations}{Student Designs}}
\label{app:diag}

This section collects the analyses that support design choices and secondary claims in
the main article: the student--teacher representation alignment that motivates the absence
of a feature-matching loss, the pixel-level imitation diagnostics, \del{the per-branch gradient
measurement behind the optimization reading of OFA,} the spectral-proxy experiment, and the
\chg{architecture ablation behind the choice of student}{comparison of three student designs under three objectives}. \add{The first three
analyses use the original experimental setup.}

\subsection*{Representation Alignment via Centered Kernel Alignment\add{ (original experimental setup)}}
\label{sec:cka}

This analysis asks whether the compact convolutional neural network (CNN) develops internal
features similar to those of the foundation model despite the Vision Transformer (ViT)-to-CNN
architecture gap. Centered Kernel Alignment (CKA)~\cite{kornblith2019similarity} scores the similarity of two feature sets extracted
from the same inputs, returning $1$ when they agree up to an orthogonal rotation and a
rescaling and near $0$ when they are unrelated; that invariance is what lets it compare
tensors of different channel count and resolution. We use the linear estimator and
\emph{spatial} CKA, with features compared on an $8\!\times\!8$ grid and each location treated as a
sample. This is both the right question for a dense-prediction model and far more
stable than pooling each image to one vector.

We compute a $4\!\times\!4$ matrix per training condition between the four encoder stages of
EfficientViT-B0 and the matched neck levels of Prithvi-EO-2.0, over 100 fixed Sen1Floods11
probe batches, and report the mean of its diagonal.\footnote{Ten conditions: the
ImageNet-initialized backbone without task training, and the three objectives on Sen1Floods11 labels
at $N\!=\!252$ and on pseudo-labels at $N\!=\!252$ and $N\!=\!2{,}500$.} Each condition is
measured at all four training seeds. Within a seed all ten conditions share one probe loader,
one probe ordering and one teacher checkpoint; \emph{across} seeds the probe seed is held
fixed and only the student checkpoint changes, so the model is the random variable and the
instrument is not. The untrained baseline is identical to four decimal places in all four
runs ($0.1866$), which is the check that this holds. Absolute CKA values are probe-dependent
and should not be compared across papers.

\begin{figure*}[t]
\centering
\includegraphics[width=0.7\linewidth]{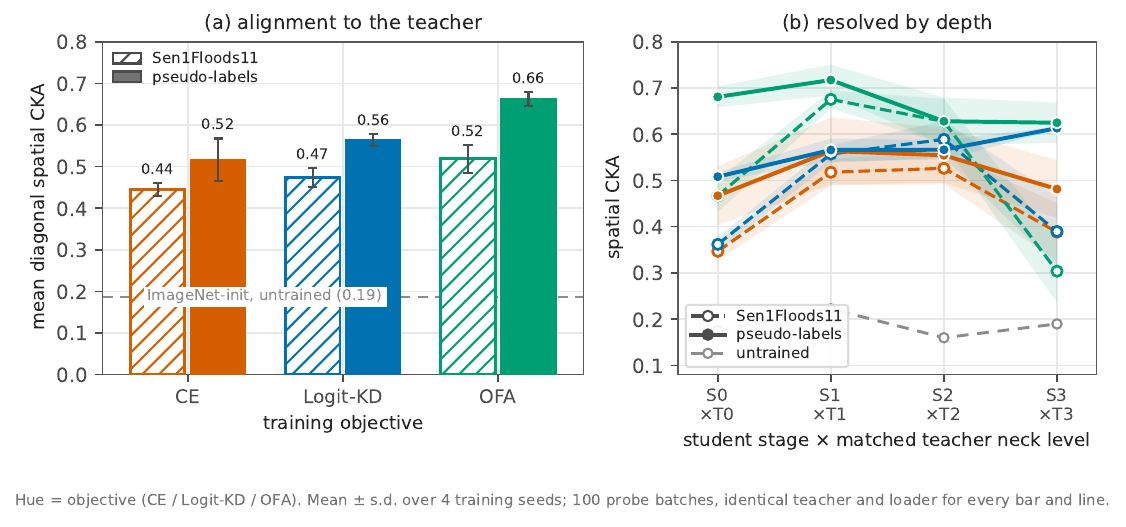}
\caption{%
  \textbf{Representation alignment by objective and label source\add{ (original experimental setup)}.}
  (a) Mean diagonal spatial CKA between EfficientViT-B0 and Prithvi-EO-2.0; the
  CE\,$\to$\,Logit-KD\,$\to$\,OFA ordering is monotone in the seed mean for both label
  sources. The Sen1Floods11 bars use $N\!=\!252$ and the plotted pseudo-label bars use
  $N\!=\!2{,}500$; the matched-budget pseudo-label values are reported in the text.
  (b) The same quantity resolved by depth, which is where the two label sources part company.
  Mean $\pm$ s.d. over four training seeds, identical teacher and probe for every bar.
}
\label{fig:cka}
\end{figure*}

Figure~\ref{fig:cka} supports three observations.

\emph{Alignment increases with the objective for both label sources.} Mean diagonal CKA
rises with the objective in every case: $0.445 \to 0.474 \to 0.518$ on Sen1Floods11 labels,
$0.390 \to 0.435 \to 0.490$ on pseudo-labels at the matched budget and
$0.517 \to 0.563 \to 0.663$ at $N\!=\!2{,}500$, against $0.187$ untrained. The endpoints
are the reliable part: OFA exceeds CE in $4/4$ seeds in all three objective comparisons, while the
individual objective steps are not all resolved at $n\!=\!4$, so we claim the ordering of the
endpoints and not a gap at each step.

\emph{The label-source ordering in alignment reverses once the budget is released.} At the matched budget
the pseudo configuration is \emph{behind} Sen1Floods11 labels at all three objectives ($-0.055$, $-0.039$,
$-0.029$); released to the full pool it is ahead at every objective and in every seed
($+0.072$, $+0.089$, $+0.144$). So a teacher-generated label is not intrinsically
better-aligned than a human one at equal count. At equal count it is measurably worse, as expected for a noisier label,
but it can be produced in quantities a human
label cannot, and alignment keeps buying with quantity where the accuracy curve of
\chg{Section~IV-E}{the earlier 224-pixel study} has already flattened. The matched-budget deficit also shrinks
monotonically as the objective strengthens, so distillation recovers part of what the label
noise costs in representation space.

\emph{The difference is concentrated in the deep stages.} Panel~(b) is the more informative view and
the most robust result here. Hand-label training aligns the shallow and middle stages and
then decreases at the deepest stage. The OFA configuration reaches $0.675$ at stage~1 and
falls to $0.304$ at stage~3, and both other Sen1Floods11 configurations show the same pattern, while
pseudo-label training at $N\!=\!2{,}500$ holds $0.63$--$0.72$ across all four stages. The
stage-3 gap of $0.32$ is present in every seed. The largest label-source difference therefore
occurs at the deepest measured stage, which is consistent with limited late-stage alignment
under the smaller training distribution. CKA is a representation-similarity diagnostic, so
we treat this as supporting evidence rather than a causal explanation of the OOD gap.

Two conclusions follow. First, the alignment ordering tracks the accuracy ordering of
\chg{Table~I}{the earlier 252-image comparison} at the level of the endpoints, and both flatten at the deployment
budget; we do not push the correspondence to the individual objective steps, since the alignment steps
and the accuracy steps do not rank the two objective steps the same way. Second, and this is why we do
not use feature distillation: the student already reaches high deep-stage alignment through
logit-level supervision alone, so an explicit feature-matching loss has little alignment left
to add. \del{MSE matching between intermediate student and teacher features did not improve results in
our experiments.} Because \chg{that loss}{a feature-matching loss} is teacher-derived, the proposition of
Section~\ref{sec:ceiling} says only that it adds no independent label source; it does not predict \chg{the null result}{its effect}.
\del{The per-term gradient-cosine logs also show no consistent conflict between feature and task
losses (cosines are near zero and roughly balanced in sign), so the simplest empirical
reading is that the added feature objective was redundant in this setup.}

Two properties of the full matrices behind these diagonal means are worth recording. The
objective ordering appears across the full matrix rather than on the diagonal alone: all
sixteen teacher-neck\,$\times$\,student-stage cells rise from CE to OFA, and in every
trained condition on both pools the strongest single entry is \emph{off-diagonal}, at
teacher T0 $\times$ student S2 ($0.784$ for CE, $0.794$ for Logit-KD and $0.810$ for
OFA). The student's third stage is the layer that best reproduces the teacher's first
neck level, and the alignment stays high without peaking again deeper. This depth
compression is consistent with the mismatch in capacity between a four-stage $0.7$\,M encoder
and a $300$\,M ViT neck. One structural limit belongs here rather than being left to be
noticed: CKA as we compute it needs discrete backbone stages to pair with the teacher's
neck levels, so the \textbf{UNet-ResNet50 capacity reference cannot be measured this way
at all} and is absent from this analysis; nothing in this subsection should be read as a
statement about it.

\subsection*{Imitation Diagnostics\add{ (original experimental setup)}}

This part collects the two pixel-level diagnostics referenced in \chg{Section~V and
Section~VII}{the main article}. Both are evaluated on STURM-S2 at
the paper's OOD protocol (512$^2$, BN eval mode, identical adapter input for all
models).

\paragraph*{Disagreement audit}
Restrict to the valid pixels where a student and the Prithvi teacher predict
different classes (2.5--2.9\% of pixels); the task is binary, so one of
the two matches the ground truth. A student that \emph{corrects} the teacher wins
more than half of these pixels; a student whose deviations are its own imitation
errors loses them.

\begin{table}[t]
\centering
\caption{%
  Disagreement audit on STURM-S2 against the Prithvi teacher\add{ (original experimental setup)}; students are the
  $N\!=\!2{,}500$ pseudo-label configurations of \chg{Table~II}{the earlier 2,500-image comparison}, 4 seeds each.
  Student-right = share of disagreement pixels where the student, not the teacher, matches
  ground truth (mean\,$\pm$\,std over seeds); above 50\% means the student corrects its
  teacher more often than not.
}
\label{tab:audit}
\small
\setlength{\tabcolsep}{4pt}
\resizebox{\columnwidth}{!}{%
\begin{tabular}{lcc}
\toprule
Student & student right & pixel acc (teacher: 0.892) \\
\midrule
\emph{EfficientViT-B0, OFA (0.7\,M)} & $\mathbf{53.8 \pm 3.5\%}$ & $\mathbf{0.895}$ \\
EfficientViT-B0, Logit-KD (0.7\,M)   & $52.6 \pm 0.8\%$          & $0.894$ \\
UNet-ResNet50, Logit-KD (32.5\,M)    & $44.1 \pm 3.7\%$          & $0.889$ \\
\bottomrule
\end{tabular}}
\end{table}

Both compact students are correct slightly more often than the teacher on pixels where they disagree: OFA on $53.8\%$ of these pixels and Logit-KD on $52.6\%$. OFA also has the highest overall pixel accuracy of the three models in this audit.

The 32.5\,M UNet is correct on only $44.1\%$ of its disagreement pixels. This is consistent with its lower STURM score in \chg{Table~II}{the earlier 2,500-image comparison} and suggests that, at this pool size, it follows the teacher more closely rather than improving on the task itself.

The advantage of the compact students is small, so we use this audit only to indicate the direction of the effect. It suggests that OFA and Logit-KD can sometimes correct teacher errors, but it does not provide a precise estimate of how large that advantage is. The audit is therefore only an indirect check and cannot replace STURM ground truth on the disagreement pixels. Section~\chg{IV-G}{V-E of the main article} examines the same question using teacher-independent reference signals.

\begin{delfloat}
\begin{figure}[h]
\revdelfloat{FIG. S4}
\centering
\includegraphics[width=\linewidth]{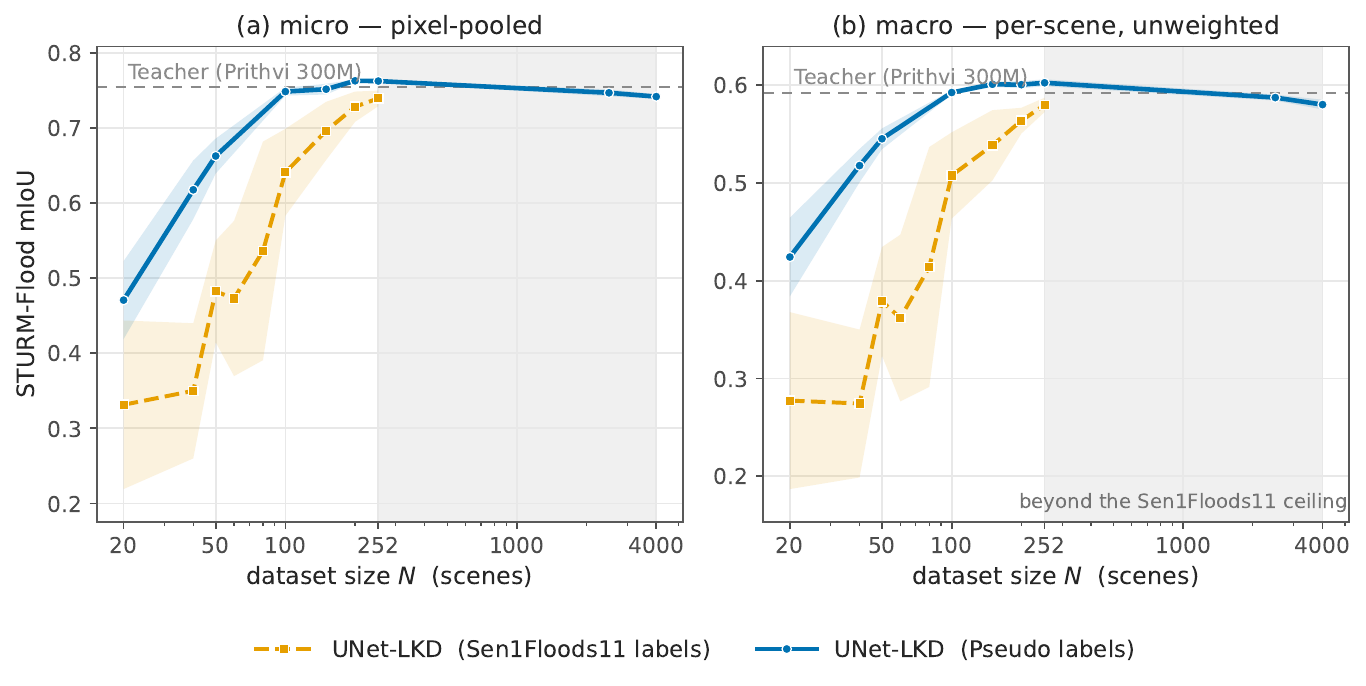}
{\footnotesize \revst{Submitted Fig.~S4. \textbf{Architecture control for Fig.~4}: the same comparison with a
  32.5\,M UNet-ResNet50 under logit distillation instead of a 0.7\,M EfficientViT-B0
  under OFA. The gap is present throughout and is much wider at small $N$, where the
  Sen1Floods11 configuration's seed spread is large. Routing the label budget is therefore not a property
  of our student. The $N=4{,}000$ run is included only as a scaling diagnostic; the UNet curve declines beyond N=252, consistent with Section~IV-C.}\par}
\end{figure}
\end{delfloat}

\paragraph*{Two-labeler audit}
This audit evaluates the UNet-ResNet50 ensemble used in the teacher-assistant experiment
(Section~\chg{S5}{\ref{sec:s_semantics}}). EfficientViT-B0 is the final student architecture in this study rather than a
separate intermediate-capacity assistant, so the UNet ensemble provides the distinct assistant
model in this test. We examine the 4.2\% of STURM pixels where the Prithvi teacher and the
UNet-ensemble relabeler disagree, because only those pixels distinguish the two labeling sources.
On that slice the foundation model is the more accurate of the two: it matches ground truth
60.9\% of the time against the ensemble's 39.1\%. The premise of the teacher-assistant
chain~\cite{mirzadeh2020improved}, in which the intermediate model should provide the better
teacher, therefore does not hold in this setting, consistent with the null relabeling result.

The students behave consistently with this. At the full pool they side with the UNet ensemble
on only $39.5\pm6.8\%$ and $40.4\pm4.8\%$ of those pixels across the two configurations tested:
the directly distilled student, trained on pool labels generated by the Prithvi teacher, and
the teacher-assistant student, the same architecture and objective trained on pool labels
written by the four-seed UNet-ResNet50 Logit-KD ensemble. Swapping the labeler moves this rate by
0.9 percentage points, within the seed spread. In the scarce regime ($N\!=\!150$), both
configurations are close to chance level ($53.2\pm6.7\%$ and $52.2\pm5.9\%$). In neither
regime does the assistant's output displace the foundation model as the effective supervisor.

\begin{deltext}
\revst{\textbf{Submitted subsection: OFA Inter-Branch Gradient Conflict} (removed with the gradient-path analysis).
This analysis supports the optimization interpretation of OFA with a per-branch gradient
diagnostic. The OFA step is not statistically resolved at the matched budget under either
label source (Section~IV-C), while objective differences are also small at the deployment budget.
The diagnostic therefore does not establish that OFA always helps; it shows that the
deep-supervision gradients are more internally conflicting when the training set is small and
become more consistent as it grows.}

\revst{We decomposed the OFA gradient into per-term components
$\{g_\text{main}, g_{\text{ofa},i}\}_{i=0}^{3}$, where $g_\text{main}$ is the gradient of
the main head's hard-label (focal) term and $g_{\text{ofa},i}$ are the four branch losses,
and recorded pairwise cosine similarities during training at $N\!\in\!\{100,300,500\}$
(2 seeds each; submitted Fig.~S5). The direction of the effect is clear and
reproducible, though the exact cosine magnitudes depend on the averaging window and should
be read qualitatively. At $N\!=\!100$ the branch gradients are in conflict: early-stage
branches (receptive field 19--43\,px) have \emph{negative} cosine both with the late
branches (RF 251--667\,px) and, more strongly, with \emph{each other}. With only 100
scenes, the branch targets are less stable and the shared backbone receives more conflicting
updates. By $N\!=\!300$, all inter-branch cosines are positive, and they increase further by
$N\!=\!500$. We describe this simply as a change in gradient consistency. The transition in the gradients ($N$ between 100
and 300) and the threshold read off accuracy (Fig.~2: OFA is behind
Logit-KD at $N\!=\!40$ and ahead from $N\!=\!50$) are not the same number. At $N\!=\!100$
the branches measurably conflict and OFA is nonetheless $+0.010$~mIoU ahead of Logit-KD on
STURM, on three of four seeds. So branch conflict does not abolish the deep-supervision
benefit wherever it is present; what the diagnostic establishes is the weaker claim that
the conflict is real, that it resolves with $N$, and that it is largest exactly where OFA
stops being the better objective. We do not claim it is the sole cause of that reversal.
Where conflict is present, the two main-head terms remain positively aligned and continue
to drive learning. The logs therefore establish the presence and disappearance of branch
conflict with $N$, but not a unique causal role for that conflict in the accuracy curve.}

\revst{Gradient surgery (PCGrad~\cite{yu2020gradient}) applied to the OFA branch set produced no
measurable improvement. In this experiment, resolving gradient direction conflicts was
therefore not sufficient to improve the small-$N$ result.}

\revst{Read together with Section~IV-C, this bounds the objective effect on both sides: at smaller $N$, branch conflict can attenuate the benefit
of deep supervision; it does not eliminate it, because OFA is already ahead of Logit-KD at $N=100$.
At larger $N$, the objective differences shrink to the point that they are no longer resolved. The window in between
is where the objective differences in Table~I occur.}
\end{deltext}

\begin{delfloat}
\begin{figure}[h]
\revdelfloat{FIG. S5}
\centering
\includegraphics[width=0.7\linewidth]{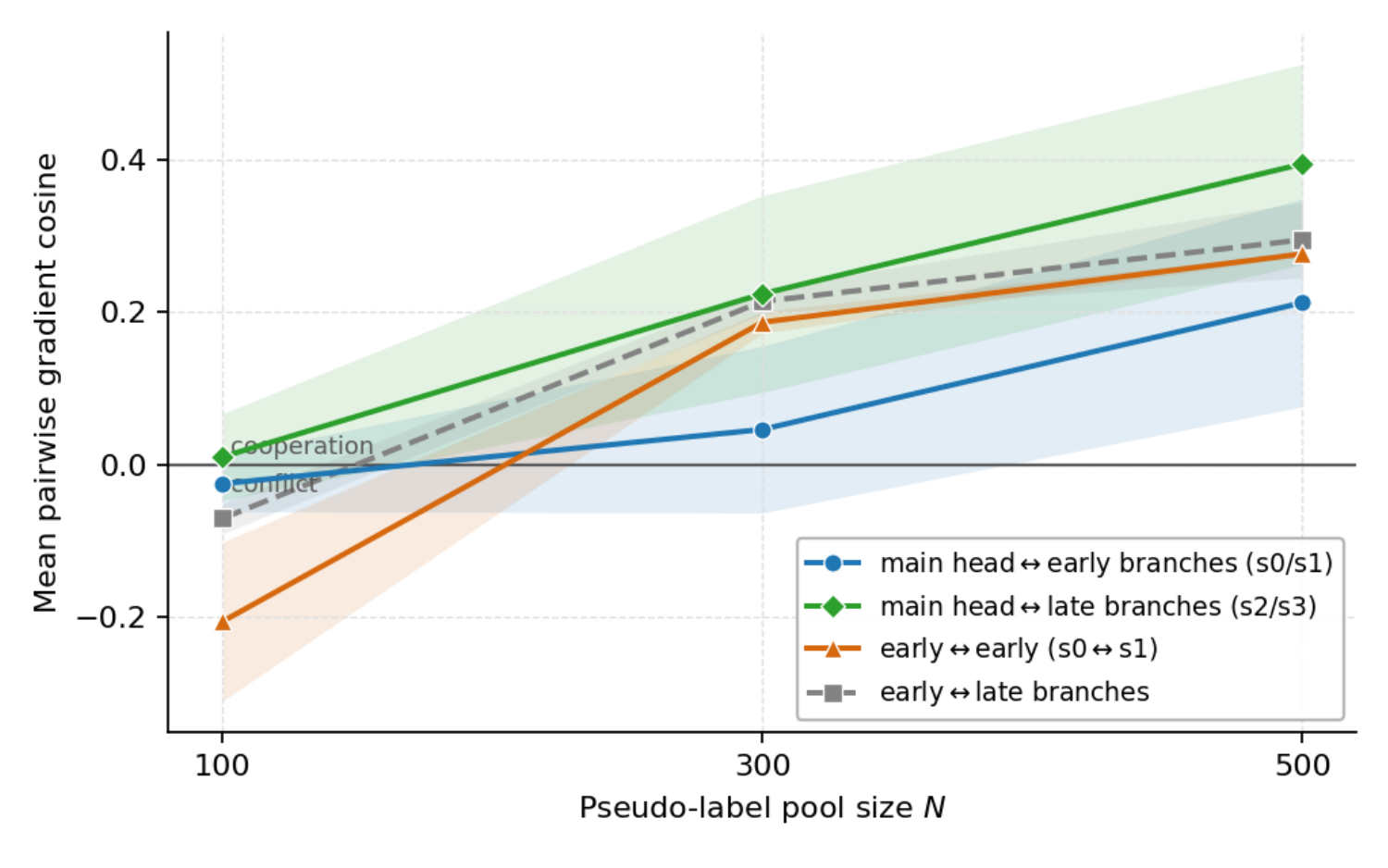}
{\footnotesize \revst{Submitted Fig.~S5. Inter-branch gradient conflict in OFA versus pseudo-label pool size
  ($N\!\in\!\{100,300,500\}$, 2 seeds; shaded bands = seed range). Negative cosine
  = conflicting gradient directions in the shared backbone. At $N\!=\!100$ the
  early branches conflict with the late branches ($-0.07$) and, most strongly, with
  each other ($-0.21$); by $N\!=\!300$ every pair has flipped positive, and
  cooperation strengthens further by $N\!=\!500$. Exact magnitudes are
  averaging-window dependent; the sign change is the robust finding.}\par}
\end{figure}
\end{delfloat}

\subsection*{\chghead{Architecture Ablation}{Three Student Designs under Three Objectives}}
\label{sec:archablation}

\begin{oldtext}
\revst{To check that EfficientViT-B0 is a reasonable architecture choice (not merely the
one we tuned most), we trained two additional compact CNN students on the same
curated pool. To keep the comparison fair, we evaluate every architecture under
the two objectives they all support (plain CE and logit-KD) and report OFA
only for our own EfficientViT-B0 student, since OFA is the method we propose rather
than a general baseline. We flag the missing OFA variants as a scoping decision, not a
result: we cannot claim OFA would fail to help the other two, only that we did not
test it.}
\end{oldtext}
\begin{newtext}
Table~\ref{tab:arch} compares EfficientViT-B0, MobileNetV3-Small and LightFormer-ResNet18 under three objectives:
labels only, logit distillation and OFA. All use the same 2,500 teacher-labeled images, image order, training setup,
number of updates and four seeds. The OFA branches for the two additional designs attach to their encoder stages in
the same way as for EfficientViT-B0. MobileNetV3-Small also appeared in an earlier ablation with a different training setup.
No design reliably benefits from OFA over logit distillation (main article, Table~III). At 2,500 images, the only
interval above zero is for LightFormer on the Sen1Floods11 test split, and this comparison does not survive Holm
correction over the 39 OFA comparisons. Logit distillation increases the mean score of every design relative to labels
only on WorldFloods and STURM-Flood. The designs share a training setup but use different pretrained weights.
EfficientViT-B0 starts from a Cityscapes segmentation model, whereas the others start from ImageNet encoders.
\end{newtext}

\begin{delfloat}
\begin{table}[t]
\revdelfloat{TABLE S3}
\centering
{\footnotesize \revst{Submitted Table~S3. Architecture ablation at a full curated pool, 4 seeds (mean\,$\pm$\,std).
  \emph{Small CNN} is a 1.64\,M convolutional segmentation baseline of our own, trained
  from scratch. $\star$ = no ImageNet pretraining. Italics mark our model; bold marks
  the best value. All rows share one recipe, so comparisons between rows are valid, but the
  absolute values are not comparable with Tables~I and~II.}\par}
\small
\setlength{\tabcolsep}{3pt}
\resizebox{\columnwidth}{!}{%
\begin{tabular}{llllll}
\toprule
Architecture & Params & Pretrain & Loss & S11 mIoU & STURM mIoU \\
\midrule
MobileNetV3-Small & 3.55\,M & ImageNet & CE         & $0.868 \pm 0.013$ & $0.734 \pm 0.020$ \\
MobileNetV3-Small & 3.55\,M & ImageNet & Logit-KD   & $0.872 \pm 0.002$ & $0.746 \pm 0.008$ \\
Small CNN         & 1.64\,M & None$^\star$ & CE      & $\mathbf{0.887 \pm 0.004}$ & $0.737 \pm 0.037$ \\
Small CNN         & 1.64\,M & None$^\star$ & Logit-KD & $0.882 \pm 0.002$ & $0.739 \pm 0.033$ \\
\midrule
EfficientViT-B0   & 0.7\,M & ImageNet & CE          & $0.861 \pm 0.008$ & $0.737 \pm 0.032$ \\
EfficientViT-B0   & 0.7\,M & ImageNet & Logit-KD    & $0.869 \pm 0.006$ & $0.754 \pm 0.023$ \\
\emph{EfficientViT-B0} & 0.7\,M & ImageNet & \emph{OFA (ours)} &
  $0.870 \pm 0.003$ & $\mathbf{0.760 \pm 0.014}$ \\
\bottomrule
\end{tabular}}
\end{table}
\end{delfloat}

\begin{table*}[t]
\revnewfloat
\centering
\caption{\add{\textbf{Three student designs under three objectives.} All use 2,500 teacher-labeled images at
$512\!\times\!512$ and four seeds. Scores are water IoU per flood event (points, mean $\pm$ standard deviation across
seeds). The labels-only objective uses focal loss at full weight on the teacher's hard labels, whereas Table~III of the main
article keeps the hard-label weight at 0.5 in both setups. Bold marks the best objective for each design and column.}}
\label{tab:arch}
\footnotesize
\begin{tabular}{lllccc}
\toprule
Design & Parameters & Objective & S11 test & WorldFloods test & STURM-Flood \\
\midrule
EfficientViT-B0 & 0.7\,M & labels only & 74.7 $\pm$ 1.0 & 78.4 $\pm$ 0.9 & 47.5 $\pm$ 2.2 \\
 &  & logit distillation & 76.1 $\pm$ 0.6 & 79.7 $\pm$ 0.2 & \textbf{50.1} $\pm$ 0.8 \\
 &  & OFA & \textbf{76.3} $\pm$ 1.0 & \textbf{79.9} $\pm$ 0.2 & 49.5 $\pm$ 0.7 \\
\midrule
MobileNetV3-Small & 3.55\,M & labels only & \textbf{77.1} $\pm$ 1.3 & 78.3 $\pm$ 1.4 & 48.7 $\pm$ 0.7 \\
 &  & logit distillation & \textbf{77.1} $\pm$ 0.6 & 79.6 $\pm$ 0.2 & 51.4 $\pm$ 0.3 \\
 &  & OFA & 76.7 $\pm$ 0.9 & \textbf{79.8} $\pm$ 0.3 & \textbf{51.6} $\pm$ 0.5 \\
\midrule
LightFormer-ResNet18 & 11.43\,M & labels only & 75.0 $\pm$ 1.4 & 75.6 $\pm$ 2.5 & 45.0 $\pm$ 1.5 \\
 &  & logit distillation & 75.9 $\pm$ 1.3 & 78.7 $\pm$ 0.6 & \textbf{49.3} $\pm$ 1.1 \\
 &  & OFA & \textbf{76.7} $\pm$ 0.6 & \textbf{78.9} $\pm$ 0.3 & 49.0 $\pm$ 0.7 \\
\bottomrule
\end{tabular}
\end{table*}

\begin{oldtext}
\revst{We read Table~S3 conservatively. Under the objectives that \emph{all}
backbones support the three architectures are within 0.015 mIoU of each other under each shared objective on STURM,
of the order of their seed spreads, so no architecture dominates on a like-for-like
objective. EfficientViT-B0 with OFA is the best row by $+0.014$ over the strongest
alternative configuration (MobileNetV3-Small with Logit-KD) at $5\times$ fewer
parameters. The only backbone without ImageNet pretraining is also the most
seed-unstable ($\pm0.037$), consistent with the initialization mattering more than the
architecture at this scale. The claim we make from this table is narrow:
EfficientViT-B0 with OFA is the strongest \emph{deployable} compact configuration we
found, and the smallest.}
\end{oldtext}

\begin{table*}[t]
\revnewfloat
\centering
\caption{\add{\textbf{Six students in five setups, and LightFormer with its authors' setup.} Water IoU per flood event
(points, mean $\pm$ standard deviation over four seeds) and false alarms on 200 water-free WorldFloods windows
(\%, averaged over 9 events). Every run of the common setup uses cross-entropy on hard labels. Shading marks the
teacher pipeline, which is compared with training on hand labels alone. Table~\ref{tab:s_bb_contrasts} gives the
paired differences.}}
\label{tab:s_bb_levels}
\scriptsize
\setlength{\tabcolsep}{3.5pt}
\begin{tabular}{lllccccc}
\toprule
Design & Parameters & Training data & S11 test & WorldFloods test & STURM-Flood & Held back & False alarms \\
\midrule
Micro student & 1,266 & 252 hand labels (no teacher) & 80.2 $\pm$ 1.0 & 79.0 $\pm$ 0.7 & 50.2 $\pm$ 0.7 & 58.0 $\pm$ 0.6 & 17.5 \\
 &  & 252 hand labels + teacher's maps & 79.5 $\pm$ 0.8 & 79.0 $\pm$ 0.4 & 48.0 $\pm$ 6.7 & 58.5 $\pm$ 0.6 & 16.2 \\
 &  & teacher labels on the same 252 images & 79.2 $\pm$ 1.0 & 78.8 $\pm$ 0.8 & 45.8 $\pm$ 6.5 & 58.5 $\pm$ 1.0 & 17.0 \\
 &  & 2,500 teacher-labeled GDACS images & 78.3 $\pm$ 1.6 & 78.7 $\pm$ 0.9 & 48.0 $\pm$ 1.6 & 59.4 $\pm$ 1.3 & 12.1 \\
\rowcolor{rowhl} &  & 2,500 teacher-labeled images + maps & 77.3 $\pm$ 1.0 & 78.0 $\pm$ 0.7 & 50.4 $\pm$ 0.4 & 58.1 $\pm$ 0.7 & 19.2 \\
\midrule
Tiny student & 12,230 & 252 hand labels (no teacher) & 81.2 $\pm$ 1.2 & 79.2 $\pm$ 0.8 & 45.8 $\pm$ 5.6 & 59.3 $\pm$ 0.5 & 11.2 \\
 &  & 252 hand labels + teacher's maps & 81.5 $\pm$ 0.7 & 79.2 $\pm$ 0.6 & 47.4 $\pm$ 1.9 & 59.1 $\pm$ 0.7 & 8.5 \\
 &  & teacher labels on the same 252 images & 80.6 $\pm$ 1.0 & 79.5 $\pm$ 0.4 & 44.4 $\pm$ 7.1 & 60.0 $\pm$ 0.1 & 9.0 \\
 &  & 2,500 teacher-labeled GDACS images & 80.2 $\pm$ 0.2 & 79.8 $\pm$ 0.2 & 48.0 $\pm$ 1.2 & 60.8 $\pm$ 0.7 & 5.9 \\
\rowcolor{rowhl} &  & 2,500 teacher-labeled images + maps & 79.6 $\pm$ 0.3 & 79.4 $\pm$ 0.2 & 48.8 $\pm$ 1.4 & 60.7 $\pm$ 0.2 & 4.3 \\
\midrule
EfficientViT-B0 & 0.7\,M & 252 hand labels (no teacher) & 74.8 $\pm$ 1.7 & 69.6 $\pm$ 3.9 & 47.2 $\pm$ 1.5 & 55.1 $\pm$ 1.2 & 11.8 \\
 &  & 252 hand labels + teacher's maps & 75.1 $\pm$ 1.3 & 74.0 $\pm$ 2.6 & 50.1 $\pm$ 1.2 & 57.3 $\pm$ 1.1 & 6.3 \\
 &  & teacher labels on the same 252 images & 75.2 $\pm$ 0.7 & 69.8 $\pm$ 2.1 & 46.1 $\pm$ 1.3 & 55.2 $\pm$ 1.6 & 9.4 \\
 &  & 2,500 teacher-labeled GDACS images & 75.5 $\pm$ 1.2 & 78.8 $\pm$ 1.4 & 48.2 $\pm$ 0.8 & 60.9 $\pm$ 0.2 & 10.6 \\
\rowcolor{rowhl} &  & 2,500 teacher-labeled images + maps & 76.1 $\pm$ 0.8 & 79.9 $\pm$ 0.5 & 50.3 $\pm$ 0.4 & 60.9 $\pm$ 0.5 & 7.7 \\
\midrule
MobileNetV3-Small & 3.55\,M & 252 hand labels (no teacher) & 76.8 $\pm$ 1.2 & 75.1 $\pm$ 1.4 & 48.3 $\pm$ 3.5 & 55.1 $\pm$ 0.5 & 15.6 \\
 &  & 252 hand labels + teacher's maps & 77.6 $\pm$ 1.3 & 77.9 $\pm$ 0.1 & 52.0 $\pm$ 0.7 & 56.8 $\pm$ 0.2 & 13.5 \\
 &  & teacher labels on the same 252 images & 76.8 $\pm$ 0.7 & 75.2 $\pm$ 0.6 & 50.8 $\pm$ 1.1 & 56.1 $\pm$ 0.4 & 17.2 \\
 &  & 2,500 teacher-labeled GDACS images & 76.2 $\pm$ 0.3 & 77.8 $\pm$ 1.0 & 49.7 $\pm$ 0.4 & 60.9 $\pm$ 0.2 & 5.9 \\
\rowcolor{rowhl} &  & 2,500 teacher-labeled images + maps & 76.6 $\pm$ 0.2 & 79.3 $\pm$ 0.3 & 51.3 $\pm$ 0.2 & 60.5 $\pm$ 0.4 & 3.4 \\
\midrule
LightFormer-ResNet18 & 11.43\,M & 252 hand labels (no teacher) & 75.6 $\pm$ 0.7 & 71.7 $\pm$ 0.4 & 48.8 $\pm$ 0.7 & 55.9 $\pm$ 0.7 & 11.0 \\
 &  & 252 hand labels + teacher's maps & 76.3 $\pm$ 1.8 & 73.8 $\pm$ 0.9 & 50.9 $\pm$ 0.7 & 57.3 $\pm$ 1.3 & 6.2 \\
 &  & teacher labels on the same 252 images & 75.1 $\pm$ 1.2 & 72.6 $\pm$ 0.8 & 48.9 $\pm$ 0.9 & 55.7 $\pm$ 1.3 & 7.8 \\
 &  & 2,500 teacher-labeled GDACS images & 74.4 $\pm$ 2.1 & 76.9 $\pm$ 2.5 & 44.9 $\pm$ 1.5 & 59.9 $\pm$ 0.2 & 17.0 \\
\rowcolor{rowhl} &  & 2,500 teacher-labeled images + maps & 75.6 $\pm$ 1.4 & 77.9 $\pm$ 0.6 & 48.6 $\pm$ 0.8 & 59.9 $\pm$ 0.4 & 14.3 \\
 &  & authors' setup, 252 hand labels & 78.3 $\pm$ 1.0 & 78.1 $\pm$ 1.9 & 50.2 $\pm$ 2.6 & 57.9 $\pm$ 0.2 & 11.1 \\
 &  & authors' setup, 2,500 teacher labels & 76.9 $\pm$ 1.0 & 77.8 $\pm$ 1.8 & 48.3 $\pm$ 2.2 & 61.5 $\pm$ 0.4 & 7.9 \\
\midrule
UNet-ResNet50 & 32.5\,M & 252 hand labels (no teacher) & 80.8 $\pm$ 0.3 & 78.0 $\pm$ 0.9 & 48.7 $\pm$ 1.5 & 56.9 $\pm$ 0.5 & 9.9 \\
 &  & 252 hand labels + teacher's maps & 79.7 $\pm$ 0.9 & 79.9 $\pm$ 0.5 & 50.8 $\pm$ 0.4 & 58.7 $\pm$ 0.6 & 5.0 \\
 &  & teacher labels on the same 252 images & 79.6 $\pm$ 0.9 & 77.9 $\pm$ 0.9 & 50.0 $\pm$ 1.6 & 58.3 $\pm$ 0.8 & 10.7 \\
 &  & 2,500 teacher-labeled GDACS images & 77.9 $\pm$ 0.7 & 77.7 $\pm$ 1.0 & 47.6 $\pm$ 0.7 & 61.1 $\pm$ 0.2 & 2.9 \\
\rowcolor{rowhl} &  & 2,500 teacher-labeled images + maps & 78.2 $\pm$ 1.4 & 79.9 $\pm$ 0.3 & 47.9 $\pm$ 0.8 & 61.6 $\pm$ 0.1 & 2.8 \\
\bottomrule
\end{tabular}
\end{table*}

\begin{table*}[t]
\revnewfloat
\centering
\caption{\add{\textbf{Paired comparisons for the six students.} Differences are left minus right, in IoU points, with
95\% intervals from resampling whole events and four seeds per setup. Positive values favor the left setting,
and an interval that excludes zero indicates a resolved difference.}}
\label{tab:s_bb_contrasts}
\scriptsize
\setlength{\tabcolsep}{3.5pt}
\begin{tabular}{llcccc}
\toprule
Design & Comparison & S11 test & WorldFloods test & STURM-Flood & Held back \\
\midrule
Micro student & maps on the 252 hand-labeled images & $-$0.7 [$-$1.6, +0.1] & 0.0 [$-$0.6, +0.8] & $-$2.1 [$-$4.0, $-$0.6] & +0.5 [+0.2, +0.9] \\
 & teacher $-$ human labels, same 252 images & $-$1.0 [$-$1.8, $-$0.3] & $-$0.2 [$-$0.7, +0.5] & $-$4.4 [$-$6.9, $-$2.3] & +0.5 [+0.2, +0.9] \\
 & teacher pipeline $-$ hand labels alone & $-$2.9 [$-$4.4, $-$1.7] & $-$1.0 [$-$2.3, +0.2] & +0.2 [$-$0.8, +1.3] & +0.1 [$-$0.7, +0.9] \\
 & 2,500 teacher-labeled $-$ 252 hand-labeled, both with maps & $-$2.2 [$-$3.0, $-$1.5] & $-$1.0 [$-$2.4, +0.2] & +2.4 [+0.6, +4.4] & $-$0.4 [$-$1.1, +0.3] \\
 & maps on the 2,500 teacher-labeled images & $-$1.0 [$-$1.6, $-$0.3] & $-$0.8 [$-$1.6, $-$0.1] & +2.4 [+0.8, +4.3] & $-$1.3 [$-$1.6, $-$0.9] \\
\midrule
Tiny student & maps on the 252 hand-labeled images & +0.4 [$-$0.3, +1.2] & 0.0 [$-$0.6, +0.6] & +1.6 [+0.6, +2.8] & $-$0.1 [$-$0.5, +0.2] \\
 & teacher $-$ human labels, same 252 images & $-$0.5 [$-$1.4, +0.6] & +0.3 [$-$0.4, +1.0] & $-$1.4 [$-$3.6, +0.3] & +0.7 [+0.3, +1.2] \\
 & teacher pipeline $-$ hand labels alone & $-$1.5 [$-$2.9, 0.0] & +0.2 [$-$1.3, +1.5] & +3.0 [+0.7, +5.5] & +1.4 [+0.6, +2.3] \\
 & 2,500 teacher-labeled $-$ 252 hand-labeled, both with maps & $-$1.9 [$-$2.9, $-$0.9] & +0.2 [$-$1.1, +1.6] & +1.4 [$-$0.4, +3.3] & +1.5 [+0.7, +2.5] \\
 & maps on the 2,500 teacher-labeled images & $-$0.6 [$-$1.3, +0.1] & $-$0.4 [$-$0.8, 0.0] & +0.8 [$-$0.4, +2.3] & $-$0.1 [$-$0.5, +0.3] \\
\midrule
EfficientViT-B0 & maps on the 252 hand-labeled images & +0.2 [$-$0.2, +0.7] & +4.4 [+1.7, +7.2] & +2.9 [+1.4, +4.5] & +2.2 [+1.4, +3.1] \\
 & teacher $-$ human labels, same 252 images & +0.4 [$-$0.3, +1.5] & +0.2 [$-$3.9, +3.2] & $-$1.1 [$-$2.4, +0.4] & +0.2 [$-$0.9, +1.0] \\
 & teacher pipeline $-$ hand labels alone & +1.3 [$-$1.4, +4.2] & +10.3 [+3.1, +20.7] & +3.2 [+0.2, +6.4] & +5.8 [+4.2, +7.8] \\
 & 2,500 teacher-labeled $-$ 252 hand-labeled, both with maps & +1.0 [$-$1.4, +3.7] & +5.9 [+0.9, +15.1] & +0.2 [$-$2.5, +3.1] & +3.6 [+2.2, +5.2] \\
 & maps on the 2,500 teacher-labeled images & +0.6 [$-$0.7, +2.0] & +1.1 [$-$0.4, +3.5] & +2.1 [+0.6, +4.6] & 0.0 [$-$0.3, +0.3] \\
\midrule
MobileNetV3-Small & maps on the 252 hand-labeled images & +0.8 [$-$0.5, +2.2] & +2.9 [+1.3, +4.7] & +3.7 [+1.9, +5.9] & +1.7 [+1.2, +2.1] \\
 & teacher $-$ human labels, same 252 images & 0.0 [$-$1.0, +1.0] & +0.1 [$-$1.6, +1.4] & +2.5 [+1.1, +4.2] & +1.0 [+0.6, +1.4] \\
 & teacher pipeline $-$ hand labels alone & $-$0.3 [$-$4.7, +3.3] & +4.3 [+1.6, +7.6] & +3.0 [+0.6, +5.8] & +5.4 [+3.7, +7.6] \\
 & 2,500 teacher-labeled $-$ 252 hand-labeled, both with maps & $-$1.1 [$-$4.7, +2.3] & +1.4 [+0.2, +3.0] & $-$0.7 [$-$2.3, +0.6] & +3.7 [+2.3, +5.6] \\
 & maps on the 2,500 teacher-labeled images & +0.3 [$-$0.4, +1.3] & +1.5 [$-$0.5, +5.0] & +1.6 [+0.3, +3.6] & $-$0.4 [$-$0.9, 0.0] \\
\midrule
LightFormer-ResNet18 & maps on the 252 hand-labeled images & +0.7 [$-$0.6, +2.5] & +2.1 [+0.1, +5.4] & +2.1 [+0.1, +4.8] & +1.4 [+0.7, +2.2] \\
 & teacher $-$ human labels, same 252 images & $-$0.4 [$-$1.2, +0.4] & +0.9 [$-$0.4, +2.7] & +0.2 [$-$1.0, +1.3] & $-$0.3 [$-$1.6, +1.2] \\
 & teacher pipeline $-$ hand labels alone & 0.0 [$-$3.0, +3.4] & +6.2 [0.0, +17.5] & $-$0.2 [$-$2.8, +2.6] & +4.0 [+2.5, +5.9] \\
 & 2,500 teacher-labeled $-$ 252 hand-labeled, both with maps & $-$0.7 [$-$2.6, +1.2] & +4.1 [$-$0.5, +12.4] & $-$2.3 [$-$4.3, $-$0.5] & +2.6 [+1.6, +3.9] \\
 & maps on the 2,500 teacher-labeled images & +1.2 [$-$0.5, +3.4] & +1.1 [$-$0.4, +2.8] & +3.7 [+1.3, +6.9] & +0.1 [$-$0.2, +0.4] \\
 & authors' $-$ common setup, 252 hand labels & +2.7 [$-$0.2, +6.0] & +6.3 [$-$0.4, +18.4] & +1.4 [$-$1.3, +4.7] & +1.9 [+0.2, +4.0] \\
 & authors' $-$ common setup, 2,500 teacher labels & +2.5 [+0.5, +4.7] & +0.9 [$-$0.3, +2.2] & +3.4 [+0.5, +7.1] & +1.6 [+0.8, +2.5] \\
\midrule
UNet-ResNet50 & maps on the 252 hand-labeled images & $-$1.1 [$-$2.8, +0.1] & +1.8 [+0.5, +3.7] & +2.1 [$-$0.3, +5.0] & +1.8 [+1.1, +2.6] \\
 & teacher $-$ human labels, same 252 images & $-$1.1 [$-$2.5, +0.6] & $-$0.1 [$-$0.4, +0.2] & +1.3 [+0.1, +2.7] & +1.4 [+0.5, +2.6] \\
 & teacher pipeline $-$ hand labels alone & $-$2.6 [$-$5.7, +1.6] & +1.9 [$-$0.1, +5.1] & $-$0.8 [$-$4.9, +3.5] & +4.7 [+3.1, +6.8] \\
 & 2,500 teacher-labeled $-$ 252 hand-labeled, both with maps & $-$1.5 [$-$4.1, +2.1] & +0.1 [$-$1.3, +1.9] & $-$2.9 [$-$6.3, $-$0.2] & +2.9 [+1.7, +4.4] \\
 & maps on the 2,500 teacher-labeled images & +0.2 [$-$0.9, +1.3] & +2.3 [+0.1, +6.3] & +0.4 [$-$0.8, +1.6] & +0.5 [+0.1, +1.0] \\
\bottomrule
\end{tabular}
\end{table*}

\begin{figure*}[t]
\revnewfloat
\centering
\includegraphics[width=\linewidth]{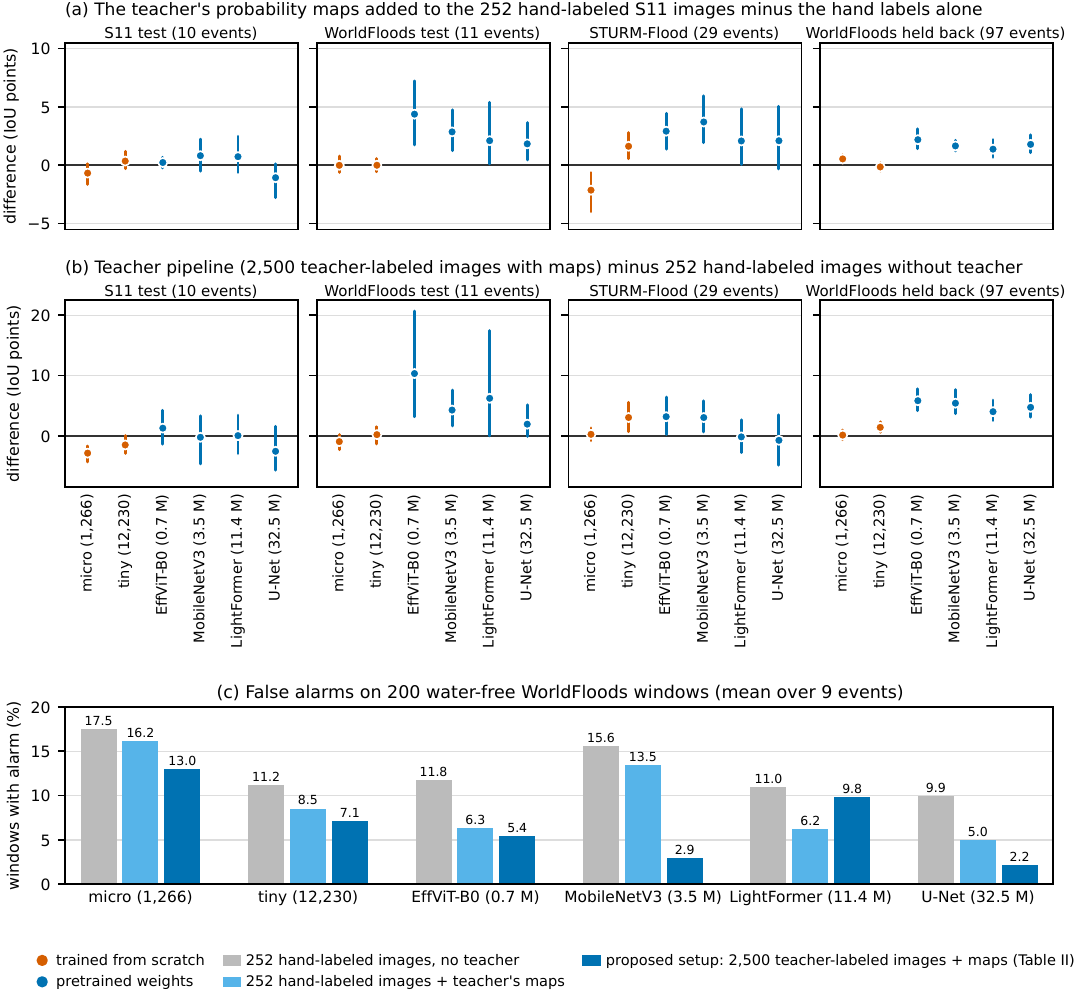}
\caption{\add{\textbf{What the teacher adds to six students, ordered by size.} (a) Adding the teacher's probability maps
to the 252 hand-labeled Sen1Floods11 images, relative to training on hand labels alone. (b) The full teacher pipeline,
using 2,500 teacher-labeled images with maps, relative to training on the 252 hand-labeled images without a teacher.
(c) False alarms on 200 water-free WorldFloods windows, averaged over 9 events, for the two setups in (a) and the proposed
setup in Table~II of the main article. In (a) and (b), every run uses 512-pixel images, 7,800 updates and four seeds. Distillation has weight 0.5 when the teacher's maps are used. Lines show 95\% intervals
from resampling whole events.}}
\label{fig:s_teacher_by_size}
\end{figure*}

\begin{figure*}[t]
\revnewfloat
\centering
\includegraphics[width=\linewidth]{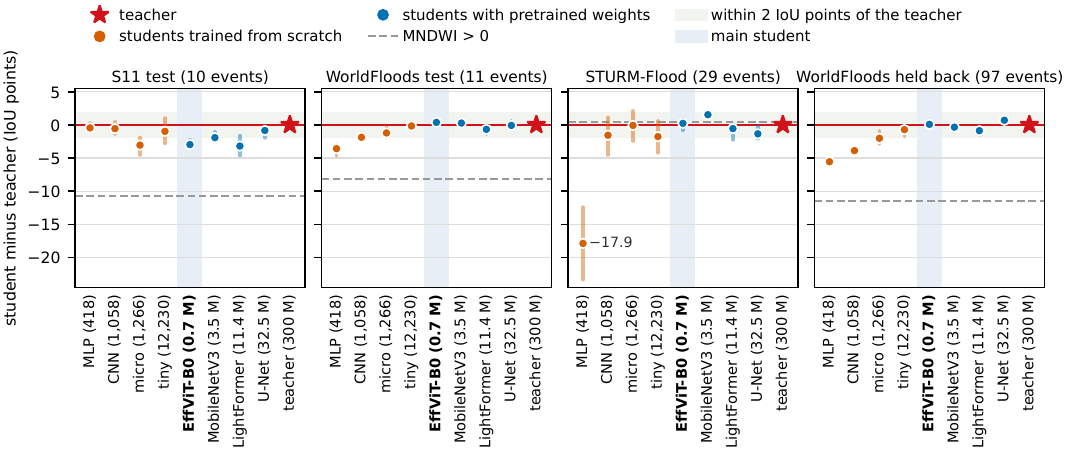}
\caption{\add{\textbf{Model size and water IoU per flood event relative to the teacher.} Every model uses the proposed
setup on the 2,500 teacher-labeled images, with 7,800 updates and $512\!\times\!512$ inputs. Dots show the means over four seeds, and lines span the lowest to highest seed.
The red star and line mark the teacher, the dashed line marks MNDWI at threshold zero, and the band marks scores
within 2 IoU points of the teacher. The shaded column marks the main student. Orange denotes training from scratch,
and blue denotes pretrained weights.}}
\label{fig:s_size_ladder}
\end{figure*}

\begin{newtext}
\section{Students of Other Sizes and Designs}
\label{sec:s_backbones}

Each of six students was trained in five setups, all with the main student's training settings: $512\!\times\!512$ images,
AdamW, batch size 16, 7,800 updates, four seeds and checkpoint selection on the Sen1Floods11 validation split.
Every run uses cross-entropy on the hard labels. Setup H uses the 252 hand labels. Setup K adds the teacher's probability
maps to H, with weight 0.5 and temperature 4. Setup T uses the teacher's labels on the same 252 images. Setup G uses the
2,500 teacher-labeled GDACS images, and setup GK adds the teacher's maps to G. The four pretrained students use a
learning rate of $5\times10^{-5}$. The tiny and micro students train from scratch with $5\times10^{-3}$.
All models were fixed on the validation split before any were evaluated outside the training data.
Table~\ref{tab:s_bb_levels} reports their scores, and Table~\ref{tab:s_bb_contrasts} reports the paired comparisons
and intervals. The findings are summarized below.

The six students reproduce the main student's findings. Teacher labels perform as well as human labels on the same
images for all six. The two label sources differ by at most 1.4 IoU points, except on STURM-Flood, where they differ by a few points in either direction.
Adding the teacher's probability maps to the 252 hand-labeled images improves every pretrained student on all three
test sets outside the training data. Adding the maps to the 2,500 teacher-labeled images still improves each of these
students on WorldFloods and STURM-Flood. For the two students trained from scratch, the maps produce gains and losses
of similar size. The larger pool provides the greatest benefit on the held-back events. There, the 2,500
teacher-labeled images outperform the 252 hand-labeled images, both with maps, for every student except the micro
student. On STURM-Flood, the larger pool reduces the scores of LightFormer and UNet-ResNet50 by 2 to 3 IoU points.

LightFormer was also trained using its authors' setup~\cite{chen2026lightformer}, on the 252 hand labels and on the
2,500 teacher labels without the teacher's maps. This setup uses AdamW with learning rates of $6\times10^{-4}$ for the
encoder and $9\times10^{-3}$ for the decoder, and weight decay 0.01. The objective combines cross-entropy with label
smoothing, Dice loss and an auxiliary loss weighted by 0.4. Augmentation comprises random rescaling by 0.5 to 2 with
$512\!\times\!512$ crops, flips, rotations, brightness and contrast changes, and distortion. The batch size is 16.
Training stops after 8 epochs without improvement on the Sen1Floods11 validation split. This ended training after
330 to 495 updates on the 252 images and 3,120 to 3,744 updates on the 2,500 images. These settings follow the authors' preprint~\cite{chen2025lightformer} and
their public code, which we adapted to six bands and two classes. Because the optimizer, losses, augmentation, training
length and checkpoint selection all differ from our common setup, this comparison shows only their combined effect. It appears in the last two rows of the LightFormer block in
Table~\ref{tab:s_bb_contrasts}.
\end{newtext}

\begin{newtext}
\label{sec:s_students}
Fig.~\ref{fig:s_teacher_by_size} shows the two main comparisons and false alarms for the six students.
Fig.~\ref{fig:s_size_ladder} shows the performance of every model trained with the proposed setup as a function of
model size. Section~V-C of the main article summarizes the results.
\end{newtext}

\begin{figure*}[t]
\revnewfloat
\centering
\includegraphics[width=\linewidth]{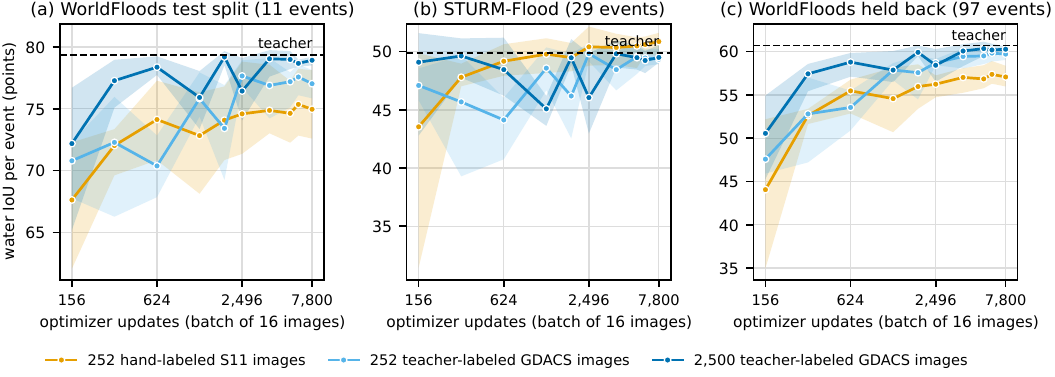}
\caption{\add{\textbf{Learning curves of the main student} on test sets outside the training data: (a) WorldFloods,
(b) STURM-Flood and (c) the 97 held-back events. Water IoU per event is shown for weights saved at ten fixed updates
on a logarithmic update axis. Lines show the means over four seeds, bands span the lowest to highest seed, and the
dashed line marks the teacher. No checkpoint was selected using these sets. Every saved point is shown without
smoothing. After the first pass over the images, one pool's mean still changes by up to 5.4 IoU points between successive
saved points, making the second half of training the fairer comparison.}}
\label{fig:s_convergence}
\end{figure*}

\begin{figure*}[tp]
\revnewfloat
\centering
\includegraphics[width=\linewidth]{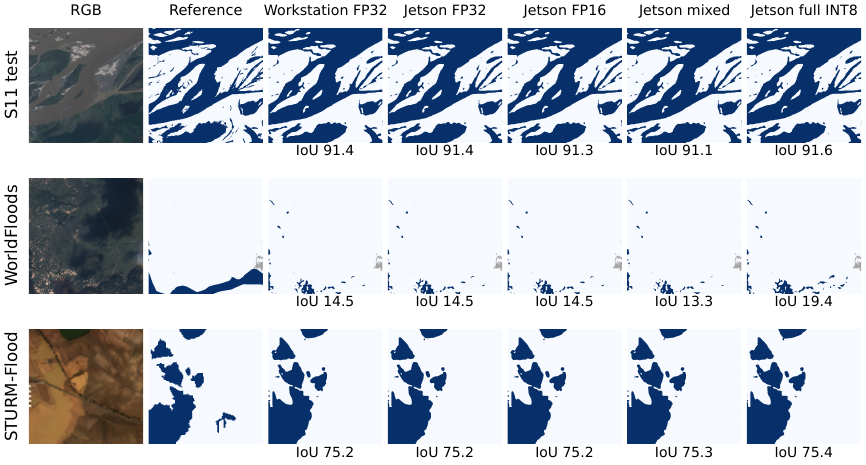}
\caption{\add{\textbf{Tiny deployment predictions}, seed 42, for the three scenes of Fig.~3 of the main article in
every number format. Dark blue denotes predicted water, and gray denotes ignored pixels. Each score
is the scene's water IoU. The formats change less than 1\% of the scored pixels; the score moves most on the low-water WorldFloods scene. Differences between formats reflect the full deployment chain, including QAT. These panels
do not replace the four-seed benchmark.}}
\label{fig:qual_tiny_precision}
\end{figure*}

\begin{figure*}[tp]
\revnewfloat
\centering
\includegraphics[width=\linewidth]{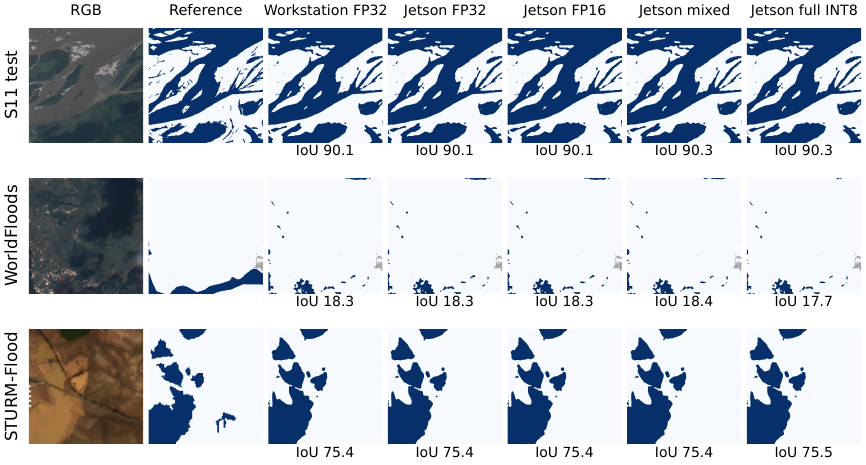}
\caption{\add{\textbf{Micro deployment predictions}, seed 42, for the three scenes of Fig.~3 of the main article in
every number format. Dark blue denotes predicted water, and gray denotes ignored pixels. Each score
is the scene's water IoU. The formats change less than 1\% of the scored pixels. Differences between formats reflect
the full deployment chain, including QAT. These panels
do not replace the four-seed benchmark.}}
\label{fig:qual_micro_precision}
\end{figure*}

\begin{figure*}[tp]
\revnewfloat
\centering
\includegraphics[width=0.714\linewidth]{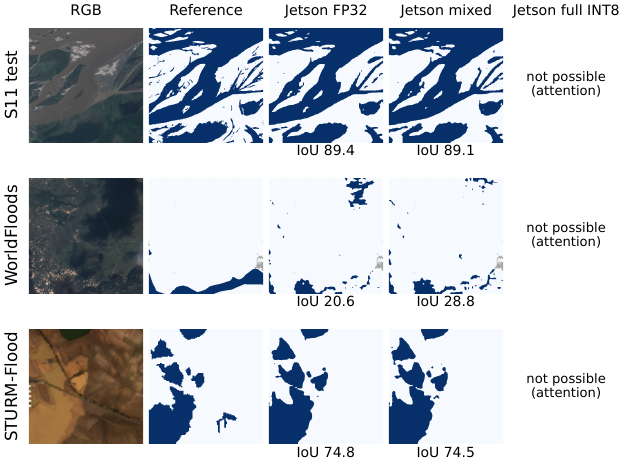}
\caption{\add{\textbf{Deployed main-student predictions} for EfficientViT-B0 with ReLU6
(Section~\ref{sec:s_deploy}), seed 42, on the three scenes of Fig.~3 of the main article. Its FP32 and mixed INT8/FP16 engines
run on the Jetson. Dark blue denotes predicted water, and gray denotes ignored pixels.
Each score is the scene's water IoU. The mixed engine changes 0.8\% to 3.5\% of the scored pixels, most on the
WorldFloods scene. Differences between the engines reflect the full deployment chain, including QAT,
and cannot be attributed solely to rounding. These panels do not replace the four-seed benchmark.}}
\label{fig:qual_b0_precision}
\end{figure*}

\begin{newtext}
\section{Learning Curves outside the Training Data}
\label{sec:s_convergence}

Twelve additional runs train the main student for 7,800 updates with the main loss, comprising 0.5 focal loss and
0.5 distillation loss. We use four seeds for each of three pools: the 252 hand-labeled Sen1Floods11 images,
252 teacher-labeled GDACS images, and the 2,500 teacher-labeled GDACS images that include those 252 images.
Weights were saved at ten fixed updates from 156 to 7,800, regardless of the scores. Each saved model was evaluated
once on the WorldFloods test split, STURM-Flood and the 97 held-back events. The final saved models were also evaluated
on the Sen1Floods11 test split. No checkpoint was selected using these sets. Fig.~\ref{fig:s_convergence} shows the curves.

These runs repeat one comparison from Table~III of the main article. At the end of training, the difference between
the 2,500 teacher-labeled images and the 252 hand-labeled images is $+4.0$ $[-0.7, +12.6]$ IoU points on WorldFloods
and $-1.4$ $[-3.5, +0.5]$ points on STURM-Flood, against $+4.9$ and $-0.6$ IoU points in Table~III. The direction
agrees, but here the WorldFloods interval includes zero. The new runs also show how performance changes during training. On WorldFloods, the
2,500-image pool leads at nine of the ten saved points and reaches 77 IoU points within 312 updates on average. On STURM-Flood, the
curves cross repeatedly in the first half of training and finish within 2 IoU points of one another. From 3,900
updates onward, the 252 hand-labeled images stay 0.5 to 1.0 IoU points above the teacher and 0.5 to 1.4 points above the
2,500-image pool. On the held-back events, the 2,500-image pool leads
at eight of the ten saved points and finishes 0.4 IoU points below the teacher. The 252 hand-labeled images never come
within 3 IoU points of the teacher and finish $+3.2$ $[+1.7, +5.1]$ points behind the 2,500-image pool.

The two teacher-labeled pools differ only in image count because the 252 images are a subset of the 2,500.
Their final difference, larger minus smaller pool, is $+1.9$ $[-0.6, +6.3]$ IoU points on WorldFloods,
$+0.2$ $[-0.8, +1.1]$ on STURM-Flood, $+0.6$ $[-0.1, +1.1]$ on the held-back events and $+0.8$ $[-1.3, +2.8]$ on the
Sen1Floods11 test split.
\end{newtext}

\section{Scaling Beyond a Fixed Manual Annotation Set}
\label{sec:ceiling}

This section formalizes the narrow information statement used in \chg{Section~V}{Sections III-C and V-B} of the main
article. Let $\mathcal{D}=\{(x_i,T(x_i))\}_{i=1}^{M}$ be the
pseudo-label dataset, where $T(x)=p_T(x)$ denotes the frozen teacher output.

\begin{proposition}[No independent label information]
Fix the teacher $T$ (including its parameters) and the input set $\{x_i\}$. Any
auxiliary training signal computed deterministically from the teacher and these inputs
(for example, soft outputs, input-sensitivity maps, output variance, saliency masks,
or intermediate teacher features) is a function of $(T,\{x_i\})$. Conditional on
$(T,\{x_i\})$, it therefore contains no additional information about the unknown
ground-truth labels. Such signals can change optimization, regularization, or inductive
bias, but they do not introduce an independent annotation source.
\label{prop:ceiling}
\end{proposition}

Let $A=f(T,X)$ denote any deterministic auxiliary signal derived from the frozen teacher and
inputs. Then
\begin{equation}
  H(A\mid T,X)=0,
\end{equation}
and therefore
\begin{equation}
  I(Y;A\mid T,X)
  =
  H(A\mid T,X)-H(A\mid Y,T,X)
  =
  0.
  \label{eq:no_independent_info}
\end{equation}
Thus $A$ may alter the optimization path or inductive bias, but conditional on the teacher and
inputs it cannot introduce an independent source of ground-truth label information.

The proposition is an information statement, not a bound on benchmark accuracy and not a
claim that all teacher-derived objectives have the same neural-network minimiser. A useful
reference is the best KL imitator available to the student class $\mathcal{H}$,
\begin{equation}
  S^*_{\mathrm{KD}}=\operatorname*{arg\,min}_{S\in\mathcal{H}}
  \mathbb{E}_{x}\,\mathrm{KL}\!\left(p_T(x)\,\middle\|\,S(x)\right),
  \label{eq:ceiling}
\end{equation}
but a mixed or auxiliary objective can select a different function through finite-sample
optimization or regularization. That different function may score above or below the
teacher on a particular ground-truth benchmark. Proposition~\ref{prop:ceiling} only says
that such a gain is not caused by a new, independent label signal. This distinction is
consistent with the disagreement audit of Section~\ref{app:diag}: some deviations from the
teacher help, while systematic annotation semantics are largely retained.

\paragraph*{The KD${+}$CE target}
For a fixed temperature, Gibbs' inequality gives the unrestricted minimizer of the KD term
at $q=p_T$. Adding a hard teacher-argmax cross-entropy term with weight $1-\alpha$ gives,
per pixel,
\begin{equation}
  q^*(x)=\alpha\,p_T(x)+(1-\alpha)\,\delta_{\hat y_T(x)},
  \label{eq:optimum}
\end{equation}
where $\delta_{\hat y_T}$ is one-hot at the teacher argmax. Thus $\alpha$ changes the
mixture between two teacher-derived targets. Changing the temperature changes the soft
teacher distribution $p_T$ itself; neither operation introduces labels independent of the
teacher. Equation~\ref{eq:optimum} is a per-pixel result for this specific KD${+}$CE
mixture and should not be read as a statement that OFA or other auxiliary objectives share
its parameter-space optimum. The main experiments use focal modulation for the hard-target
term, so their exact optimum need not equal the convex mixture in Eq.~\ref{eq:optimum};
this does not affect Proposition~\ref{prop:ceiling}, because both targets remain deterministic
functions of the same frozen teacher.

\paragraph*{Binary soft targets}
For binary segmentation ($C\!=\!2$), a normalized class distribution has one independent
probability degree of freedom per pixel. The teacher's soft value can retain confidence
information that its argmax removes, but its numerical precision should not be confused
with an additional source of ground-truth supervision. Without assumptions about teacher
calibration, $p_T$ alone does not provide a quantitative bound on how much class
information the soft target adds over the argmax. We therefore interpret the measured
\chg{CE-to-Logit-KD differences in Tables~I and~II}{effects of the probability maps in Table~III} of the main article as empirical
sample-efficiency effects, not estimates of an information bound. \add{These effects are largest with few images, consistent with
this interpretation.} Multi-class tasks may behave differently and are outside the scope of this study.

\begin{newtext}
\section{Measures: Original Experimental Setup and This Paper}
\label{sec:s_measures}

An earlier version of this study reported the mean IoU of water and land over all pixels (mIoU) and excluded permanent
water from the STURM-Flood and WorldFloods scores. This paper reports water IoU per flood event and scores all surface
water. Table~\ref{tab:s_measures} shows three models under each measure. Two changes affect the values. First, water
IoU excludes the easier land class, whose IoU of 86 to 97 for the main student increases mIoU. Second, averaging over
events gives each event equal weight. STURM-Flood contains many events with little water, which is hard to map, so its
per-event score falls to about 50. On WorldFloods, a few large scenes account for most water pixels and reduce
the pooled score because every model performs worse on these scenes. Differences between models change little.
On the Sen1Floods11 test split, the teacher and main student have pooled water IoU scores of 81.7 and 78.5,
close to the earlier values of 82.2 and 78.7.
\end{newtext}
\begin{table}[t]
\revnewfloat
\centering
\caption{\add{\textbf{Measures used in the earlier version and this paper.} The earlier version reports mIoU pooled over
all pixels, excludes permanent water on WorldFloods and STURM-Flood, and uses the One-for-All student. This paper scores
all surface water and uses the logit-distillation student. The two objectives differ by less than 1 point.}}
\label{tab:s_measures}
\scriptsize
\setlength{\tabcolsep}{3pt}
\begin{tabular}{llcccc}
\toprule
Model & Test set & Earlier & mIoU, & Water IoU, & Water IoU, \\
 & & mIoU & pooled & pooled & per event \\
\midrule
Teacher & S11 test & 89.7 & 89.5 & 81.7 & 79.1 \\
 & WorldFloods test & 77.1 & 81.5 & 73.9 & 79.4 \\
 & STURM-Flood & 75.5 & 76.6 & 67.7 & 49.9 \\
\midrule
Main student & S11 test & 87.7 & 87.6 & 78.5 & 76.1 \\
 & WorldFloods test & 76.4 & 82.5 & 75.4 & 79.7 \\
 & STURM-Flood & 75.6 & 77.2 & 68.3 & 50.1 \\
\midrule
MNDWI $>0$ & S11 test & 78.9 & 78.9 & 65.0 & 68.3 \\
 & WorldFloods test & 76.3 & 81.4 & 74.8 & 71.2 \\
 & STURM-Flood & 74.7 & 76.0 & 67.4 & 50.4 \\
\bottomrule
\end{tabular}
\end{table}
\begin{table*}[t]
\revnewfloat
\centering
\caption{\add{\textbf{Pooled and per-event water IoU.} Water IoU pooled over all pixels of a test set compared with water
IoU per flood event (points, mean of four seeds), for the models of Table~II of the main article.}}
\label{tab:s_pooled}
\footnotesize
\begin{tabular}{lcccccc}
\toprule
 & \multicolumn{2}{c}{S11 test} & \multicolumn{2}{c}{WorldFloods test} & \multicolumn{2}{c}{STURM-Flood} \\
\cmidrule(lr){2-3}\cmidrule(lr){4-5}\cmidrule(lr){6-7}
Model & pooled & per event & pooled & per event & pooled & per event \\
\midrule
Teacher & 81.7 & 79.1 & 73.9 & 79.4 & 67.7 & 49.9 \\
MNDWI $>0$ & 65.0 & 68.3 & 74.8 & 71.2 & 67.4 & 50.4 \\
\midrule
Main student & 78.5 & 76.1 & 75.4 & 79.7 & 68.3 & 50.1 \\
Tiny student & 80.6 & 78.1 & 76.5 & 79.2 & 67.3 & 48.1 \\
Micro student & 78.7 & 76.0 & 75.8 & 78.1 & 68.1 & 49.8 \\
Main student, no teacher & 77.5 & 74.8 & 61.5 & 69.6 & 63.4 & 47.2 \\
Main student, with teacher's maps & 78.3 & 75.7 & 67.9 & 74.9 & 67.2 & 50.9 \\
\midrule
MobileNetV3-Small & 79.3 & 77.1 & 75.0 & 79.6 & 69.0 & 51.4 \\
LightFormer-ResNet18 & 78.6 & 75.9 & 73.8 & 78.7 & 68.3 & 49.3 \\
UNet-ResNet50 & 80.9 & 78.2 & 74.7 & 79.3 & 67.4 & 48.5 \\
MobileNetV3-Small, no teacher & 79.0 & 76.8 & 71.9 & 75.1 & 61.1 & 48.3 \\
LightFormer-ResNet18, no teacher & 77.4 & 75.6 & 63.8 & 71.7 & 65.2 & 48.8 \\
LightFormer, authors' setup, no teacher & 80.2 & 78.3 & 75.7 & 78.1 & 67.8 & 50.2 \\
UNet-ResNet50, no teacher & 82.8 & 80.8 & 74.1 & 78.0 & 64.0 & 48.7 \\
Tiny student, no teacher & 83.4 & 81.2 & 76.1 & 79.2 & 62.9 & 45.8 \\
Micro student, no teacher & 82.2 & 80.2 & 77.9 & 79.0 & 68.2 & 50.2 \\
\bottomrule
\end{tabular}
\end{table*}
\begin{newtext}

Table~\ref{tab:s_pooled} compares pooled and per-event water IoU for every model of Table~II of the main article.
Pooling increases every learned model's score by 1.9 to 2.8 IoU points on the Sen1Floods11 test split and by 12.8 to
19.2 points on STURM-Flood. It reduces the scores by 1.0 to 8.1 IoU points on WorldFloods. The comparison between the
main student and teacher remains unchanged. With pooling, the main student is 3.1 IoU points below the teacher on the
Sen1Floods11 test split, and 1.5 and 0.6 points above the teacher on WorldFloods and STURM-Flood. With per-event
averaging, the main student is 3.0 points below the teacher on the Sen1Floods11 test split, and 0.4 and 0.2 points above
the teacher on WorldFloods and STURM-Flood.
\end{newtext}

\begin{newtext}
Table~\ref{tab:s_bridge} isolates the change from scoring flood water to scoring all surface water while keeping the
model weights fixed. The teacher, MNDWI and the earlier student are evaluated under both rules using the present code.
The first column can therefore differ from the reported values in Table~\ref{tab:s_measures} by up to 0.7 mIoU points.
Scoring all surface water improves every method on both test sets. The model ordering remains unchanged except on
WorldFloods.
\end{newtext}
\begin{table}[t]
\revnewfloat
\centering
\caption{\add{\textbf{Scoring flood water only (earlier rule) and all surface water (this paper).} The same model weights
are used for both rules. Scores are mIoU over all pixels and water IoU per event under the current rule, which is this
paper's primary measure.}}
\label{tab:s_bridge}
\scriptsize
\setlength{\tabcolsep}{3pt}
\begin{tabular}{llccc}
\toprule
Model & Test set & Flood water, & All water, & All water, \\
 & & mIoU & mIoU & water IoU per event \\
\midrule
Teacher & STURM-Flood & 75.4 & 76.6 & 49.9 \\
Teacher & WorldFloods & 77.2 & 81.5 & 79.4 \\
MNDWI $>0$ & STURM-Flood & 74.7 & 76.0 & 50.4 \\
MNDWI $>0$ & WorldFloods & 75.9 & 81.4 & 71.2 \\
Earlier student & STURM-Flood & 76.0 & 77.3 & 50.3 \\
Earlier student & WorldFloods & 77.1 & 82.7 & 79.5 \\
\bottomrule
\end{tabular}
\end{table}

\ifdefined\ARXIVCOMBINED
\else
\bibliographystyle{IEEEtran}
\bibliography{references}
\end{document}
\fi

\fi

\end{document}